\documentclass[10pt,twocolumn]{article}
\PassOptionsToPackage{table}{xcolor}

\usepackage[utf8]{inputenc}
\usepackage[T1]{fontenc}
\usepackage{times}
\usepackage{amsmath,amssymb,amsfonts,amsthm}
\usepackage{graphicx}
\graphicspath{{image_assets/}}
\usepackage{tikz}
\usepackage{pgfplots}
\pgfplotsset{compat=1.18}
\usetikzlibrary{arrows.meta,positioning,calc,patterns,decorations.pathreplacing,decorations.pathmorphing,shapes.geometric,fit,backgrounds,shadows}
\usepackage{algorithm}
\usepackage{algpseudocode}
\usepackage{booktabs}
\usepackage{multirow}
\usepackage{xcolor}
\usepackage{eso-pic}

\usepackage[
  colorlinks=true,
  linkcolor=blue!70!black,
  citecolor=blue!50!black,
  urlcolor=blue!70!black,
  pdftitle={Continual Field-Adaptive Models (CFAMs) for Post-Deployment Physical AI: Autonomous Few-Shot Continual On-Device Adaptation Post-Deployment},
  pdfauthor={Amarjot Singh et al.},
  pdfsubject={Continual learning for deployed physical AI systems},
  pdfkeywords={continual learning, physical AI, robotics, edge AI}
]{hyperref}
\usepackage{cleveref}
\crefname{appendix}{Appendix}{Appendices}%
\Crefname{appendix}{Appendix}{Appendices}
\usepackage{microtype}
\usepackage[margin=0.70in]{geometry}
\usepackage{float}
\usepackage{placeins}
\usepackage{dblfloatfix}
\usepackage{caption}
\usepackage{subcaption}
\usepackage{enumitem}
\usepackage{tabularx}

\newcommand{\CFUR}{\textsc{cfur}}          %

\newcommand{\vz}{\mathbf{z}}          %
\newcommand{\va}{\mathbf{a}}          %
\newcommand{\vh}{\mathbf{h}}          %
\newcommand{\sal}{\kappa}             %
\newcommand{\binarize}{\textrm{bin}}    %

\newcommand{\rcu}{u}                   %
\newcommand{\rcufield}{\mathcal{F}}    %
\newcommand{\resid}{\Delta\va}         %
\newcommand{\srad}{\rho}               %
\newcommand{\trust}{q}                 %
\newcommand{\phase}{\psi}              %

\newcommand{\abase}{\va_{\text{base}}}

\newcommand{\aout}{\va_{\text{out}}}

\newcommand{\astaraction}{\va^{*}}
\newcommand{\aanchor}{\va_{\text{anchor}}}     %
\newcommand{\aadd}{\va_{\text{add}}}           %

\newcommand{\alphaaction}{\alpha_{\text{action}}}

\definecolor{corrcolor}{RGB}{239,68,68}
\definecolor{augcolor}{RGB}{74,222,128}
\definecolor{novcolor}{RGB}{139,92,246}
\definecolor{passcolor}{RGB}{148,163,184}
\definecolor{tabhead}{RGB}{226,233,242}
\definecolor{tabzebra}{RGB}{245,247,250}
\definecolor{abstractbg}{RGB}{239,247,255}
\definecolor{abstractborder}{RGB}{116,151,181}
\definecolor{abstracttitle}{RGB}{54,91,122}
\providecommand{\PH}[1]{\textcolor{red}{#1}}%
\makeatletter
\providecommand{\CFAMmiss}[2]{\textbf{\textcolor{red}{[#1: #2]}}}
\newcommand{\CFAMlookup}[2]{%
  \expandafter\ifx\csname cfam@#1@#2\endcsname\relax
    \CFAMmiss{#1}{#2}%
  \else
    \csname cfam@#1@#2\endcsname
  \fi}

\newcommand{\ConfigValue}[1]{\CFAMlookup{CONFIG}{#1}}
\makeatother

\expandafter\def\csname cfam@RESULT@EXP-06P.packet-bytes-max\endcsname{10{,}241}
\expandafter\def\csname cfam@RESULT@EXP-06P.packet-bytes-mean\endcsname{10{,}235}
\expandafter\def\csname cfam@RESULT@EXP-06P.packet-bytes-min\endcsname{10{,}225}
\expandafter\def\csname cfam@RESULT@EXP-06P.packet-bytes-sd\endcsname{5.95}
\expandafter\def\csname cfam@RESULT@EXP-10R.false-activation-at-10000\endcsname{0}
\expandafter\def\csname cfam@RESULT@EXP-10R.mean-active-set-at-10000\endcsname{1}
\expandafter\def\csname cfam@RESULT@EXP-10R.query-latency-ms-at-10000\endcsname{34.98}
\expandafter\def\csname cfam@RESULT@EXP-10R.top1-accuracy-at-10000\endcsname{1}

\expandafter\def\csname cfam@DERIVED@packet.kib\endcsname{9.99}
\expandafter\def\csname cfam@DERIVED@retrieval.scan-growth-factor\endcsname{78}

\expandafter\def\csname cfam@LIT@memoryvla.se-bridge\endcsname{71.9}
\expandafter\def\csname cfam@LIT@memoryvla.se-fractal\endcsname{72.7}
\expandafter\def\csname cfam@LIT@memoryvla.se-fractal-vm\endcsname{77.7}
\expandafter\def\csname cfam@LIT@memoryvla.se-fractal-va\endcsname{67.7}
\expandafter\def\csname cfam@LIT@memoryvla.libero\endcsname{96.5}
\expandafter\def\csname cfam@LIT@memoryvla.real\endcsname{84.0}
\expandafter\def\csname cfam@LIT@cronusvla.widowx-vm\endcsname{60.4}
\expandafter\def\csname cfam@LIT@cronusvla.google-vm\endcsname{78.6}
\expandafter\def\csname cfam@LIT@cronusvla.google-va\endcsname{73.8}
\expandafter\def\csname cfam@LIT@cronusvla.simplerenv-avg\endcsname{70.9}
\expandafter\def\csname cfam@LIT@cronusvla.libero-wrist\endcsname{97.0}
\expandafter\def\csname cfam@LIT@cronusvla.libero-nowrist\endcsname{92.2}
\expandafter\def\csname cfam@LIT@cronusvla.real\endcsname{72.6}
\expandafter\def\csname cfam@LIT@dreamvla.se-bridge\endcsname{71.4}
\expandafter\def\csname cfam@LIT@dreamvla.se-fractal\endcsname{60.5}
\expandafter\def\csname cfam@LIT@dreamvla.libero\endcsname{97.2}
\expandafter\def\csname cfam@LIT@dreamvla.real\endcsname{71.4}
\expandafter\def\csname cfam@LIT@openvlaoft.se-fractal-repro\endcsname{54.3}
\expandafter\def\csname cfam@LIT@openvlaoft.se-fractal-vm\endcsname{63.0}
\expandafter\def\csname cfam@LIT@openvlaoft.libero\endcsname{97.1}
\expandafter\def\csname cfam@LIT@openvlaoft.real-repro\endcsname{33.3}
\expandafter\def\csname cfam@LIT@jetson-nano.gflops\endcsname{472}
\expandafter\def\csname cfam@LIT@jetson-nano.power-modes\endcsname{5\,W and 10\,W}
\expandafter\def\csname cfam@LIT@jetson-orin.tops\endcsname{275}
\expandafter\def\csname cfam@LIT@jetson-orin.power-range\endcsname{15--60\,W}

\expandafter\def\csname cfam@CONFIG@eval.seeds-physical\endcsname{3}
\expandafter\def\csname cfam@CONFIG@eval.seeds-sim\endcsname{3}
\expandafter\def\csname cfam@CONFIG@eval.demos-gradient\endcsname{10}
\expandafter\def\csname cfam@CONFIG@eval.demos-cfam\endcsname{1}
\expandafter\def\csname cfam@CONFIG@control.cycle-budget-ms\endcsname{33}
\expandafter\def\csname cfam@CONFIG@substrate.n-locations\endcsname{2^{20}}
\expandafter\def\csname cfam@CONFIG@substrate.address-dim\endcsname{1024}
\expandafter\def\csname cfam@CONFIG@retrieval.key-bits\endcsname{1024}
\expandafter\def\csname cfam@CONFIG@retrieval.query-flips\endcsname{8}
\makeatletter
\@ifundefined{iflatexml}{\newif\iflatexml}{}%
\makeatother
\iflatexml
  \renewcommand{\rowcolor}[2][]{}
  \renewcommand{\rowcolors}[4][]{}%
  \newcolumntype{L}[1]{l}
  \newcolumntype{C}[1]{c}
  \newcolumntype{P}[1]{l}%
\else
  \newcolumntype{L}[1]{>{\raggedright\arraybackslash}p{#1}}%
  \newcolumntype{C}[1]{>{\centering\arraybackslash}p{#1}}%
  \newcolumntype{P}[1]{p{#1}}%
\fi

\title{\LARGE \textbf{Continual Field-Adaptive Models (CFAMs)}\\[0.3em] \Large \textbf{for Post-Deployment Physical AI}\\[0.45em] \large \emph{Autonomous Few-Shot Continual On-Device Adaptation Post-Deployment}}

\author{
 Amarjot Singh$^{1}$ \quad Tanmay R. Pancholi$^{1}$ \quad Jainam Kothari$^{1}$ \\[0.35em]
 Shrirang Mahajan$^{1}$ \quad Ketan Bansal$^{1}$ \quad Zackory Erickson$^{2}$ \\[0.35em]
 Giuseppe Loianno$^{3}$ \quad Alexandre M. Bayen$^{3}$ \quad Jeff Schneider$^{2}$ \quad Vince Nakayama$^{1}$ \\[0.65em]
 \normalsize $^{1}$Skylark Labs \qquad $^{2}$Carnegie Mellon University \qquad $^{3}$University of California, Berkeley
}

\date{}

\begin{document}
\raggedbottom
\twocolumn[
 \begin{@twocolumnfalse}
 \maketitle
 \iflatexml
\begin{abstract}
\noindent\emph{Unattended interactive autonomy}---machines that step into danger in place of humans and complete tasks with the very tools humans use---is the missing capability in mission-critical operations. The domains that need it offer scarce training data and only the compute the asset carries, yet the field keeps presenting novelty that a deployed model cannot learn without erasing what it already knows.

We introduce \textbf{Continual Field-Adaptive Models (CFAMs)}, a model class built for this regime: it learns efficiently in the lab, and then keeps learning after deployment through autonomous, gradient-free, on-device updates. A CFAM is a brain-inspired complementary learning system. Its slow-learning part, frozen after the lab in the role of the slow-learning cortical component of complementary learning systems, is three cortices: a Sensor cortex that lifts multimodal input into 3D-grounded geometry, a Reasoning cortex that decomposes tasks into skills and judges their outcomes, and an Action cortex, a geometric skill model, that executes each skill. Its fast-learning part is the Capsule Field, a hippocampus-like memory where all field learning is written one-shot and gradient-free as Competence Capsules, one capsule per stored competence element. Skill installation is therefore few-shot in the lab on top of the pretrained prior and continual in the field; open-world novelty is outside its scope.

We validate CFAM across five embodiments (manipulator, quadruped, humanoid, quadrotor, and off-road vehicle), with every baseline policy ($\pi_0$, CogACT, SpatialVLA) trained on the same in-house multi-embodiment dataset for the physical-platform comparisons. On the training side, CFAM is data-efficient: it reaches the operating point of the standard policy trained on the full prior-training dataset while using only 40\% of that data ($2.5\times$ fewer prior-training trajectories). At test time, it grows: autonomous capture of verified near-OOD cases raises action success by $13.9$ percentage points while adaptation baselines fall short. And, in the sequential simulation suite, earlier competence is retained: backward transfer is $-0.5$ percentage points, versus $-11.4$ percentage points for LoRA. Together these give a bounded form of post-deployment physical intelligence: learn a task few-shot in the lab, keep growing it autonomously after deployment from verified, slightly out-of-distribution experience, and keep what was gained.
\end{abstract}
\else
\begin{center}
\setlength{\fboxsep}{7pt}%
\setlength{\fboxrule}{0.8pt}%
\fcolorbox{abstractborder}{abstractbg}{%
\begin{minipage}{0.94\textwidth}
{\centering\color{abstracttitle}\bfseries Abstract\par}
\vspace{0.4em}
\small
\setlength{\parindent}{1em}%
\end{minipage}%
}
\end{center}
\fi

 \vspace{1.5em}
 \end{@twocolumnfalse}
]

\begin{figure*}[t]
\centering
\includegraphics[width=0.78\textwidth]{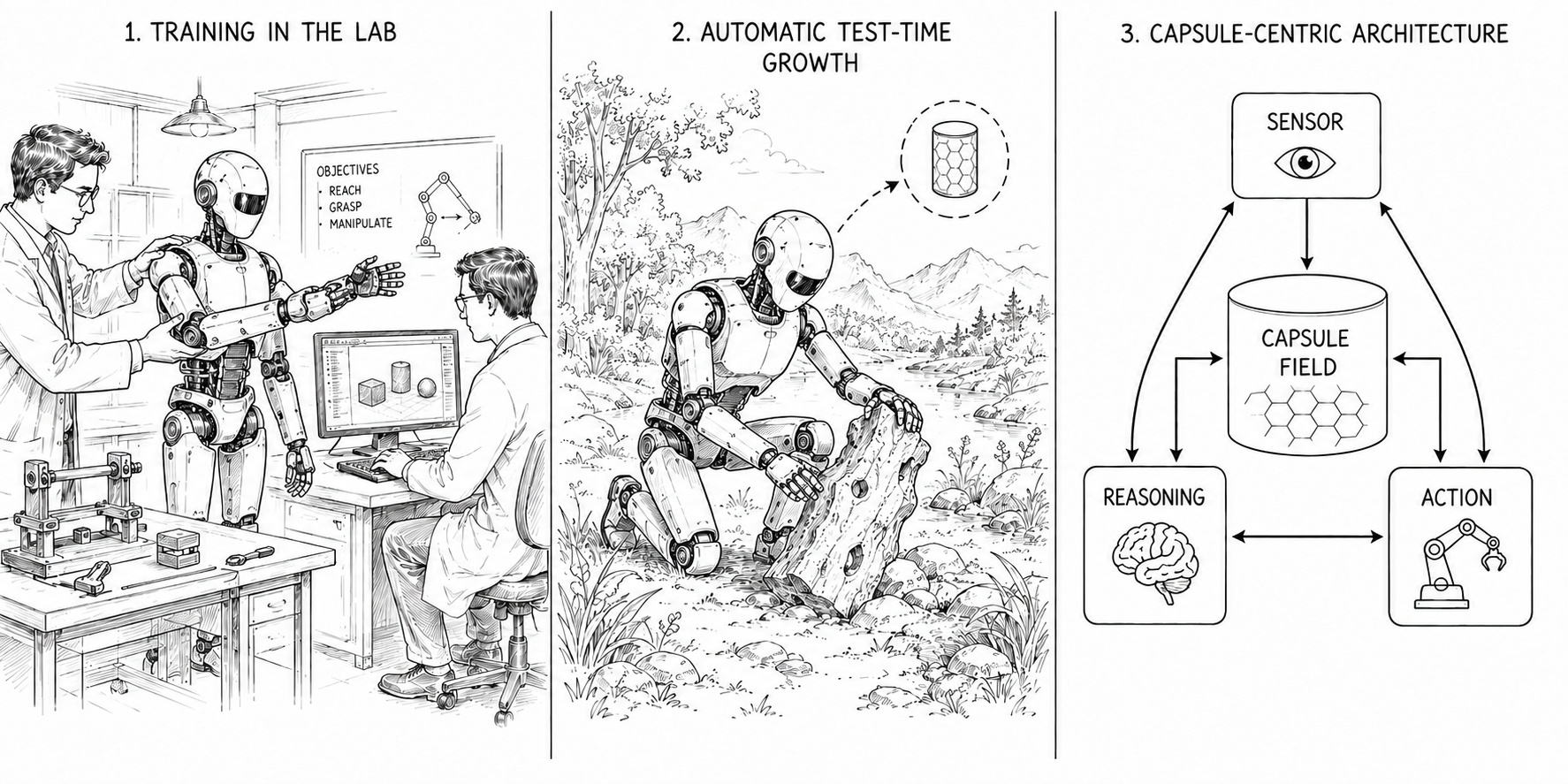}
\caption{\textbf{The Continual Field-Adaptive Model (CFAM) lifecycle and architecture.} \textbf{(1) Few-shot Build:} a handful of operator demonstrations install seed skills as Competence Capsules without retraining the frozen prior. \textbf{(2) Autonomous Test-Time Growth:} after deployment, confidence and retrieval distance identify verified successful near-OOD cases, stored by an on-device, one-shot write with no operator label; each capture expands the support envelope, bounded to known skill families. \textbf{(3) Capsule Field Architecture:} the Sensor, Reasoning, and Action modules retrieve, execute, and consolidate bounded capsules in a fixed-budget field.}
\label{fig:teaser}
\label{fig:motivation}
\end{figure*}

\section{Introduction}
\label{sec:introduction}

Autonomy in defense and other mission-critical domains has so far relied on two systems.
Surveillance systems observe and report, and one-shot strike systems deliver a single pre-committed action.
Neither system interacts with or adapts to the world it operates in.
Yet mines, seaports, battlefields, and public-safety operations are full of tasks that demand exactly this interaction, and today every one of them puts a human in harm's way.
A suspected explosive device must still be cleared by hand, a shaft graded for imminent collapse must still be walked by an inspector, and an unstable structure after an earthquake must still be searched by the people who climb into it.
The missing capability is \emph{unattended interactive autonomy}, machines that step into danger in place of humans and complete their tasks with the very tools humans use.

The same missing capability also keeps many new applications out of reach entirely.
Beyond the tasks that endanger humans lie missions for which no training data exists and none can be collected in advance: the hadal ocean below 6{,}000 meters, the Martian surface with its 4--24 minute communication delay, radiation-saturated reactor interiors.
The hardware, sensors, and control algorithms for these missions exist; the data to train a model for them does not, so a machine can enter such a world only if it can adapt once it is there.

Such systems are difficult to build due to three key challenges:
\begin{enumerate}[label=(\roman*),leftmargin=*,nosep,topsep=2pt]
\item \emph{Data.} Modern robot-learning models acquire their competence from thousands of hours of demonstrations collected through instrumented, internet-scale pipelines~\cite{openvla2024,pi02024}; mission-critical operations offer no such pipeline (their environments are hazardous to instrument, access to them is restricted, and much of what they record is classified), so the demonstrations that exist number in the handfuls and rarely transfer from one mission to the next.
\item \emph{Compute.} Robot-learning models are built for clusters with reliable connectivity, but there is no data center behind a quadruped in a mine shaft or a drone over a disaster site: the model must run on the edge device the asset carries, often disconnected for the whole mission; this is achieved not by distilling a large model into a small one but by designing efficient architectures that capture representations of the problem and fit within that device's compute, memory, and power.
\item \emph{New field cases.} Deployment introduces new object arrangements, loads, terrain conditions, and combinations from the first day in the field. These cases are slightly out of distribution but still close to known skills; a frozen model degrades against them, laboratory retraining is too slow or unavailable, and naive in-place updates can erase earlier knowledge through catastrophic forgetting. Unrelated open-world tasks are outside this paper's scope.
\end{enumerate}
A model class for this setting must therefore learn from a few examples, run on hardware a machine can carry, and keep learning without forgetting; \Cref{sec:diagnosis} formalizes these conditions as the \emph{field learning regime}, and this paper introduces an architecture built for it.

We present \textbf{Continual Field-Adaptive Models (CFAMs)}, a new architecture inspired by the brain, built to give mission-critical Physical AI what it is missing: a model that learns tasks few-shot and then autonomously captures verified, slightly out-of-distribution experience on-device after it leaves the lab (\Cref{fig:teaser}).
The unit of this learning is the \textbf{Competence Capsule (CC)}: a compact, callable record of one skill that jointly captures the perception and action information bound to the situation that activates it, and from which the system generalizes---one capsule, learned from as little as one example, is warped geometrically onto new scenes rather than retrained.
The design mirrors complementary learning in the brain~\cite{mcclelland1995complementary}: frozen substrates provide slow, stable competence in the role of the slowly learning \emph{neocortex}, while a new-learning layer rapidly encodes new episodes like the \emph{hippocampus} and consolidates them over time.

Concretely, a CFAM is a custom VLA stack with two learning phases.
\emph{In the lab}, it learns efficiently: a \textbf{Sensor module}~\cite{singh2017shdl,singh2019thesis,singh2023shdlpatent} lifts multi-modal inputs into 3D-grounded geometry, a \textbf{Reasoning module} (a vision-language model custom-trained for mission-critical operations) decomposes tasks into skills and judges their outcomes, and an \textbf{Action module} executes each skill as one geometrically warped emission onto the current 3D scene via the Geometric Residual Transform (GRT), so a handful of demonstrations per task, not thousands, install the skill library on top of the pretrained prior.
\emph{In the field}, it keeps learning: the \textbf{new-learning layer}, governed by the Continual Field Update Rule ($\CFUR$, \Cref{sec:growth_exec}), writes new competence on the edge device, autonomously, with no operator label and no gradient step; each write is one forward pass and one memory insertion, designed to fit the control cycle.
The capturable events are \textbf{near-OOD cases} (or \textbf{near-edge novelty}): situations slightly out of distribution but close enough that a stored skill still warps into a verified success. Each capture becomes a new capsule, one-shot, within a fixed memory budget and with near-zero interference with what is already stored, so the envelope the library covers widens with use without overwriting what is already stored; no existing adaptation family (gradient-based, reinforcement learning, memory-augmented, or test-time adaptation) offers this combination (\Cref{sec:value_iteration_fails}), and the design carries formal deployment guarantees (locality, bounded authority, and convergence), each proved under its stated assumptions in the supplementary material.
The Action module and the near-edge extension path of the new-learning layer are the novel contributions of this work, and both are evaluated here; correction from detected field failures and open-world novelty are outside this paper's scope (\Cref{subsec:limitations}), and the Sensor and Reasoning modules are described as designed and integrated, with isolated evaluation in \Cref{tab:reasoning_pathway,tab:perception_pathway}.

\paragraph{Contributions.}
This paper makes four contributions:
\begin{enumerate}[leftmargin=*,itemsep=2pt,topsep=4pt]
 \item \textbf{Architecture}: CFAMs, an end-to-end field-adaptive model class for Physical AI, four pieces (Sensor / Reasoning / Action / new-learning layer) built for settings where training data is limited in the first place. It includes the Action module's \emph{unit-of-inference} shift (per-skill capsule emissions via GRT, $M \ll T$ action-decoding calls per task) and the new-learning layer's architectural interface (the capsule schema every field write must produce, and the guarantees imposed on any writer, proved in the supplementary material).

 \item \textbf{Mission-critical dataset}: Skylark's in-house multi-embodiment dataset (2.6 million$+$ trajectories across five physical platforms) on which CFAM and every standard-policy baseline ($\pi_0$, CogACT, SpatialVLA) are trained for the physical-platform comparisons, so those comparisons are matched-data (the simulation benchmarks use the public backbones).

 \item \textbf{Efficient lab learning}: on the learning curve, CFAM matches the standard policy trained on 100\% of $D_\mathrm{train}$ using only 40\% of it, and leads every matched-data baseline on the held-out split (\Cref{tab:action_pathway}).

 \item \textbf{Autonomous post-deployment growth with near-zero measured forgetting}: on the variation stream $D_\mathrm{var}$, autonomous test-time capture of new cases raises action success by $13.9$\,pp ($74.0 \to 87.9$, \Cref{tab:real_growth_family}) while adaptation baselines fall short (\Cref{tab:vla_adaptation}); in the sequential simulation suite retention stays near-perfect, backward transfer $-0.5$\,pp vs.\ $-11.4$\,pp for LoRA~\cite{hu2022lora} (\Cref{tab:continual}).

\end{enumerate}

\paragraph{Paper organization.}
\Cref{sec:diagnosis} first derives the field-learning requirements from deployment conditions; \Cref{sec:related_work} then evaluates prior adaptation families against those requirements and isolates the remaining gap.
\Cref{sec:theory} states CFAM's architectural response; \Cref{sec:system_overview} presents the four-piece architecture, \Cref{sec:capsule} defines the Competence Capsule, and \Cref{sec:skill_composition} covers how skills are selected, warped, and executed.
\Cref{sec:execution} develops test-time growth, the $\CFUR$ capture rule, and compression.
\Cref{sec:experiments} reports the pre-novelty experiments, \Cref{sec:discussion} discusses limitations, and \Cref{sec:conclusion} concludes.
\section{The Field Learning Regime: Why a Deployed Model Must Keep Growing}
\label{sec:diagnosis}

\begin{figure*}[t]
\centering
\iflatexml
\includegraphics[width=0.7\textwidth]{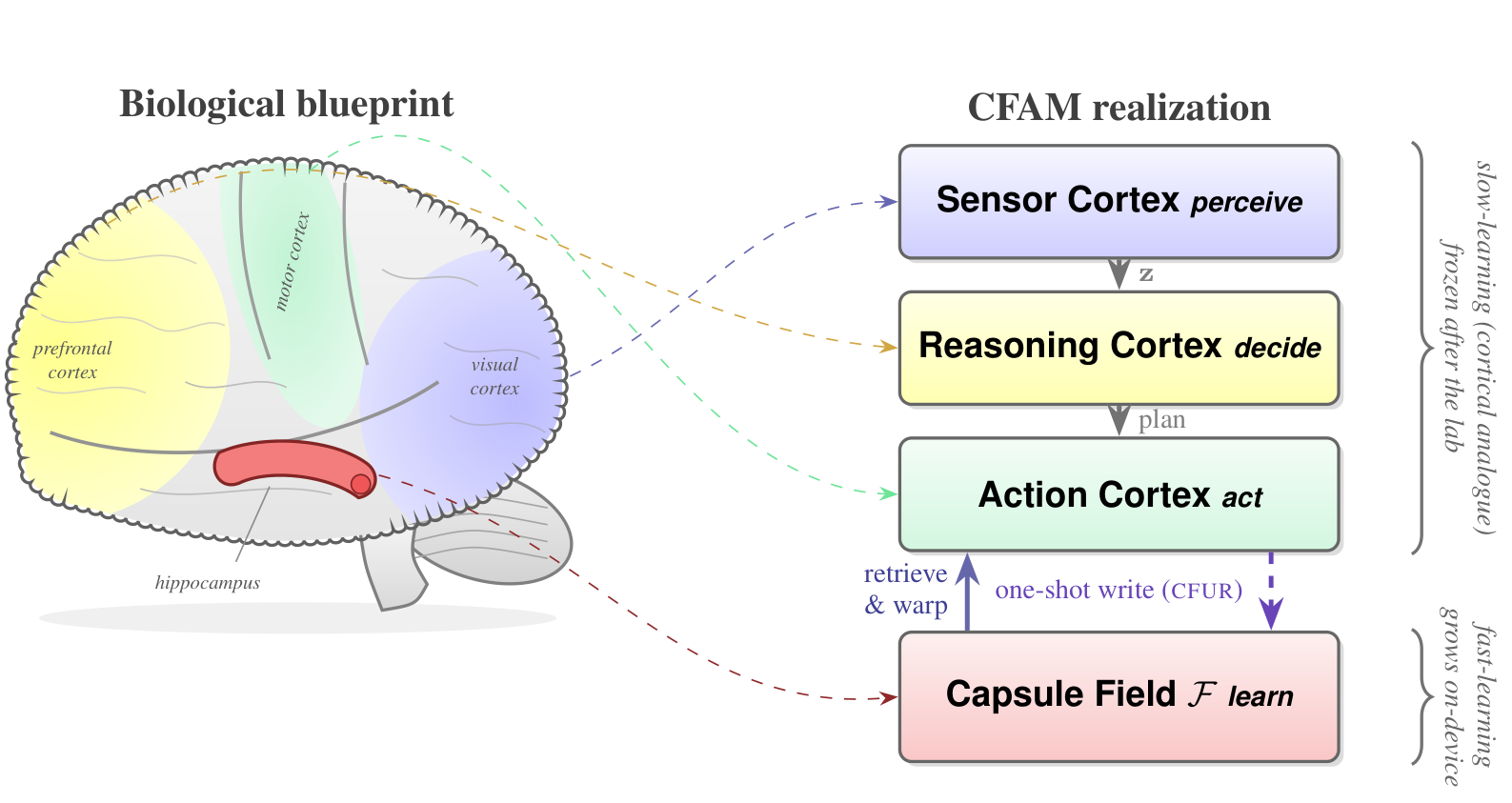}
\else
\resizebox{0.7\textwidth}{!}{%
\begin{tikzpicture}[
 >=Stealth,
 every node/.style={font=\small},
 cortexbox/.style={draw=black!60, rounded corners=3pt, thick, drop shadow={shadow xshift=0.045cm, shadow yshift=-0.045cm, fill=black, opacity=0.14}, minimum width=3.9cm, minimum height=1.0cm, align=center, font=\small\sffamily\bfseries},
 arr/.style={->, very thick, black!55},
 maparr/.style={->, dashed, thin},
 ]

 \node[font=\normalsize\bfseries, text=black!75] at (2.9,5.0) {Biological blueprint};
 \node[font=\normalsize\bfseries, text=black!75] at (10.3,5.0) {CFAM realization};

 \fill[black!14, opacity=0.4] (3.0,0.45) ellipse (2.3 and 0.14);

 \fill[top color=black!5, bottom color=black!17, draw=black!50, thick, rotate around={-18:(4.62,1.28)}] (4.62,1.28) ellipse (0.84 and 0.48);
 \foreach \yy in {0.98,1.13,1.28,1.43}
 \draw[black!35, thin] (4.02,\yy) .. controls (4.55,\yy+0.13) .. (5.22,\yy);

 \fill[top color=black!4, bottom color=black!15, draw=black!50, thick] (3.55,1.25) .. controls (3.6,0.9) .. (3.75,0.52) .. controls (4.05,0.57) .. (4.15,0.75) .. controls (4.0,1.0) .. (4.05,1.3) -- cycle;

 \begin{scope}
 \clip plot [smooth cycle, tension=0.75] coordinates {(0.55,2.0) (0.5,2.9) (1.1,3.85) (2.4,4.45) (3.8,4.3) (4.95,3.55) (5.35,2.5) (4.95,1.75) (4.2,1.35) (3.2,1.15) (2.2,1.2) (1.25,1.4) (0.7,1.6)};
 \shade[top color=black!3, bottom color=black!12] (-0.5,0) rectangle (6.0,5.0);
 \shade[inner color=yellow!44, outer color=yellow!20] (0.85,2.85) circle (1.55); 
 \shade[inner color=blue!26, outer color=blue!12] (4.95,2.35) circle (1.4); 
 \shade[inner color=augcolor!34, outer color=augcolor!15, rotate around={14:(2.95,3.6)}] (2.95,3.6) ellipse (0.55 and 1.5); 
 \draw[black!38, line width=0.8pt] (2.5,4.42) .. controls (2.45,3.7) and (2.6,3.2) .. (2.75,2.75);
 \draw[black!38, line width=0.8pt] (3.42,4.32) .. controls (3.38,3.6) and (3.5,3.15) .. (3.62,2.7);
 \draw[black!42, line width=0.9pt] (0.8,2.1) .. controls (1.5,1.85) and (2.4,1.88) .. (3.15,2.08) .. controls (3.6,2.2) and (3.95,2.35) .. (4.25,2.55);
 \draw[yellow!45!black!32, thin] (0.72,3.15) .. controls (1.05,2.95) and (1.35,3.25) .. (1.7,3.05) .. controls (1.95,2.9) and (2.15,3.15) .. (2.35,3.05);
 \draw[yellow!45!black!32, thin] (0.68,2.5) .. controls (1.0,2.35) and (1.3,2.6) .. (1.65,2.45);
 \draw[yellow!45!black!32, thin] (1.05,3.7) .. controls (1.4,3.5) and (1.7,3.8) .. (2.1,3.6);
 \draw[black!26, thin] (3.75,3.55) .. controls (4.05,3.35) and (4.3,3.6) .. (4.6,3.4);
 \draw[blue!40!black!30, thin] (4.3,2.9) .. controls (4.6,2.7) and (4.85,2.95) .. (5.15,2.75);
 \draw[blue!40!black!30, thin] (4.35,2.25) .. controls (4.65,2.05) and (4.9,2.3) .. (5.2,2.1);
 \draw[black!26, thin] (1.6,1.6) .. controls (2.1,1.45) and (2.7,1.7) .. (3.3,1.55);
 \draw[black!26, thin] (2.0,2.55) .. controls (2.3,2.4) and (2.6,2.65) .. (2.9,2.5);
 \end{scope}
 \draw[black!62, thick, decorate, decoration={bumps, amplitude=1.2pt, segment length=6.5pt}] plot [smooth cycle, tension=0.75] coordinates {(0.55,2.0) (0.5,2.9) (1.1,3.85) (2.4,4.45) (3.8,4.3) (4.95,3.55) (5.35,2.5) (4.95,1.75) (4.2,1.35) (3.2,1.15) (2.2,1.2) (1.25,1.4) (0.7,1.6)};

 \fill[corrcolor!70, draw=corrcolor!55!black, thick]
 (2.3,1.92) .. controls (2.65,2.08) and (3.3,2.06) .. (3.62,1.82)
 .. controls (3.74,1.72) and (3.7,1.55) .. (3.54,1.52)
 .. controls (3.28,1.72) and (2.75,1.78) .. (2.48,1.65)
 .. controls (2.3,1.58) and (2.2,1.78) .. (2.3,1.92) -- cycle;
 \fill[corrcolor!90, draw=corrcolor!55!black, thin] (3.56,1.64) circle (0.085);

 \node[font=\tiny\itshape, text=black!65, align=center] at (1.0,2.75) {prefrontal\\cortex};
 \node[font=\tiny\itshape, text=black!65, rotate=76] at (2.95,3.62) {motor cortex};
 \node[font=\tiny\itshape, text=black!65, align=center] at (4.75,2.6) {visual\\cortex};
 \node[font=\tiny\itshape, text=black!65] at (2.2,0.75) {hippocampus};
 \draw[black!45, thin] (2.75,1.62) -- (2.45,0.95);

 \node[cortexbox, top color=blue!3, bottom color=blue!19] (sensor) at (10.3,4.15)
 {Sensor Cortex\;{\scriptsize\itshape perceive}};
 \node[cortexbox, top color=yellow!10, bottom color=yellow!30] (reason) at (10.3,2.85)
 {Reasoning Cortex\;{\scriptsize\itshape decide}};
 \node[cortexbox, top color=augcolor!6, bottom color=augcolor!24] (action) at (10.3,1.55)
 {Action Cortex\;{\scriptsize\itshape act}};
 \node[cortexbox, top color=corrcolor!8, bottom color=corrcolor!30, minimum height=1.15cm] (mem) at (10.3,-0.25)
 {Capsule Field $\rcufield$\;{\scriptsize\itshape learn}};

 \draw[arr] (sensor) -- (reason) node[midway, right=1pt, font=\scriptsize, text=black!50]{$\vz$};
 \draw[arr] (reason) -- (action) node[midway, right=1pt, font=\scriptsize, text=black!50]{plan};
 \draw[->, very thick, blue!45!black!60] ($(mem.north)+(-1.35,0)$) -- ($(action.south)+(-1.35,0)$)
 node[midway, left=1pt, font=\scriptsize, text=blue!45!black!75, align=right]{retrieve\\\& warp};
 \draw[->, very thick, novcolor!75!black, dashed] ($(action.south)+(1.35,0)$) -- ($(mem.north)+(1.35,0)$);
 \node[font=\scriptsize, text=novcolor!75!black] at (10.3,0.69) {one-shot write ($\CFUR$)};

 \draw[decorate, decoration={brace, amplitude=5pt}, black!55, thick] (12.9,4.68) -- (12.9,1.02);
 \node[font=\scriptsize\itshape, text=black!60, align=center, rotate=-90] at (13.4,2.85) {slow-learning (cortical analogue)\\frozen after the lab};
 \draw[decorate, decoration={brace, amplitude=5pt}, black!55, thick] (12.9,0.35) -- (12.9,-0.85);
 \node[font=\scriptsize\itshape, text=black!60, align=center, rotate=-90] at (13.4,-0.25) {fast-learning\\grows on-device};

 \draw[maparr, blue!50!black!60] (5.42,2.6) to[out=25,in=180] (sensor.west);
 \draw[maparr, yellow!45!orange!80!black] (1.3,3.95) to[out=32,in=175] (reason.west);
 \draw[maparr, augcolor!80] (3.1,4.42) to[out=40,in=178] (action.west);
 \draw[maparr, corrcolor!60!black] (3.72,1.72) to[out=-15,in=185] (mem.west);

\end{tikzpicture}%
}
\fi
\caption{\textbf{The brain's learning architecture, transcribed.} \emph{Left:} the slow-learning neocortical structures (visual: perceive; prefrontal: decide; motor: act) and the fast-learning hippocampus, which encodes new episodes in one exposure. \emph{Right:} CFAM realizes each structure (dashed mappings): the Sensor, Reasoning, and Action cortices are trained in the lab and frozen at deployment, and the \textbf{Capsule Field} $\rcufield$ is the hippocampal memory, written one-shot and gradient-free by the new-learning layer ($\CFUR$), one Competence Capsule per stored competence element. In CFAM the division is strict in both directions: the slow part is frozen after the lab (a deliberate idealization of the cortical side, which in the biological account learns slowly rather than not at all), and the fast part never acts on the world directly.}
\label{fig:two_part}
\end{figure*}
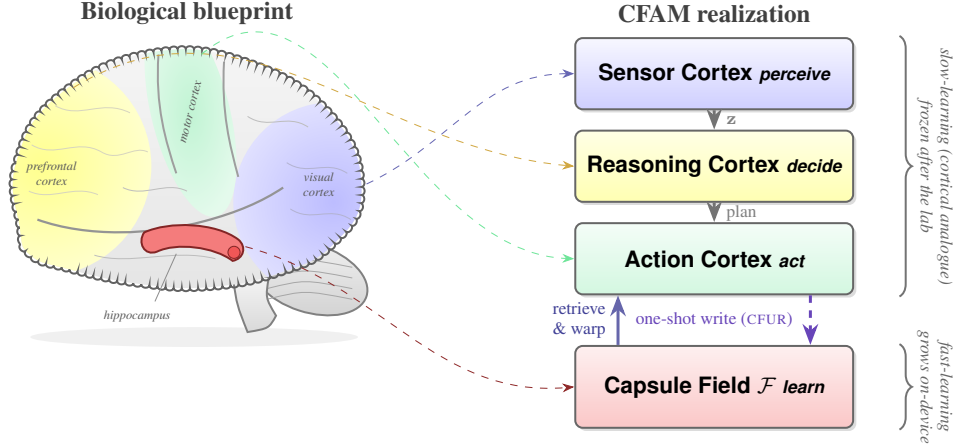

The world a mission presents cannot be captured in any pre-deployment dataset, and in mission-critical domains the pre-deployment data is limited in the first place (\Cref{sec:introduction}), so the system must learn the world after deployment.
But ``after deployment'' is not the lab: no curator selects training examples, no class distribution is balanced, and the machine is alone with whatever compute it carried in and whatever data the world decides to show it.
What the world shows it is a \emph{stream}: events arrive one at a time, drift steadily further from the training distribution as the mission ages, and occasionally present something genuinely new, the shape the evaluation protocol of \Cref{sec:exp:setup} makes measurable.

We call these conditions the \textbf{field learning regime}, and three properties define it.
Connectivity is absent or intermittent.
Supervision is episodic rather than continuous, arriving as a human demonstration here or a sensor spike there.
And the operational tempo requires that any on-device write complete within the controller's cycle time, without interrupting or catastrophically delaying the task the system was deployed to perform.

\paragraph{Field-learning requirements.}
\label{sec:requirements}
Any architecture that meets this regime by \emph{growing} rather than retraining must satisfy four requirements:
\begin{itemize}[leftmargin=*,itemsep=2pt,topsep=3pt]
 \item \textbf{R1 (Efficient learning)}\label{req:detection}: the system must build a usable competence library from a handful of demonstrations, not the thousands a from-scratch policy needs.
 \item \textbf{R2 (Typed growth)}\label{req:typed_adapt}: the system must extend competence through distinct mechanisms matched to the gap: \emph{correction} of a familiar mistake and \emph{extension} to a within-family instance, without retraining the base.
 \item \textbf{R3 (Bounded consolidation)}\label{req:consolidation}: the system must merge redundant competence and hold a fixed memory footprint as it grows, so new writes never corrupt the skills it already has.
 \item \textbf{R4 (Field-feasible growth)}\label{req:tempo}: each new capsule must be written in one shot, gradient-free and on-device, within fixed compute and memory and the control-cycle budget.
\end{itemize}
The current model class satisfies none of them: it is data-hungry at build time (R1) and cannot grow after deployment (R2--R4); \Cref{sec:related_work} provides the cited, family-by-family assessment.
The core memory unit (\Cref{sec:capsule}) and the architecture built around it (\Cref{sec:system_overview}) are designed to fill both gaps.

\begin{figure*}[t]
\centering
\iflatexml
\includegraphics[width=0.78\textwidth]{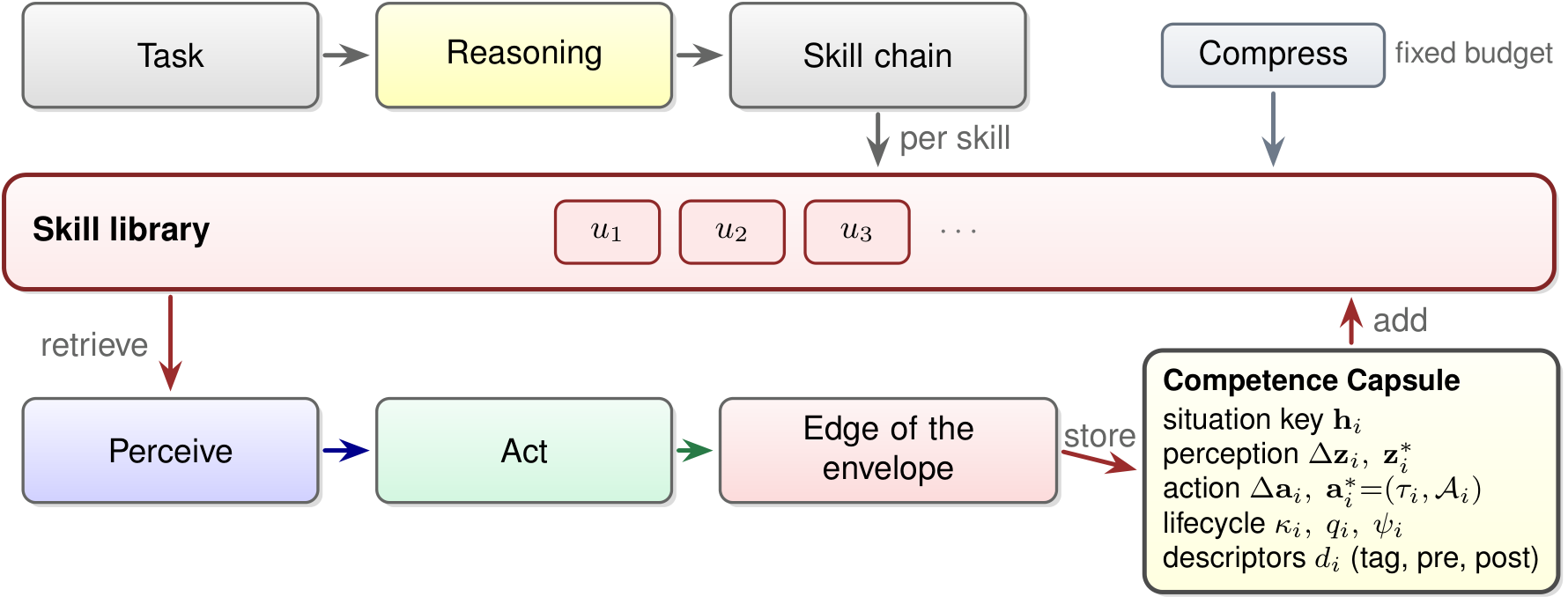}
\else
\resizebox{0.78\textwidth}{!}{%
\begin{tikzpicture}[
 >=Stealth,
 every node/.style={font=\small},
 box/.style={draw=black!60, rounded corners=3.5pt, drop shadow={shadow xshift=0.045cm, shadow yshift=-0.045cm, fill=black, opacity=0.14}, thick, font=\small\sffamily, align=center, minimum height=1.0cm, text width=2.6cm},
 chip/.style={draw=corrcolor!60!black, rounded corners=3pt, thick, fill=corrcolor!12, font=\small\sffamily, align=center, minimum height=0.6cm, minimum width=1.0cm},
 arr/.style={->, very thick, shorten >=1.5pt, shorten <=1.5pt, line cap=round},
 lbl/.style={font=\small\sffamily, text=black!60},
 ]
 \node[box, top color=black!2, bottom color=black!13] (task) at (0,5.15) {Task};
 \node[box, top color=yellow!9, bottom color=yellow!27] (reason) at (3.4,5.15) {Reasoning};
 \node[box, top color=black!2, bottom color=black!13] (chain) at (6.8,5.15) {Skill chain};
 \draw[arr, black!60] (task) -- (reason);
 \draw[arr, black!60] (reason) -- (chain);

 \draw[rounded corners=6pt, very thick, draw=corrcolor!55!black, top color=corrcolor!3, bottom color=corrcolor!11,
 drop shadow={shadow xshift=0.045cm, shadow yshift=-0.045cm, fill=black, opacity=0.14}]
 (-1.6,2.9) rectangle (13.3,4.0);
 \node[font=\small\sffamily\bfseries, anchor=west] at (-1.45,3.45) {Skill library};
 \node[chip] at (4.2,3.45) {$\rcu_1$};
 \node[chip] at (5.4,3.45) {$\rcu_2$};
 \node[chip] at (6.6,3.45) {$\rcu_3$};
 \node[font=\small\sffamily, text=black!55] at (7.6,3.45) {$\cdots$};
 \node[draw=passcolor!70!black, rounded corners=3pt, thick, top color=passcolor!10, bottom color=passcolor!26,
 font=\small\sffamily, align=center, minimum height=0.6cm, text width=1.9cm] (compress) at (10.6,5.15) {Compress};
 \node[lbl, anchor=west, font=\footnotesize\sffamily] at (11.65,5.15) {fixed budget};
 \draw[arr, passcolor!75!black] (compress.south) -- (10.6,4.02);

 \node[box, top color=blue!3, bottom color=blue!18] (perc) at (0,1.35) {Perceive};
 \node[box, top color=augcolor!7, bottom color=augcolor!24] (act) at (3.4,1.35) {Act};
 \node[box, top color=corrcolor!5, bottom color=corrcolor!19, text width=3.0cm] (edge) at (6.9,1.35) {Edge of the envelope};
 \draw[arr, blue!55!black] (perc) -- (act);
 \draw[arr, augcolor!55!black] (act) -- (edge);

 \node[draw=black!70, very thick, rounded corners=5pt, top color=yellow!3, bottom color=yellow!12,
 drop shadow={shadow xshift=0.045cm, shadow yshift=-0.045cm, fill=black, opacity=0.14},
 font=\footnotesize\sffamily, align=left, inner sep=5pt] (cc) at (11.35,1.15)
 {\textbf{Competence Capsule}\\[1pt]
 situation key $\vh_i$\\ perception $\Delta\vz_i,\ \vz^{*}_i$\\
 action $\resid_i,\ \astaraction_i{=}(\tau_i,\mathcal{A}_i)$\\ lifecycle $\sal_i,\ \trust_i,\ \phase_i$\\ descriptors $d_i$ (tag, pre, post)};

 \draw[arr, black!60] (chain.south) -- node[lbl, right=2pt]{per skill} (6.8,4.02);
 \draw[arr, corrcolor!65!black] (0,2.88) -- node[lbl, left=2pt]{retrieve} (perc.north);
 \draw[arr, corrcolor!65!black] (edge.east) -- node[lbl, above]{store} (cc.west);
 \draw[arr, corrcolor!65!black] (cc.north) -- node[lbl, right=2pt]{add} (11.35,2.88);
\end{tikzpicture}%
}
\fi
\caption{\textbf{How a CFAM runs a task, and how its library grows.} The Reasoning cortex decomposes the task into a chain of skills; for each skill the system perceives the scene in 3D, retrieves the matching capsule, and warps it onto that scene to act---no gradient step, no per-step decoding. An execution that succeeds at the edge of a skill's envelope is stored as a new \textbf{Competence Capsule} (\Cref{eq:rcu_full}), so the library grows with use, while decay and consolidation (\emph{Compress}) hold it within a fixed memory budget. Colors follow \Cref{fig:two_part}.}
\label{fig:cc_schema}
\label{fig:cc_anatomy}
\end{figure*}

\begin{figure*}[t]
\centering
\iflatexml
\includegraphics[width=0.78\textwidth]{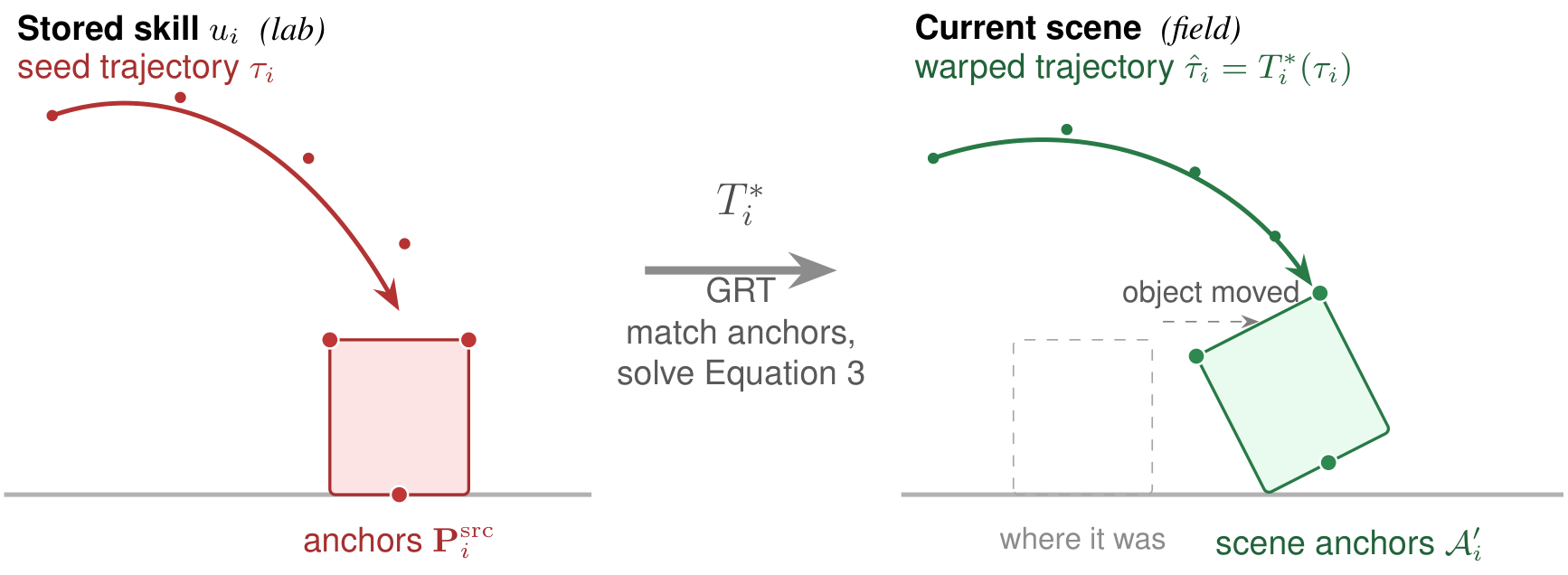}
\else
\resizebox{0.78\textwidth}{!}{%
\begin{tikzpicture}[
 >=Stealth,
 every node/.style={font=\small},
 src/.style={circle, fill=corrcolor!80!black, draw=white, semithick, inner sep=0pt, minimum size=4.6pt},
 dst/.style={circle, fill=augcolor!62!black, draw=white, semithick, inner sep=0pt, minimum size=4.6pt},
 lbl/.style={font=\small\sffamily, text=black!65},
 ttl/.style={font=\small\sffamily\bfseries},
 ]
 \draw[black!30, very thick] (-0.3,0) -- (5.2,0);
 \node[ttl, anchor=west] at (-0.3,4.35) {Stored skill $\rcu_i$ \ {\normalfont\itshape (lab)}};
 \draw[fill=corrcolor!14, draw=corrcolor!70!black, thick, rounded corners=1.5pt] (2.75,0) rectangle (4.05,1.45);
 \node[src] (a1) at (2.75,1.45) {};
 \node[src] (a2) at (4.05,1.45) {};
 \node[src] (a3) at (3.4,0) {};
 \node[lbl, anchor=north, text=corrcolor!70!black] at (3.4,-0.18) {anchors $\mathbf{P}^{\mathrm{src}}_i$};
 \draw[very thick, corrcolor!75!black, ->] (0.15,3.55) .. controls (1.3,3.95) and (2.5,3.3) .. (3.4,1.72);
 \foreach \x/\y in {0.15/3.55, 1.35/3.72, 2.55/3.15, 3.45/2.35}
 \node[circle, fill=corrcolor!75!black, inner sep=0pt, minimum size=3pt] at (\x,\y) {};
 \node[lbl, anchor=west, text=corrcolor!70!black] at (-0.3,3.98) {seed trajectory $\tau_i$};

 \draw[->, line width=2.2pt, black!45] (5.7,2.1) -- (7.5,2.1);
 \node[font=\large\sffamily\bfseries, text=black!70] at (6.6,2.72) {$T^{*}_i$};
 \node[lbl, align=center] at (6.6,1.5) {GRT\\match anchors,\\solve \Cref{eq:grt_transform}};

 \draw[black!30, very thick] (8.1,0) -- (14.3,0);
 \node[ttl, anchor=west] at (8.1,4.35) {Current scene \ {\normalfont\itshape (field)}};
 \begin{scope}[shift={(12.1,0.30)}, rotate=27]
 \draw[fill=augcolor!12, draw=augcolor!55!black, thick, rounded corners=1.5pt] (-0.65,0) rectangle (0.65,1.45);
 \node[dst] (b1) at (-0.65,1.45) {};
 \node[dst] (b2) at (0.65,1.45) {};
 \node[dst] (b3) at (0,0) {};
 \end{scope}
 \node[lbl, anchor=north, text=augcolor!45!black, align=center] at (12.3,-0.18) {scene anchors $\mathcal{A}'_i$};
 \draw[very thick, augcolor!55!black, ->] (8.4,3.15) .. controls (9.7,3.6) and (11.1,3.15) .. (11.95,1.95);
 \foreach \x/\y in {8.4/3.15, 9.65/3.42, 10.85/3.02, 11.6/2.42}
 \node[circle, fill=augcolor!55!black, inner sep=0pt, minimum size=3pt] at (\x,\y) {};
 \node[lbl, anchor=west, text=augcolor!45!black] at (8.1,3.98) {warped trajectory $\hat\tau_i = T^{*}_i(\tau_i)$};
 \draw[dashed, black!35, rounded corners=1.5pt] (9.15,0) rectangle (10.45,1.45);
 \node[lbl, anchor=north, text=black!45, font=\footnotesize\sffamily] at (9.8,-0.18) {where it was};
 \draw[->, dashed, black!50] (10.55,1.62) -- node[lbl, above, font=\footnotesize\sffamily]{object moved} (11.45,1.62);
\end{tikzpicture}%
}
\fi
\caption{\textbf{The Geometric Residual Transform: one stored trajectory, any object pose.} A capsule stores a seed trajectory $\tau_i$ with the 3D anchors of the object it was demonstrated on \emph{(left)}; in the field the object is translated and rotated \emph{(right)}. GRT matches stored to current anchors under the task-weighted distance $D_w$, selects the admissible transform $T^{*}_i$, and carries the whole trajectory through it in one pass: the skill is re-aimed, not re-learned. The phase objective $\Phi_i$ verifies the emission; a failure receives a bounded local residual restricted to the failing phase. \Cref{fig:grt_real} shows the operator on real hardware.}
\label{fig:grt_mechanism}
\end{figure*}
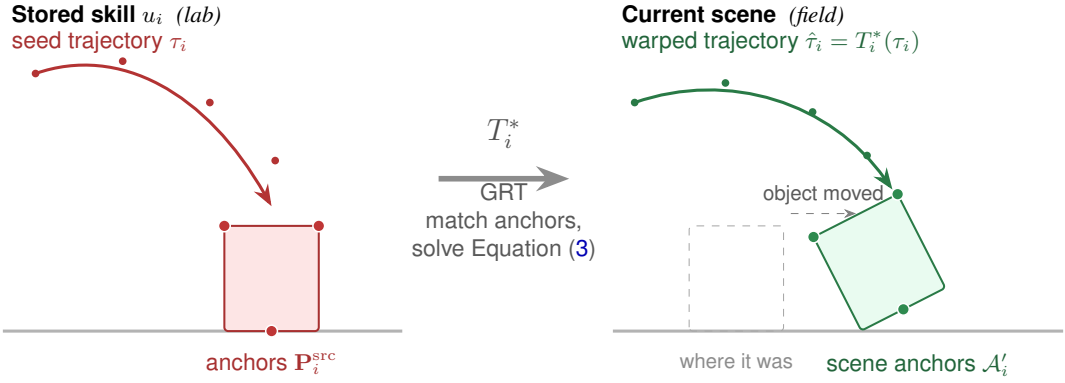

\begin{figure*}[t]
\centering
\iflatexml
\includegraphics[width=0.78\textwidth]{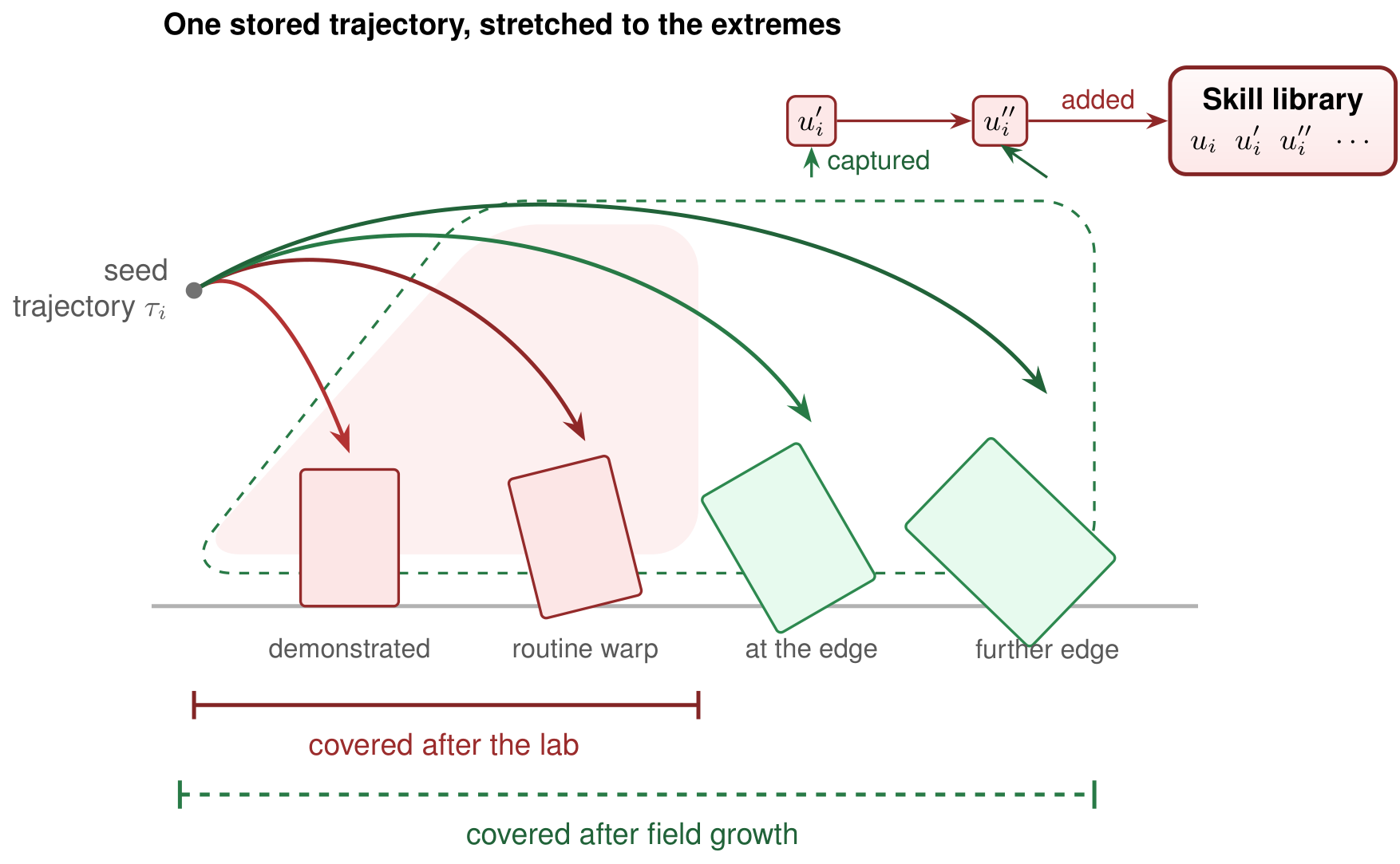}
\else
\resizebox{0.78\textwidth}{!}{%
\begin{tikzpicture}[
 >=Stealth,
 every node/.style={font=\small},
 lbl/.style={font=\small\sffamily, text=black!65},
 ttl/.style={font=\small\sffamily\bfseries},
 chip/.style={draw=corrcolor!60!black, rounded corners=2.5pt, thick, fill=corrcolor!12, font=\small\sffamily, inner sep=3pt},
 ]
 \fill[corrcolor!8, rounded corners=14pt] (1.35,0.55) -- (4.5,4.05) -- (6.7,4.05) -- (6.7,0.55) -- cycle;
 \draw[dashed, augcolor!55!black, thick, rounded corners=16pt] (1.2,0.35) -- (4.3,4.3) -- (10.9,4.3) -- (10.9,0.35) -- cycle;
 \draw[corrcolor!55!black, very thick] (1.35,-1.05) -- (6.7,-1.05);
 \draw[corrcolor!55!black, very thick] (1.35,-0.9) -- (1.35,-1.2);
 \draw[corrcolor!55!black, very thick] (6.7,-0.9) -- (6.7,-1.2);
 \node[lbl, anchor=north, text=corrcolor!65!black] at (4.0,-1.22) {covered after the lab};
 \draw[augcolor!55!black, very thick, dashed] (1.2,-2.0) -- (10.9,-2.0);
 \draw[augcolor!55!black, very thick] (1.2,-1.85) -- (1.2,-2.15);
 \draw[augcolor!55!black, very thick] (10.9,-1.85) -- (10.9,-2.15);
 \node[lbl, anchor=north, text=augcolor!45!black] at (6.0,-2.17) {covered after field growth};

 \draw[black!30, very thick] (0.9,0) -- (12.0,0);
 \node[ttl, anchor=west] at (0.9,6.15) {One stored trajectory, stretched to the extremes};

 \foreach \x/\rot/\sc/\col/\lab in {%
 3.0/0/1.0/corrcolor/{demonstrated},
 5.5/14/1.05/corrcolor/{routine warp},
 7.9/30/1.15/augcolor/{at the edge},
 10.4/46/1.28/augcolor/{further edge}} {
 \begin{scope}[shift={(\x,0.30)}, rotate=\rot, scale=\sc]
 \draw[fill=\col!13, draw=\col!62!black, thick, rounded corners=1.5pt] (-0.52,-0.30) rectangle (0.52,1.15);
 \end{scope}
 \node[lbl, anchor=north, font=\footnotesize\sffamily] at (\x,-0.22) {\lab};
 }

 \draw[very thick, corrcolor!75!black, ->] (1.35,3.35) .. controls (1.9,3.75) and (2.5,2.9) .. (3.0,1.62);
 \draw[very thick, corrcolor!60!black, ->] (1.35,3.35) .. controls (2.7,4.15) and (4.6,3.4) .. (5.5,1.75);
 \draw[very thick, augcolor!55!black, ->] (1.35,3.35) .. controls (3.4,4.55) and (6.6,3.8) .. (7.9,1.95);
 \draw[very thick, augcolor!45!black, ->] (1.35,3.35) .. controls (4.0,5.0) and (8.8,4.25) .. (10.4,2.25);
 \node[circle, fill=black!55, inner sep=0pt, minimum size=5pt] at (1.35,3.35) {};
 \node[lbl, anchor=east, align=right] at (1.2,3.35) {seed\\trajectory $\tau_i$};

 \node[chip] (n1) at (7.9,5.15) {$\rcu_i'$};
 \node[chip] (n2) at (9.9,5.15) {$\rcu_i''$};
 \draw[->, thick, augcolor!55!black] (7.9,4.55) -- node[lbl, right=1pt, font=\footnotesize\sffamily, text=augcolor!45!black]{captured} (n1.south);
 \draw[->, thick, augcolor!45!black] (10.4,4.55) -- (9.9,4.9);
 \node[draw=corrcolor!55!black, very thick, rounded corners=5pt,
 top color=corrcolor!3, bottom color=corrcolor!12, inner sep=6pt,
 font=\small\sffamily, align=center] (lib) at (12.9,5.15)
 {\textbf{Skill library}\\[1pt]$\rcu_i\ \ \rcu_i'\ \ \rcu_i''\ \ \cdots$};
 \draw[->, thick, corrcolor!60!black] (n1.east) -- (n2.west);
 \draw[->, thick, corrcolor!60!black] (n2.east) -- node[lbl, above, font=\footnotesize\sffamily, text=corrcolor!65!black]{added} (lib.west);
\end{tikzpicture}%
}
\fi
\caption{\textbf{Generalization beyond the extremes: the envelope grows with use.} GRT re-aims one stored seed trajectory at whatever pose the object takes, so a single skill covers a band of situations around the demonstration \emph{(pink)}. Each success at the edge of the envelope is captured as a new Competence Capsule ($\rcu_i'$, $\rcu_i''$), so the covered region expands with use \emph{(dashed)}, with no gradient step and no failure oracle. \Cref{sec:growth_exec} formalizes the capture rule; \Cref{sec:exp:growth} measures the resulting rise.}
\label{fig:growth_envelope}
\end{figure*}

\label{sec:failure_types}%
\label{sec:adaptation_regimes}%
The new cases the field presents are concrete: a known grasp with the object farther away or occluded, a familiar carry chain meeting a heavier load, a route learned on grass arriving at dense weeds (\Cref{fig:field_deployment})---slightly out-of-distribution, within-family variants for which prior geometry and action structure remain usable.
A failure is classified by which part of the frozen model's prior knowledge is intact and which is missing, and before genuine novelty arrives two regimes cover what a deployed model encounters:
\begin{itemize}[leftmargin=*,itemsep=2pt,topsep=3pt]
 \item \textbf{Familiar mistake $\to$ Correction:} a known category with the wrong parameter (a grasp overshoot, a wrist rotation off by a few degrees), needing a tight, surgical fix rather than a new skill.
 \item \textbf{Within-family gap $\to$ Extension:} an unseen member of a known family (the model grasps cups but not thermoses, walks on concrete but not gravel), needing competence extended across related contexts.
\end{itemize}
Each regime is read off from signals the system already computes (the base policy's confidence and the Hamming distance from stored competence in the sparse distributed memory~\cite{kanerva1988}), so no external oracle is needed; the capsule's regime-dependent storage realizes the mapping (\Cref{sec:cc_definition}). CFAM expands competence within skill families it already holds---a failure with no relevant family is open-world novelty, outside this architecture's scope (\Cref{subsec:limitations}).

\section{Related Work}
\label{sec:related_work}

\label{sec:value_iteration_fails}%
\label{sec:unbalanced}%
\label{sec:two_timescales_field}%
Section~\ref{sec:diagnosis} established four field-learning requirements: efficient few-shot learning (R1), typed correction and extension (R2), bounded consolidation without interference (R3), and one-shot, gradient-free, on-device growth (R4). No prior family supplies the complete operational primitive these define---a closed adaptation object that captures competence at the edge of a frozen model's envelope, re-applies it during operation, and consolidates it under a fixed budget---so we assess four families against them before stating the architectural gap CFAM addresses (\Cref{sec:rw_gap}).

\subsection{Foundation Policies and Gradient-Free Adaptation}
\label{sec:rw_vla}

Vision-language-action models (RT-2~\cite{brohan2023rt2}, OpenVLA~\cite{openvla2024}, Octo~\cite{octo2024}, CogACT~\cite{cogact2024}, GR00T~\cite{groot2024}) collapse perception, reasoning, and action into one model that is \emph{frozen at deployment}; improvement requires centralized retraining, infeasible on edge hardware. CFAM factors the stack (\Cref{sec:system_overview}) and adds the new-learning layer no VLA provides.
Among gradient-free alternatives, skill-library and self-critique agents (Voyager~\cite{wang2023voyager}, ExpEL~\cite{zhao2024expel}, Reflexion~\cite{shinn2023reflexion}, REFLECT~\cite{liu2023reflect}, HELPER~\cite{sarch2025helper}, TidyBot~\cite{wu2023tidybot}, Statler~\cite{yoneda2024statler}, AdaPlanner~\cite{sun2023adaplanner}) act on symbolic skills with no physical execution operator, while residual and in-context adapters (Side-Tuning~\cite{zhang2020sidetuning}, MoS-VLA~\cite{mosvla2025}, MAC~\cite{tack2024mac}, MAP-VLA~\cite{mapvla2025}, TT-VLA~\cite{ttvla2025}, EVOLVE-VLA~\cite{evolvevla2025}, residual policy learning~\cite{silver2018residual,johannink2019residual}) produce global or gradient-trained residuals, and classical reinforcement learning needs rewards, enumerable states, and replayable trajectories the field does not offer, at per-step value-query cost.
Real-world policy improvement through repeated physical execution and feedback has also been explored in ENPIRE~\cite{xiao2026enpire}. CFAM occupies a third position: physically grounded capsules executed one shot per skill by GRT and sequenced at $O(M)$ emissions (\Cref{sec:skill_composition}), written one-shot, prediction-error-gated, and gradient-free. The skill primitive is a synthesis of established ingredients (movement primitives~\cite{ijspeert2013dmp}, task-parameterized mixtures~\cite{calinon2016tpgmm}, trajectory transfer~\cite{schulman2013trajectory}, keypoint affordances~\cite{manuelli2019kpam}, structured skill representations~\cite{yu2026skillgraph}) over a single persistent, consolidating capsule (\Cref{sec:capsule}).
DAgger~\cite{ross2011dagger} and RMA~\cite{kumar2021rma} share the structure but require an expert or privileged teacher; CFAM captures competence from self-observed signals alone (ebbing base confidence, rising retrieval distance) and accumulates explicit residuals over the lifetime.

\subsection{Memory-Augmented Policies and Retrieval}
\label{sec:rw_memory_augmented}

Memory-augmented robots organize storage by \emph{cognitive type}: MemoryVLA~\cite{memoryvla2025}, RoboMemory~\cite{robomemory2025} (after Tulving~\cite{tulving1972episodic}), EchoVLA~\cite{chen2025echovla}, and 3DLLM-Mem~\cite{yang2025threedllmmem}; lifelong skill-memory and skill-library approaches include ViReSkill~\cite{kagaya2025vireskill}, LRLL~\cite{tziafas2024lrll}, and Uni-Skill~\cite{xie2026uniskill}. CFAM organizes by \emph{learning function}, corrective and extension regimes with distinct support radii and generalization bandwidths (\Cref{sec:cc_definition}), and targets runtime growth under fixed memory and constant-time retrieval rather than temporal or spatial reasoning: a different desideratum, not a strict superset.
Its substrate extends Kanerva's sparse distributed memory~\cite{kanerva1988,kanerva1993sparse} (cf.\ Memory Layers at Scale~\cite{wu2024memorylayers}, Sparse Memory Finetuning~\cite{chen2024sparsememory}) with confidence-gated blending, one-shot capsule writes, and self-compressing consolidation.
One-shot imitation methods (Instant Policy~\cite{vosylius2025instant}, Coarse-to-Fine Imitation~\cite{johns2021coarse}) inspire GRT's geometric reasoning but operate statelessly; CFAM couples one-shot encoding with continual retrieval, so performance improves with experience.

\subsection{Test-Time and Continual Learning}
\label{sec:rw_tta}

Test-time adaptation (TENT~\cite{wang2021tent}, TTT~\cite{sun2020ttt}, CoTTA~\cite{wang2022cotta}, MEMO~\cite{zhang2022memo}) and parameter-efficient adapters (VPT~\cite{jia2022vpt}, IA$^3$~\cite{liu2022ia3}) adapt backbones by back-propagating through them, transiently, on unsupervised signals; CFAM uses no backward pass, gates writes on the base's own confidence and retrieval distance, and accumulates persistent capsules routed to typed memory regimes.
Classical continual learning, regularization (EWC~\cite{ewc}), dynamic architectures~\cite{progressive_nets}, replay, and meta-learning (MAML~\cite{finn2017model}), pursued under programs such as DARPA L2M~\cite{darpa2017l2m}, assumes training-time access to the task stream and gradient updates, both absent on a deployed edge device (\Cref{sec:diagnosis}).
Closest in structure are backbone prototype methods, NCM and incremental variants~\cite{mensink2013distance,rebuffi2017icarl,janson2022simple}, which mitigate forgetting by construction through additive per-class enrollment; CFAM is their physical-action generalization, replacing the per-class centroid with a typed, radius-addressed, geometrically warpable capsule that both corrects and grows (\Cref{sec:exp:continual}).
Both families' protocols announce task boundaries or score stationary test sets; the deployment-stream protocol of \Cref{sec:exp:setup} measures what they hold fixed, whether performance rises while the evaluation itself gets harder.

\subsection{Perception and Neuroscience Grounding}
\label{sec:rw_perception}

The Sensor cortex's SHDL~\cite{singh2017dtcwt,singh2017shdl} sits among hybrid parametric encoders rather than the monolithic deep encoders that dominate VLA work (ViT~\cite{dosovitskiy2020vit} and kin), trading end-to-end capacity for fixed front-end invariances, label efficiency, and continual category acquisition~\cite{singh2019thesis,singh2020humanlike}; the Scattering Vision Transformer~\cite{patro2023svt} externally validates the pattern at transformer scale.
The capsule's prediction-error lifecycle echoes, but was not reverse-engineered from, predictive coding and the free energy principle~\cite{friston2010free,rao1999predictive,clark2013whatever}, complementary learning systems~\cite{mcclelland1995complementary,kumaran2016}, population coding~\cite{pouget2000,georgopoulos1986}, and salience-gated consolidation~\cite{kandel2001molecular,friston2008,tulving1985memory}; the mapping is collected in the supplementary neuroscience-grounding appendix.

\subsection{Requirement-Level Gap and CFAM's Response}
\label{sec:rw_gap}

The comparison is consistent across families. Foundation policies provide broad initial competence but require centralized gradient retraining, missing R1 and R4. Symbolic skill libraries and retrieval memories can add information without changing the base, but they do not type correction versus extension or maintain a bounded, non-interfering physical competence store, missing R2 and R3. Test-time and continual-learning methods update parameters or transient state and therefore miss persistent, one-shot field growth under the edge budget, R3--R4. Prototype methods preserve old entries, but do not encode or geometrically execute physical skills.

CFAM is designed as the response to this requirement-level gap: a frozen slow-learning stack for broad competence (R1), typed Competence Capsules for corrective and extension writes (R2), a capsule field that consolidates them under a fixed budget (R3), and a one-shot, back-propagation-free write path (R4). The next section derives that architecture.

\section{A Brain-Inspired Architecture for Continual Learning}
\label{sec:theory}

The field learning regime of \Cref{sec:diagnosis} poses a structural question before it poses an engineering one: \emph{what shape must a model have to learn from single field events without erasing the competence it already holds?} We address this stability--plasticity tradeoff through explicit architectural separation rather than within one set of weights, which would have to be simultaneously plastic enough to absorb a one-shot event at the edge and stable enough to protect everything learned before it, the dilemma that per-weight remedies (regularization, replay, parameter isolation) mitigate but do not remove (\Cref{sec:related_work}). The resolution the brain arrives at is architectural, not parametric: split learning across \emph{two} subsystems that learn at different rates, a \textbf{complementary learning system}~\cite{mcclelland1995complementary,kumaran2016,dupoux2025}.

CFAM's architecture is inspired directly by the brain: it adopts this complementary-learning formulation as its design principle (\Cref{fig:two_part}). A \textbf{slow-learning phase} acquires broad competence efficiently in the laboratory (where data, compute, and supervision are plentiful) and is then frozen, playing the role of the slow-learning cortical component of a complementary learning system (an idealization: biological cortex learns slowly rather than not at all), so that at deployment it runs at minimum compute with no gradient machinery on board. What is stored during this slow-learning phase becomes the substance of the slow part that executes: its frozen internal parameters supply everything the deployed control cycle needs. A \textbf{fast-learning phase} is where all new learning happens in the field, encoding every competence acquired after deployment as a discrete unit in a shared memory, the way the hippocampus rapidly encodes new episodes. This learning is what happens during the fast part, and it is what grows the fast memory. The two phases learn at different rates and never interfere: new knowledge is never written back into the base. Non-interference is thus architectural: forgetting is mitigated by construction because field learning has no write path into the slow weights.

Functionally, the two parts divide the work of a deployed control cycle.
The \textbf{slow part executes}: everything that must happen every cycle, at control rate, inside the edge budget. It \emph{perceives}, it \emph{decides}, and it \emph{acts}, and none of this changes a single parameter.
The \textbf{fast part learns}: everything that changes the system. It \emph{selects} what is worth keeping, \emph{stores} it one-shot and gradient-free, \emph{re-applies} it whenever the situation recurs, and \emph{maintains} the store within a fixed memory budget.
The division is strict in both directions: the slow part never learns, and the fast part never acts on the world directly; it only modulates what the slow part is about to do.

The formulation also informs what the parts must contain. Because the slow part carries the full deployed control cycle, it must span the sensor-to-action pipeline, which gives it the form of \textbf{three cortices}: a \emph{sensor cortex} that lifts multimodal input into stable, 3D-grounded geometry (perceive); a \emph{reasoning cortex} that decomposes tasks into skills and judges their outcomes (decide); and an \emph{action cortex}, a geometric skill model that predicts a stored skill's trajectory in the current scene and executes each skill (act). Because the fast part must capture, in one write, everything a single field event teaches, its unit must bind the perception and the action content of that event to the situation that triggered it, and it must be compact, inspectable, and self-contained so that competence can be audited or removed. \Cref{sec:system_overview} details this realization: the three slow-learning cortices and the fast-learning memory they read from, whose unit is the Competence Capsule.

\section{The CFAM Robotics Architecture: Slow-Learning Cortices and Fast-Learning Memory}
\label{sec:system_overview}

The formulation of \Cref{sec:theory} motivates an end-to-end architecture, not a single module. A CFAM powers a physical asset the way a vision-language-action (VLA) policy does, and can be read as a specialization of the VLA robotics stack. We organize the deployed system as its four functional pieces (\Cref{fig:teaser}): the \textbf{Sensor cortex} supplies the 3D-grounded embedding; the \textbf{Reasoning cortex} sequences a task plan over the capsule field and serves as the outcome oracle; the \textbf{Action cortex} emits each planned skill as one geometrically warped emission; and \textbf{the new-learning layer} creates, refines, migrates, and consolidates skill competence over the deployment lifetime under a single update law, the Continual Field Update Rule ($\CFUR$).
The capsule field $\rcufield$ is the shared fast-learning memory, whose unit is defined in \Cref{sec:capsule}: it is read by the three runtime cortices at execution time and written only by the new-learning layer.
This section details the three cortices (we use \emph{cortex} and \emph{module} interchangeably) and then the fast-learning memory they read from. Among the four pieces, the \textbf{Action module} (unit-of-inference shift, \Cref{sec:action_module_overview}) and the near-edge extension path of \textbf{the new-learning layer} (unit-of-learning shift) are this paper's novel, empirically evaluated contributions; \Cref{sec:execution} covers that extension path, test-time growth, and compression. Correction from detected field failures, open-world novelty, field-time perception writes, and language-to-trajectory synthesis are outside this paper's scope (\Cref{subsec:limitations}).

\subsection{Sensor Cortex: ScatterNet Hybrid Deep Learning}
\label{sec:sensor_module}

\begin{figure*}[t]
\centering
\includegraphics[width=0.21\textwidth,height=4.3cm]{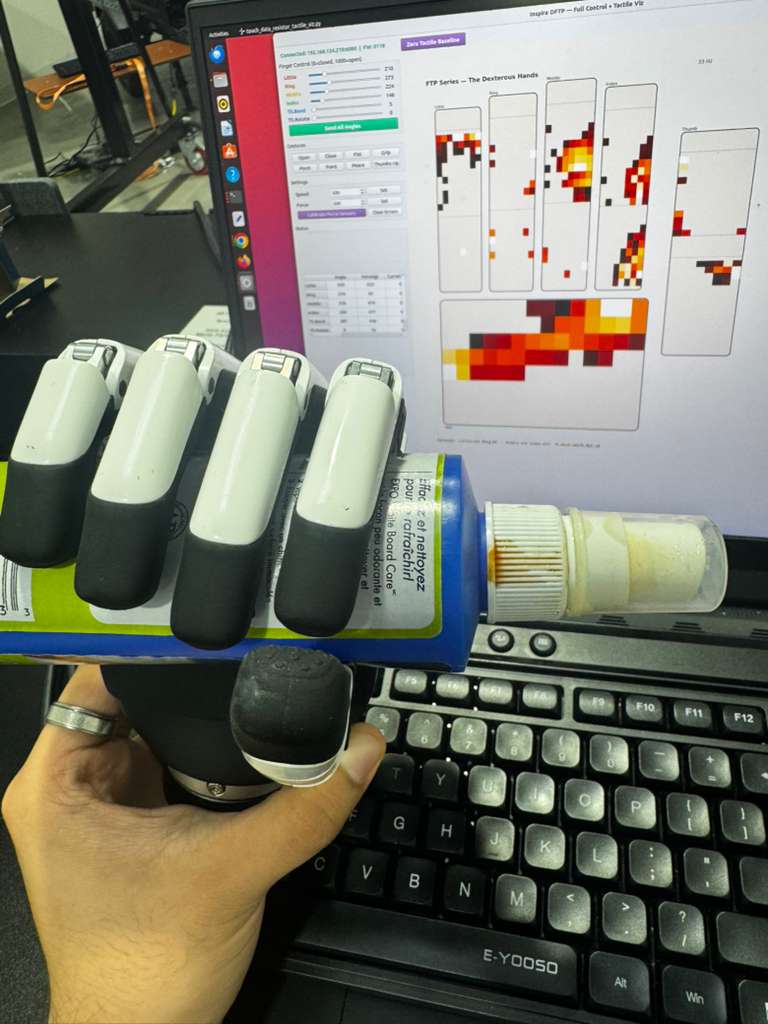}%
\includegraphics[width=0.21\textwidth,height=4.3cm]{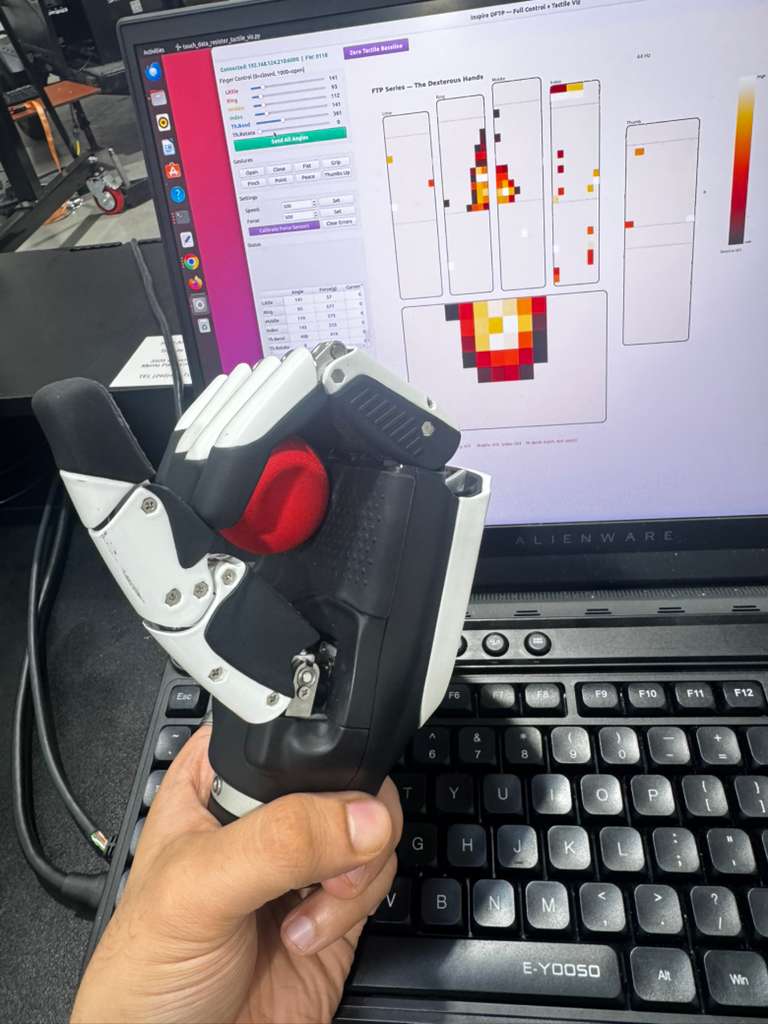}%
\includegraphics[width=0.21\textwidth,height=4.3cm]{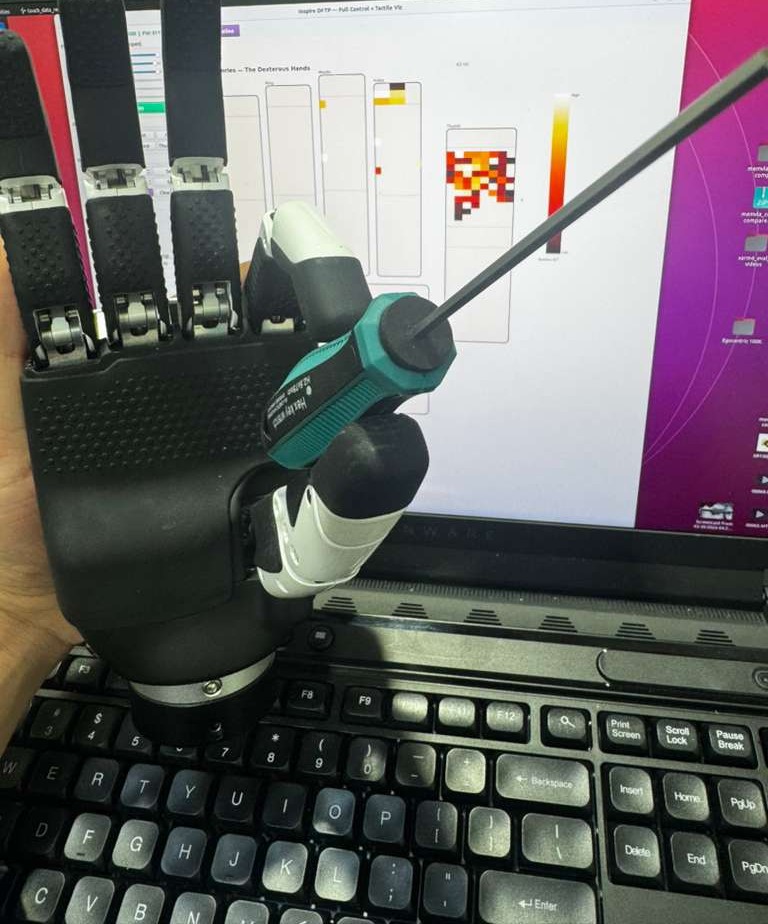}%
\includegraphics[width=0.21\textwidth,height=4.3cm]{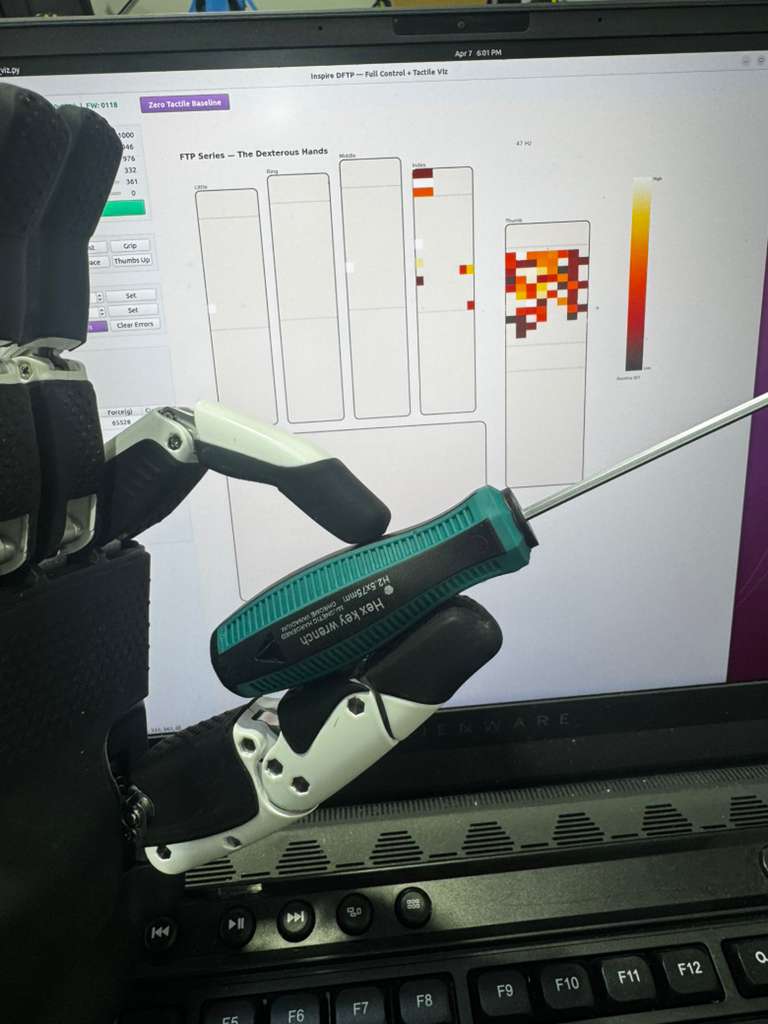}
\caption{\textbf{The tactile modality of the Sensor module.} Per-finger and palm tactile arrays during grasps of a bottle, a ball, and a hex key, with live per-taxel response maps; these sensor events ($\delta_{\text{sensor}}$: contact, slip, and force spikes) are inputs to failure detection.}
\label{fig:tactile}
\end{figure*}

The Sensor cortex's purpose is to \emph{see}: it converts raw multi-modal inputs (vision, audio, LiDAR, tactile) into the two products the rest of the stack consumes---a discriminative embedding $\vz \in \mathbb{R}^D$, binarized to the capsule field's retrieval address $\vh = \binarize(\vz)$ (\Cref{sec:action_module}), and per-modality sensor-event signals $\delta_{\text{sensor}}$ (tactile/force spikes, LiDAR proximity alerts, audio anomalies; the tactile arrays are shown in \Cref{fig:tactile}).

It adopts the \textbf{ScatterNet Hybrid Deep Learning (SHDL)} architecture~\cite{singh2017shdl,singh2019thesis,singh2023shdlpatent}: a semi-supervised, brain-inspired encoder that learns efficiently. Its three stages mirror the V1$\to$V2/V4$\to$IT pathway of the visual cortex (a fixed wavelet edge stage~\cite{singh2017dtcwt}, an unsupervised closed-form mid-level stage, and a small supervised classifier~\cite{singh2017efficient}), so only the final stage needs labels---giving classification performance comparable to deep CNNs from substantially smaller training sets~\cite{singh2017shdl,singh2019thesis}---and the fixed front-end keeps embeddings, and hence capsule addresses, stable under lighting, viewpoint, and pose variation. The same three-stage template applies to every modality, one architecture rather than a separate encoder per modality.

The load-bearing interface choice is that \emph{the Sensor cortex outputs 3D geometry, not pixels}: downstream of SHDL the scene is lifted into a 3D representation (point clouds from RGB-D, stereo, or LiDAR) from which task-relevant anchors (object centroids, grasp points, contact points, bottleneck waypoints) are extracted. Both the embedding $\vz$ that indexes $\rcufield$ and the current-scene anchor set $\mathcal{A}'_i$ that drives GRT live downstream of the same geometric lift, which is what makes the Action module's policy tractable (\Cref{sec:action_module_overview}).

\subsection{Reasoning Cortex: A Mission-Tuned VLM with Two Roles}
\label{sec:reasoning_module}

The Reasoning cortex's purpose is to \emph{understand}: it interprets the task stated verbally, infers the user's intent, plans at high level by decomposing the task into a sequence of stored skills, and serves as the oracle that judges whether each executed skill achieved its goal.

In detail, it is a single vision-language model $\mathcal{R}$ playing two roles inside the stack: capsule sequencer and outcome oracle.
\emph{$\mathcal{R}$ is custom-trained for mission-critical operations before deployment and then frozen: CFAM requires no planner that learns in the field.} In the deployed product configuration $\mathcal{R}$ is mission-tuned; in all experiments reported here $\mathcal{R}$ is an off-the-shelf frozen Qwen2.5-VL-7B (\Cref{sec:exp:setup}), so no result below depends on mission tuning. The ``CFAM Reasoning'' row of \Cref{tab:reasoning_pathway} is this same frozen model operating inside the CFAM stack, prompted at run time with the capsule descriptors of \Cref{eq:rcu_full}; the gap to the raw Qwen2.5-VL-7B rows measures what the capsule descriptors add to the sequencer, not a different model.
The mission tuning supplies the domain vocabulary and task structure of the target operations; what $\mathcal{R}$ is never trained on is the skill library itself. A planner that learns task structure during deployment violates non-iterative updates and fixed memory, and a planner trained against a lab-time library could not anticipate the post-deployment skills it will eventually have to sequence over, so $\mathcal{R}$ reads capsule descriptors at run time instead: in our implementation, $\mathcal{R}$ is a small VLM prompted with the capsule descriptors written at capsule-creation time.

\paragraph{Role A: capsule sequencer.}
Given a task instruction $L$ and a current observation $o_t$, $\mathcal{R}$ reads, for every capsule, its situation key $\vh_i$ (a field of $\rcu_i$) together with its descriptor tuple $d_i = (\mathrm{tag}_i,\,\mathrm{pre}_i,\,\mathrm{post}_i;\,\ldots)$ written at capsule creation time (\Cref{eq:rcu_full}), and emits an ordered plan
\begin{equation}
\label{eq:skill_plan}
\pi_L = (\rcu_{i_1},\, \rcu_{i_2},\, \ldots,\, \rcu_{i_M}),
\quad \rcu_{i_m} \in \rcufield \cup \{\textsc{synth}\},
\end{equation}
where each entry is either a capsule retrieved from the field or the symbol \textsc{synth} marking a library miss at that step.
The Action module then executes one capsule per plan step under GRT (\Cref{sec:action_module}). $\mathcal{R}$ is a read-only query layer over $\rcufield$, not part of $\CFUR$, and does not modify capsules.

\paragraph{Role B: outcome oracle.}
At a phase boundary $\mathcal{R}$ produces a binary verdict on whether the executed skill achieved its post-condition, $J_t = \mathrm{VLM}(o_t, L) \in \{0,1\}$. Two judgments are kept apart: $\CFUR$ writes are gated by the sensor-grounded outcome predicate $\Phi^{\mathrm{succ}}_i$ of \Cref{sec:grt}, never by $J_t$; $J_t$ is used only for semantic task progression and high-level post-condition evaluation, and the success rates of \Cref{sec:experiments} are scored independently of both.
Two routing questions follow: does the field contain a covering capsule for this step (otherwise the sequencer emits \textsc{synth}), and is the covering capsule performing adequately (otherwise per-phase diagnostics route to refinement).
$\mathcal{R}$ additionally emits bounded trajectory edits when a warp alone cannot reach the goal; because those edits act on the Action cortex's output, they are covered with execution (\Cref{sec:grt}).

The \textbf{Action cortex} that retrieves, warps, and executes the skill library is detailed next (\Cref{sec:action_module_overview}); the shared fast-learning memory it reads from, the Competence Capsule and the field it forms, follows (\Cref{sec:capsule}), and the growth and compression that run alongside are the subject of \Cref{sec:execution}.

\begin{figure*}[t]
\centering
\setlength{\tabcolsep}{2pt}%
\begin{tabular}{@{}ccc@{}}
\includegraphics[width=0.26\textwidth]{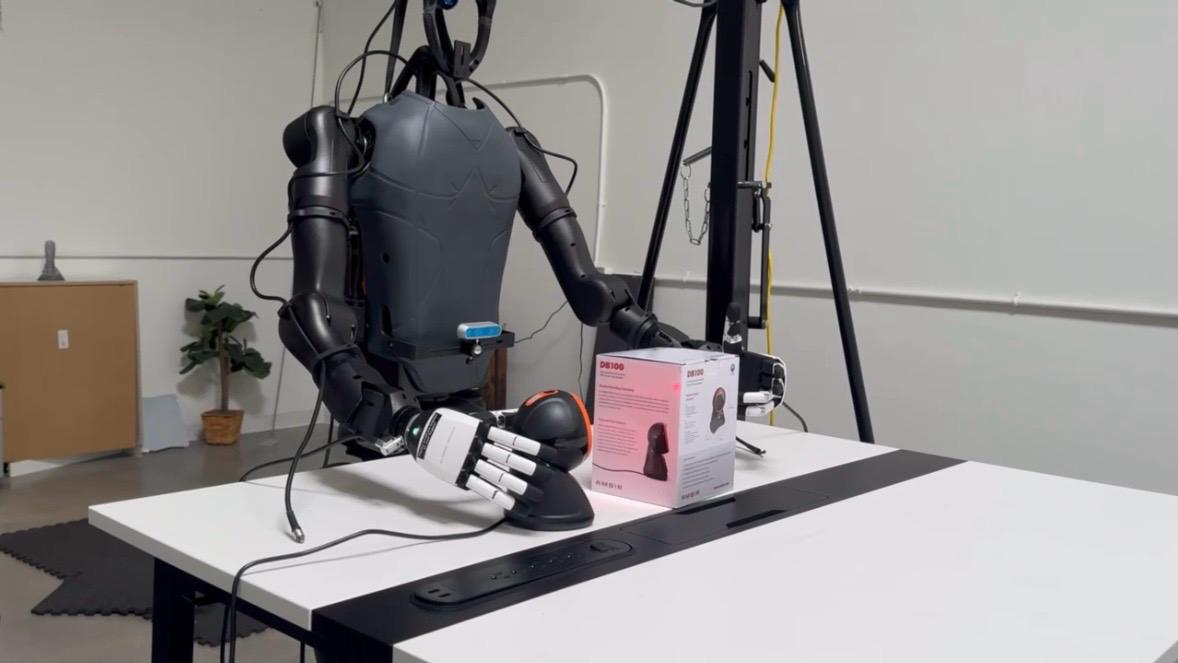} &
\includegraphics[width=0.26\textwidth]{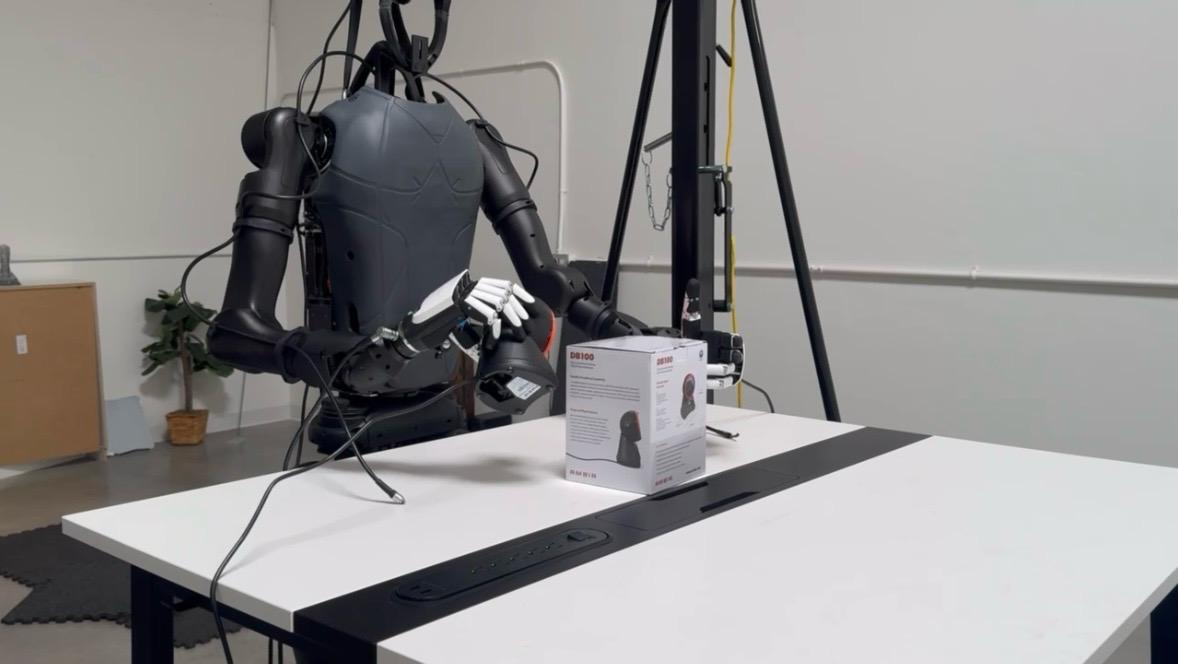} &
\includegraphics[width=0.26\textwidth]{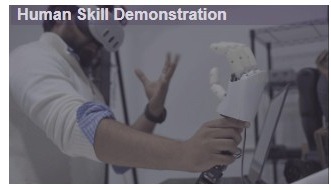} \\[2pt]
\includegraphics[width=0.26\textwidth]{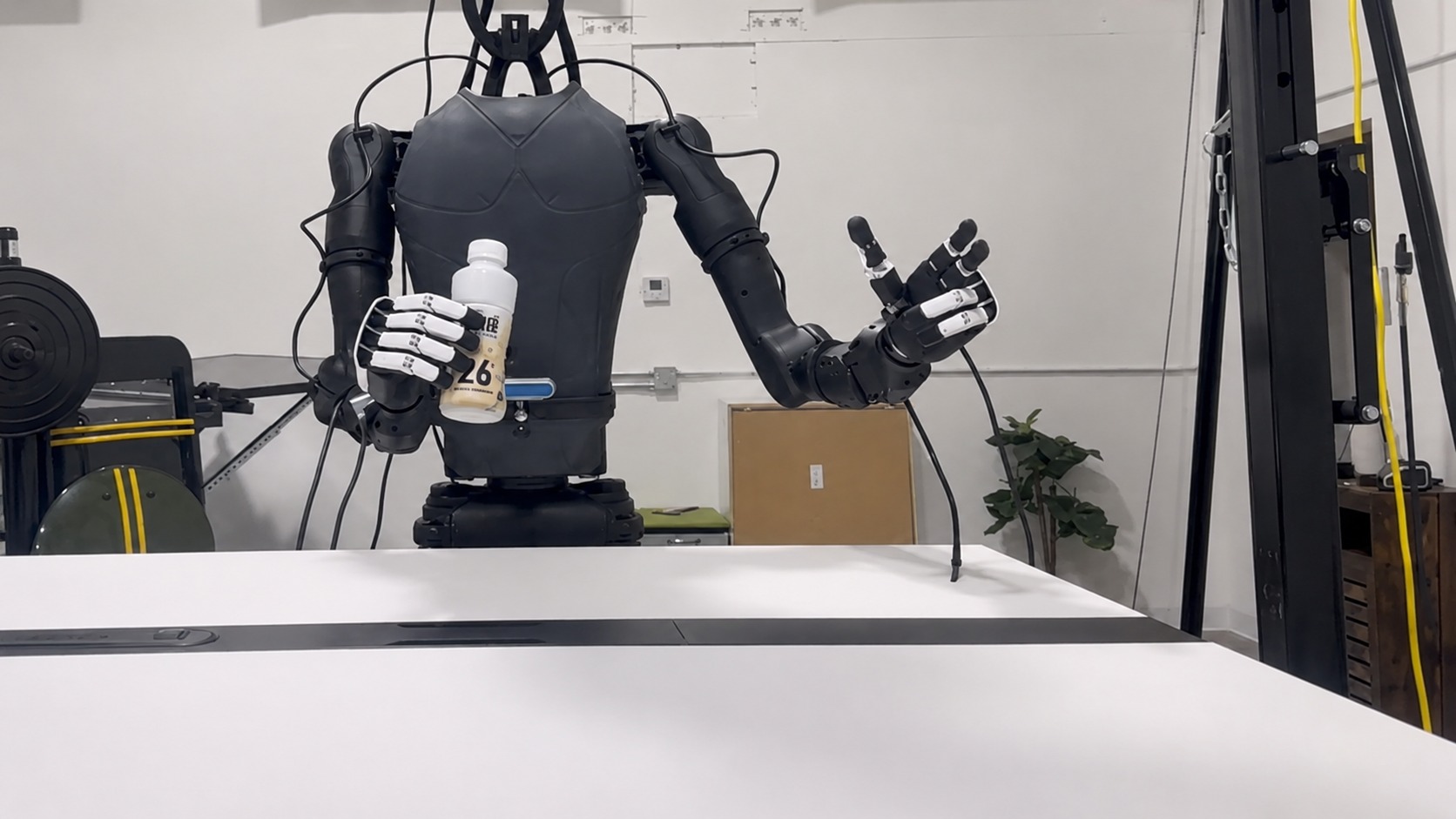} &
\includegraphics[width=0.26\textwidth]{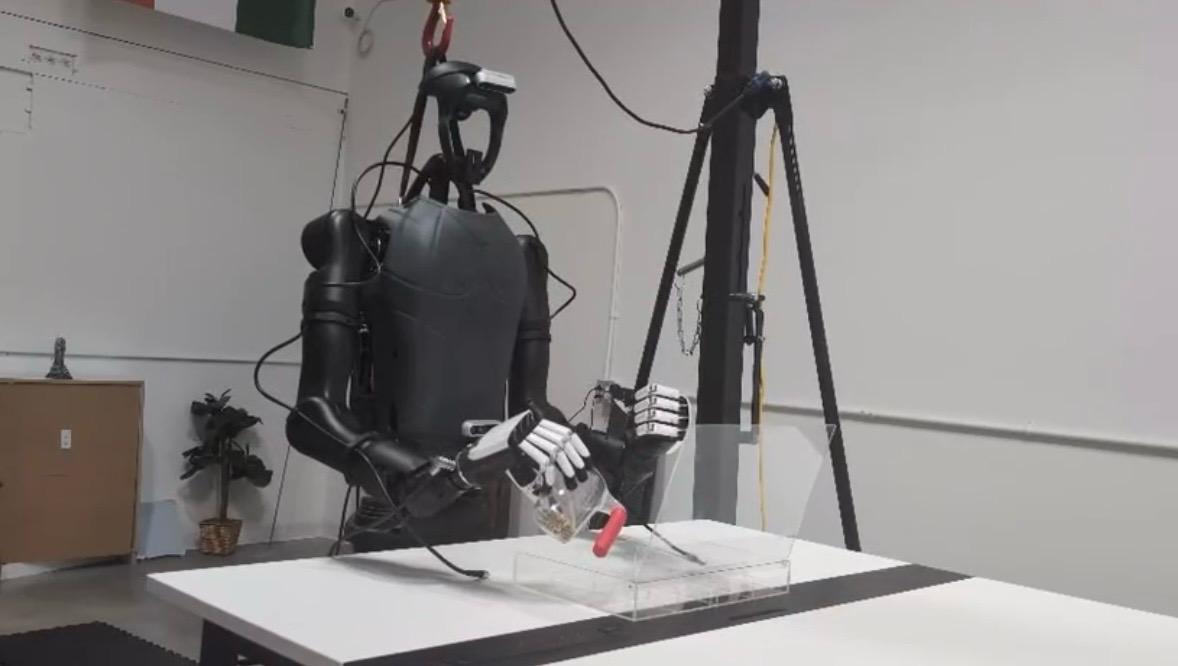} &
\includegraphics[width=0.26\textwidth]{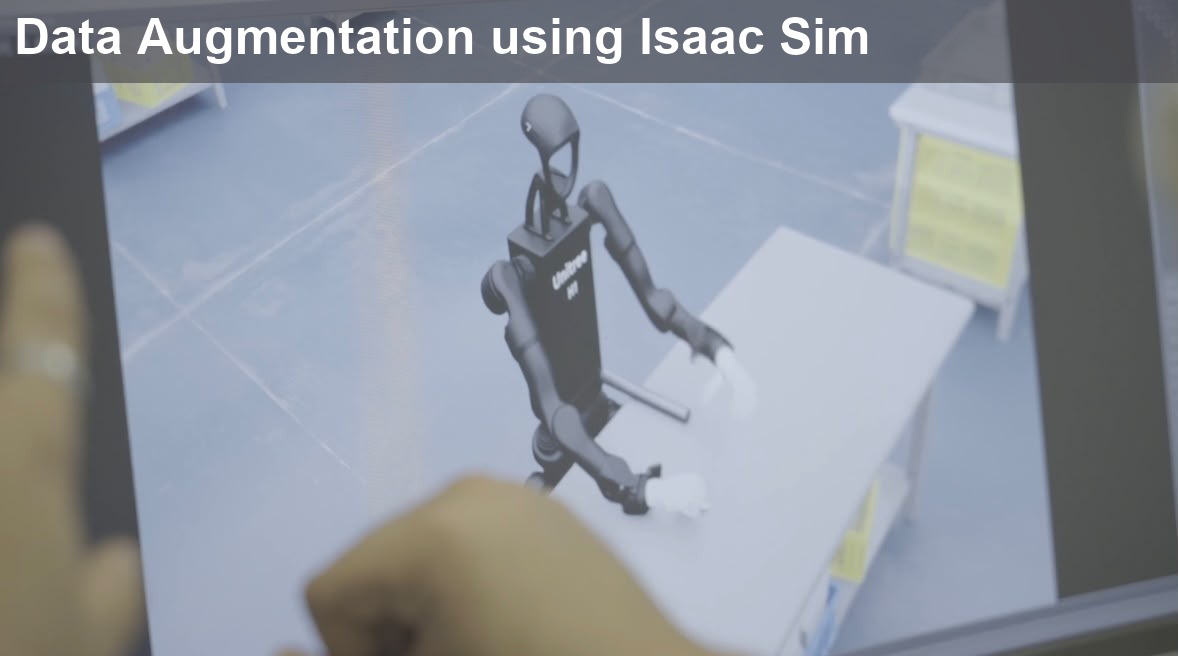} \\
\end{tabular}
\caption{\textbf{Building the library in the lab.} Seed skills are acquired from human demonstration on the bimanual humanoid, then augmented in simulation (Isaac Sim, bottom right); each demonstration is encoded into one Competence Capsule by a single one-shot write (\Cref{sec:capsule}).}
\label{fig:lab_training}
\end{figure*}

\begin{figure*}[t]
\centering
\setlength{\tabcolsep}{2pt}%
\begin{tabular}{@{}ccc@{}}
\includegraphics[width=0.27\textwidth]{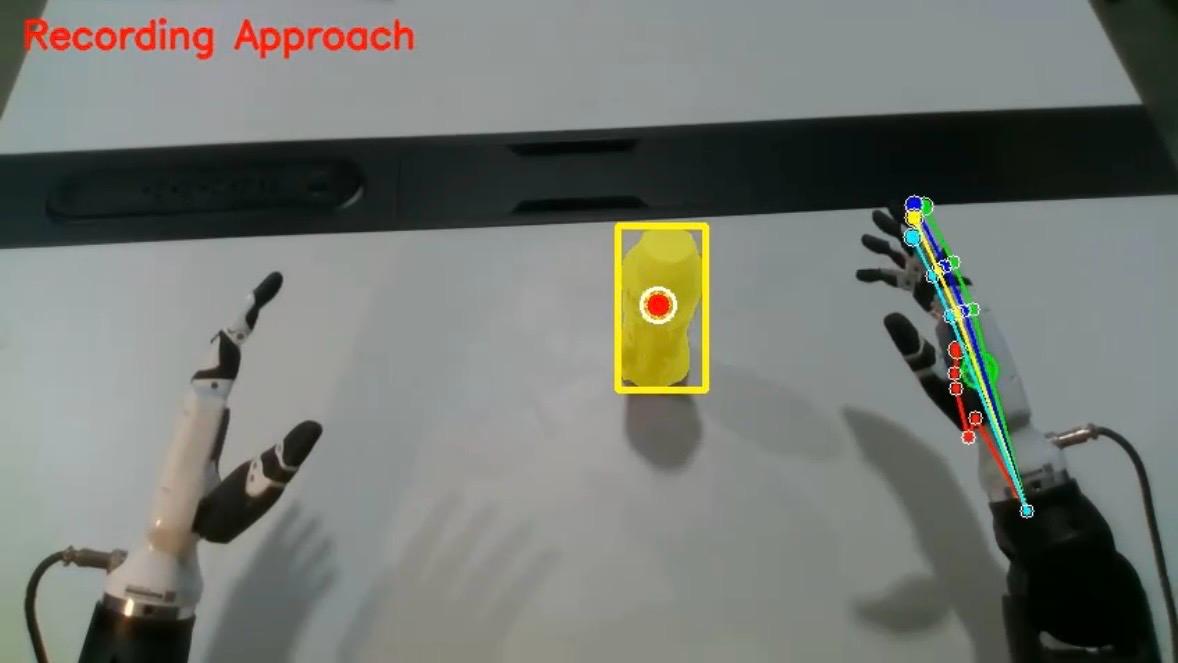} &
\includegraphics[width=0.27\textwidth]{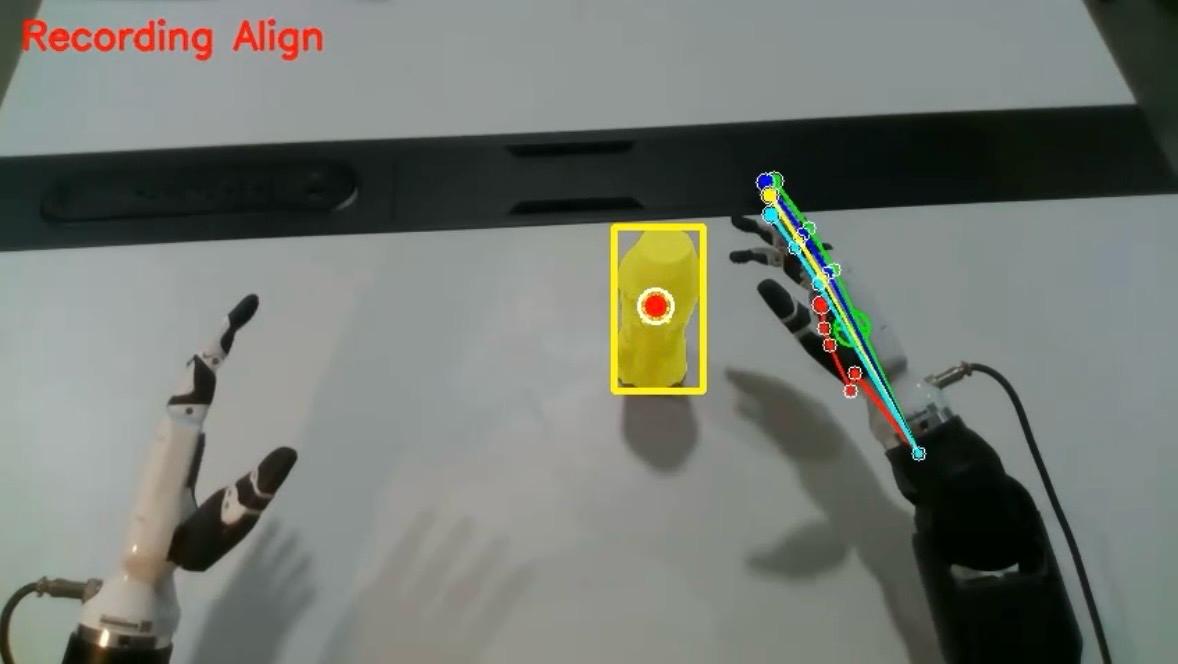} &
\includegraphics[width=0.27\textwidth]{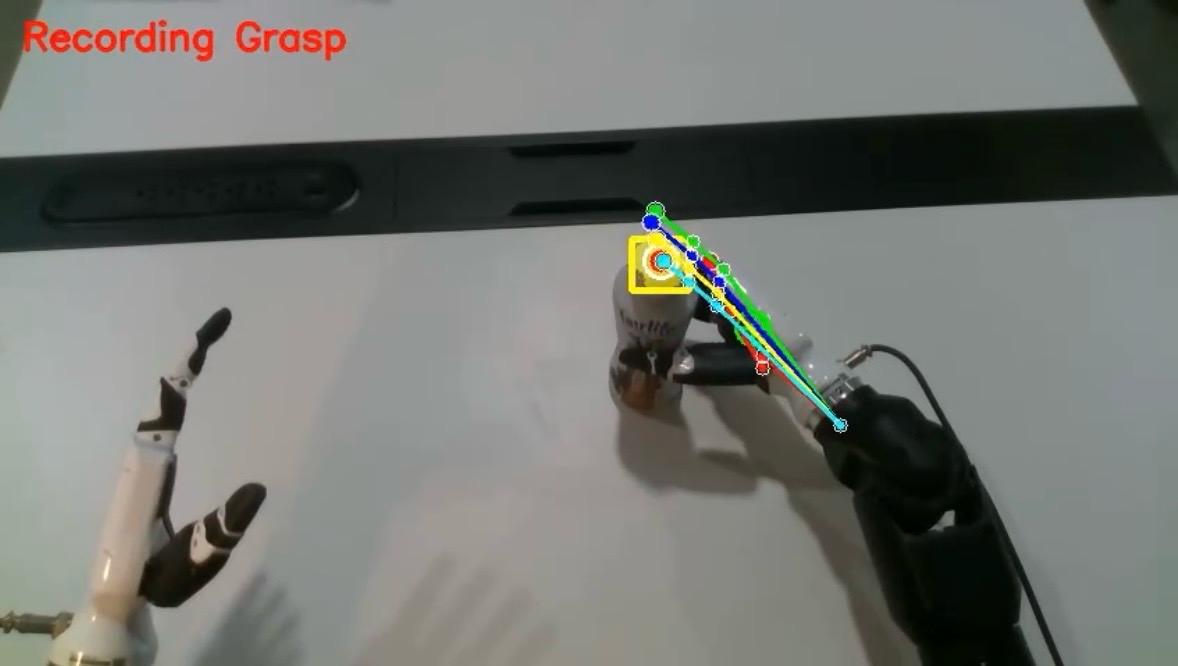} \\[1pt]
{\footnotesize (a) Approach} & {\footnotesize (b) Align} & {\footnotesize (c) Grasp} \\
\end{tabular}
\caption{\textbf{One skill, executed phase by phase.} A stored manipulation skill on the bimanual humanoid: (a)~approach, (b)~align, (c)~grasp, with the detected object (yellow box) and live hand-pose keypoints overlaid; each phase objective $\Phi$ must pass before the next begins (\Cref{sec:skill_composition}).}
\label{fig:mission_chain}
\end{figure*}

\begin{figure*}[t]
\centering
\begin{tabular}{@{}c@{\hspace{2pt}}c@{\hspace{2pt}}c@{\hspace{2pt}}c@{}}
\includegraphics[width=0.215\textwidth]{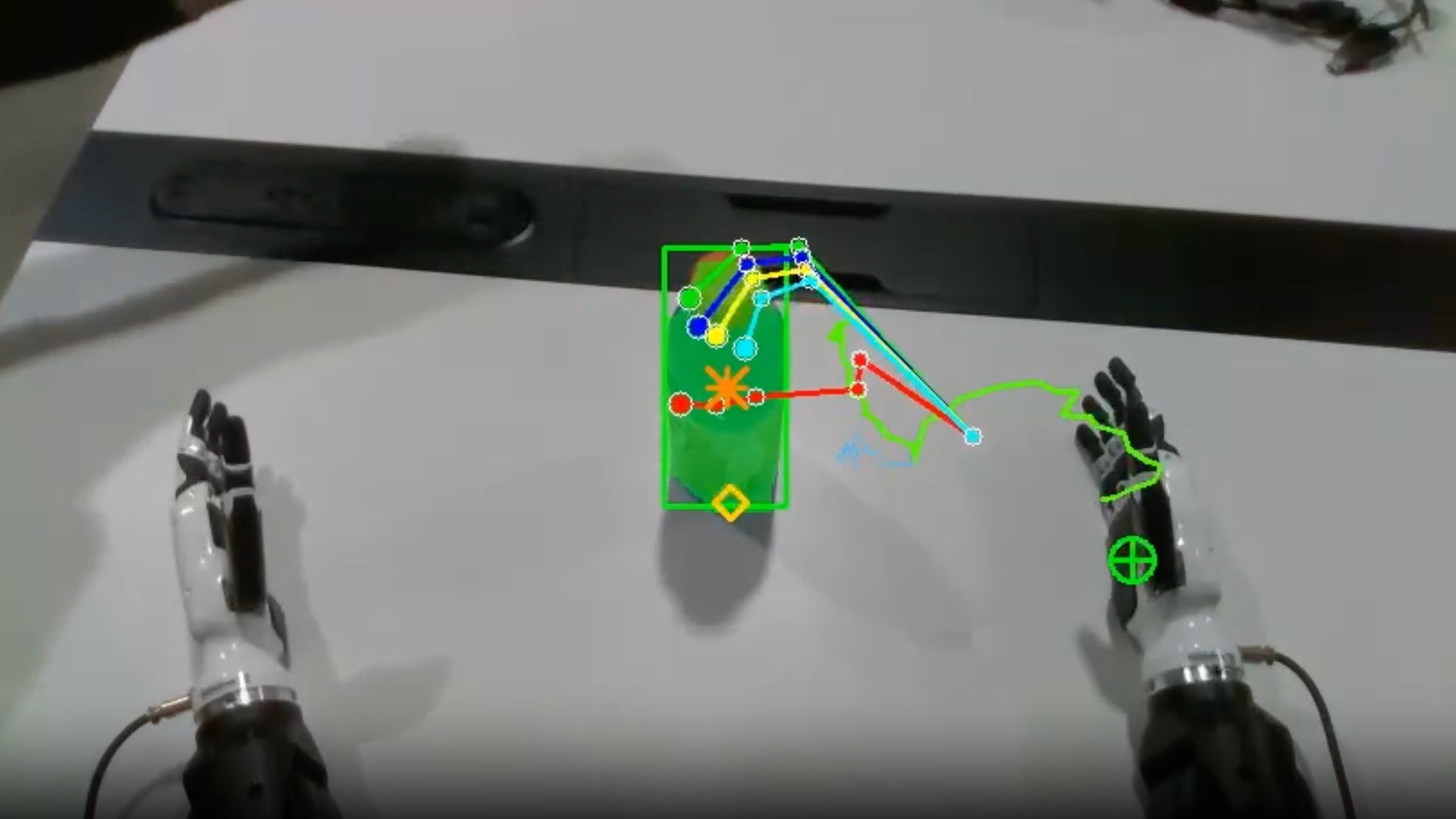} &
\includegraphics[width=0.215\textwidth]{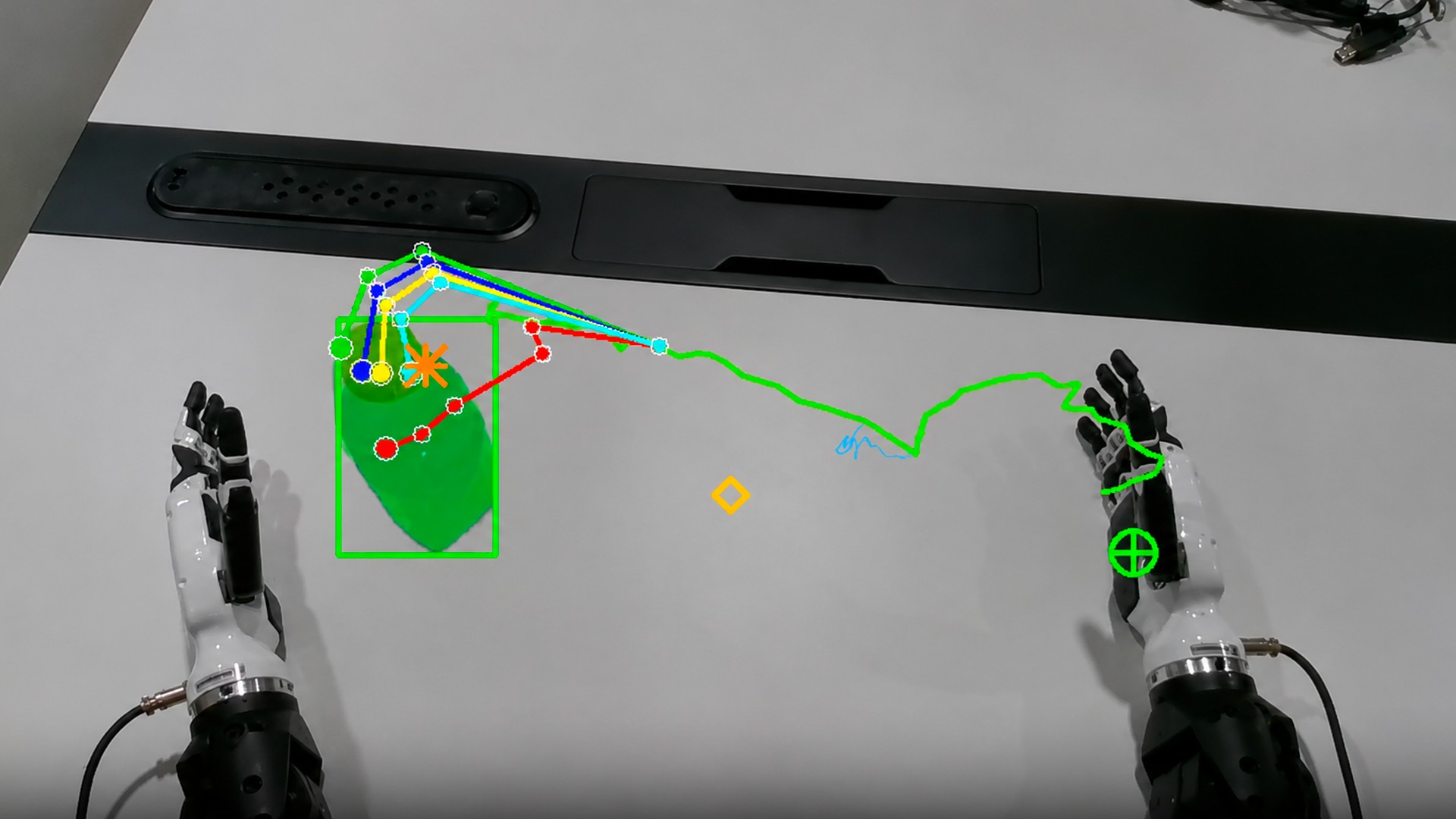} &
\includegraphics[width=0.215\textwidth]{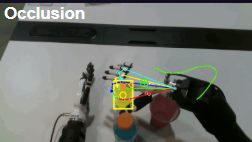} &
\includegraphics[width=0.215\textwidth]{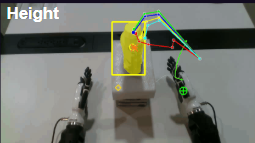} \\[-1pt]
{\scriptsize (a)} & {\scriptsize (b)} & {\scriptsize (c)} & {\scriptsize (d)}
\end{tabular}
\caption{\textbf{GRT on real hardware: one stored skill, any scene.} Four deployments of the \emph{same} stored grasp capsule: (a)~nominal pose, (b)~target displaced, (c)~occluded among clutter, (d)~raised to a different height, the latter two engaging the reasoning-mediated edit library (detour, raise) on top of the warp. Overlays: detected target, matched anchors, grasp point, bottleneck state, and end-effector trail. The capsule is stored once; per scene, GRT recomputes the transform and emits the warped trajectory in one shot, no retraining between panels.}
\label{fig:grt_real}
\end{figure*}

\subsection{Action Cortex: A Geometric Skill Model for Execution}
\label{sec:action_module_overview}
\label{sec:action_module_preview}%
\label{sec:action_module}%
\label{sec:architecture}%
\label{sec:skill_composition}

Once the task is understood, the Action cortex turns sensor information into action: it reads the plan $\pi_L$ from the Reasoning module, retrieves the corresponding capsules from $\rcufield$, and emits one geometrically warped skill per phase onto the 3D scene the Sensor cortex supplies (the capsule's stored fields read here are defined in \Cref{sec:capsule}).

\paragraph{Geometry in, skills out.}
\label{sec:geometry_frame}%
The Action module reads the 3D interface of \Cref{sec:sensor_module}, the embedding $\vz$ and the current-scene anchor set $\mathcal{A}'_i$, not raw pixels, and its online optimization is over the skill's geometric transform $T^{*}_i$ and a bounded phase/control residual $\Delta\theta_i$: low-dimensional, in physically meaningful coordinates, which is what makes deployment-time skill warping tractable on edge hardware (the full tractability argument is in the supplementary material).
Execution is per-skill, not per-step: a task is decomposed into physical phases (approach, align, grasp, lift, transport, place), and each phase is realized by retrieving a callable capsule from $\rcufield$ and emitting the warped skill in one deterministic forward pass---one GRT pass plus the clamped blend below, at $O(K_{\text{kp}} \cdot d_{\text{net}})$ cost, where $K_{\text{kp}}$ is the number of matched 3D anchors (keypoints) of the skill and $d_{\text{net}}$ the width of the Sensor cortex's embedding ($D$ in \Cref{sec:sensor_module}), with no diffusion steps and no per-control-step decoding. A task that a frame-by-frame VLA decodes in $T$ control steps therefore needs only $M \ll T$ capsule emissions: the number of neural action-decoding calls per task drops from $T$ to $M$, independent of state-space size and reward structure (\Cref{sec:value_iteration_fails}); the low-level controller still closes its loop at every one of the $T$ control steps, but no network is decoded there.

\paragraph{Selecting and blending the matching capsule(s).}
\label{sec:inference}
\label{sec:generalization}
\label{sec:pgt_apply}%
The plan $\pi_L$ settles \emph{which skills, in what order} (\Cref{sec:reasoning_module}); retrieval settles \emph{which stored capsule(s) realize each step} in the current scene.
For each step, the capsule field $\rcufield$ is queried read-only, within R4's fixed compute and memory envelope: the current observation is matched by similarity against stored situation keys, the closest capsules within the regime-specific support radius are selected, and when several nearby capsules apply their stored values are combined under a bounded (\emph{clamped}) blend, weighted by proximity and confidence. No learning happens here (\Cref{sec:cc_definition} covers how the stored values are acquired).

Retrieval runs two passes. A perception pass shifts the embedding $\vz$ to a corrected address $\vz'$ by a clamped, confidence-weighted sum of the stored perception values, pulling toward the perception anchor $\vz^*_i$ when the encoder itself is unreliable; an action pass then re-queries memory at $\vz'$ and combines the stored action values the same way into an additive estimate $\aadd$ (base output plus blended residuals) and an anchor estimate $\aanchor$ (blended stored actions). Both sums act on per-control-step action vectors, never on stored trajectories: for each active capsule the anchor pair $(\tau_i, \mathcal{A}_i)$ is first warped into the current scene by GRT (\Cref{eq:grt_transform}), and the warped emission's action at the current phase step is the ``stored action'' the anchor sum combines, while the residual sum combines the $\resid_i$ directly; the bounded-authority statement therefore compares action vectors of the same dimensionality, and the largest single stored value is the largest warped per-step action among the active capsules. This per-step blend is a fixed-cost vector operation applied to an emission that GRT has already produced for the whole phase; it is not the per-control-step network decode that a frame-by-frame VLA performs. The base output $\abase$ entering \Cref{eq:blend_rcu} is likewise not decoded per control step: the base emits its action for the whole phase once ($\pi_0$ emits an action chunk; the in-house prior emits a phase-length chunk), \Cref{eq:blend_rcu} blends within that chunk at each step, and the confidence signal that gates the blend and the capture trigger is computed once per phase from that emission. Each clamped sum divides by $\max(1, \sum w_i \trust_i)$, with $w_i$ decaying with Hamming distance from the query address, so a combination of several capsules can never overshoot the largest single stored value (bounded authority). The output blends the two estimates under a confidence gate:
\begin{equation}
 \aout = \alphaaction\,\aadd + (1 - \alphaaction)\,\aanchor, \label{eq:blend_rcu}
\end{equation}
where $\alphaaction$ is computed from the base policy's confidence (output entropy, logit spread, or an auxiliary head): $\alphaaction \approx 1$ in \textsc{corr} (pure additive residual), decreasing toward the anchor-dominant blend in \textsc{ext} as base reliability drops, recovering the per-regime storage behavior of \Cref{sec:cc_definition}.
Because both gates measure the base's \emph{self-agreement}, not correctness, the independently evaluated phase objective $\Phi_i$, which reads perceived and proprioceptive quantities rather than the model's own estimate, is the backstop against a confidently wrong base.
The full per-pathway blending and gate equations are in the supplementary material. In this paper the stored perception fields are captured directly from the demonstration at build time.

\paragraph{Warping and executing a skill: GRT.}
\label{sec:grt}
The \textbf{Geometric Residual Transform (GRT)} warps a selected skill onto the current physical scene, acting as a \emph{geometric skill model} (not a learned model of environment dynamics): it predicts the skill's full spatial trajectory $\hat{\tau}_i$ before execution and checks it against the phase objective $\Phi_i$ (\Cref{fig:grt_mechanism}; real-hardware deployments in \Cref{fig:grt_real}).
For a capsule with stored anchors $\mathcal{A}_i$, seed trajectory $\tau_i$, and admissible transform family $\mathcal{T}_i$, GRT selects
\begin{align}
\label{eq:grt_transform}
 T^{*}_i &= \arg\min_{T \in \mathcal{T}_i}\;
 D_w\bigl(T\,\mathcal{A}_i,\; \mathcal{A}'_i\bigr)
 + \lambda\,\Omega(T),\\
 \hat{\tau}_i &= T^{*}_i(\tau_i),
 \quad \text{executed iff } \Phi^{\mathrm{adm}}_i\bigl(\hat{\tau}_i,\; o\bigr) \leq \varepsilon^{\mathrm{adm}}_i,
\nonumber
\end{align}
where $\mathcal{A}'_i$ are the current-scene 3D anchors, $D_w$ is a task-weighted, skill-conditioned anchor discrepancy, and $\Omega(T)$ charges unnecessary transform complexity (preferring the simplest admissible family, rigid $\mathrm{SE}(3)$ by default). The first line is the geometric alignment: for the default rigid family it is solved in closed form by weighted Procrustes, and the similarity, affine, and non-rigid families use their corresponding solvers. Two predicates are kept apart. $\Phi^{\mathrm{adm}}_i$ is the \emph{admissibility} check on the candidate before it moves (collision margin, reachability, joint limits, contact geometry), the second line above; $\Phi^{\mathrm{succ}}_i$ is the sensor-grounded \emph{outcome} check on the executed trajectory (contact and force, closure, tracking tolerance, stability), which cannot be known before execution and is the only predicate that gates a $\CFUR$ write (\Cref{sec:growth_exec}). Both are evaluated from onboard sensing alone, so no VLM sits in the loop; together they are the $\Phi_i$ of \Cref{eq:rcu_full}.
The pipeline is therefore three-stage: weighted Procrustes generates the \emph{candidate} transform, $\Phi^{\mathrm{adm}}_i$ admits it, and after execution $\Phi^{\mathrm{succ}}_i$ scores it; a candidate that fails admissibility, or an execution that fails a phase, receives a bounded local residual $\Delta\theta^{*}_i$ ($\|\Delta\theta\| \leq \varepsilon_i$) restricted to the failing phase---if alignment succeeds but grasp fails, only grasp is updated, with no gradient through the frozen VLA.
The predicates are not hand-coded per skill. They are instantiated automatically at Build from the demonstration itself: the phase-objective template of the supplementary material (goal, alignment, contact, obstacle, smoothness, joint, and vision terms with phase-specific weights) is filled in from the demonstrated trajectory's phase segmentation and the sensed contact, force, and tracking statistics of the demonstration, which set the active terms and their tolerances, and a field-written capsule inherits the predicate of the capsule it was warped from. A new skill family therefore needs a demonstration, not a predicate author, and ``no operator label'' holds at Build as well as in the field. The number of distinct predicate instances and their false-positive and false-negative rates are not reported (\Cref{subsec:limitations}).
For locomotion, aerial, and wheeled skills the same operator applies with different anchors: the anchor set is the terrain, waypoint, and obstacle frames of the task (the task frames of the supplementary material), the seed trajectory is the skill's body or base trajectory expressed in those frames (footstep and body-height profile for a gait phase, waypoint path for a flight segment, path and speed profile for a wheeled traverse), and $\Phi^{\mathrm{succ}}_i$ is the corresponding stability or tracking predicate; what GRT re-aims is the skill's trajectory in its task frames, not an object.
The novelty is not geometric alignment per se but GRT as a continual, phase-conditioned, task-weighted, bounded-authority operator inside a persistent skill memory (\Cref{sec:related_work} positions it against movement primitives, trajectory transfer, keypoint affordances, and residual policy learning); its execution-time guarantees, \emph{locality} and \emph{bounded authority} (\Cref{sec:cc_definition}), hold by construction, with proofs, full instantiations, and the transform-family hierarchy in the supplementary material.

\paragraph{Reasoning-mediated trajectory editing.}
\label{eq:reasoning_edit}%
GRT absorbs variations a stored capsule can reach under its admissible transform family (rigid $\mathrm{SE}(3)$ for object pose and orientation; similarity $\mathrm{Sim}(3)$ where scale changes; affine for mild deformation) but not variations that require topological change to the trajectory itself: reaching around an occluder, raising the wrist over a height obstacle.
For these, the reasoning model emits an edit on top of the warped trajectory, drawn from a bounded library of waypoint operations (detour, raise, dwell) parameterized by the obstacle geometry $\mathcal{R}$ extracts from $o_t$ and clamped to a per-step authority limit analogous to \Cref{eq:blend_rcu}'s bound, so the trajectory deviation stays bounded even when the edit is wrong.
An edit that succeeds twice on related scenes is written back into $\rcufield$ as a new capsule via the standard $\CFUR$ path, so the skill that needs the edit is acquired over time rather than re-derived by $\mathcal{R}$ on every encounter. The edits themselves are exercised in \Cref{fig:grt_real}(c)--(d).

\paragraph{Library miss.}
When $\mathcal{R}$ cannot map a step of $L$ to any capsule in $\rcufield$, the plan emits \textsc{synth} at that position.

The capsules executed this way are \emph{validated competences}, each demonstrated in the lab or captured from a verified test-time success, so no exploration mechanism is required or used in this paper. The full execution model (skill-capsule view, task frames, bottleneck-state chaining, phase objectives) is developed in the supplementary material.
The fast-learning memory all of this reads, and the new-learning layer writes, is defined next.

\subsection{The Fast-Learning Memory: The Competence Capsule}
\label{sec:capsule}
\label{sec:cc_definition}
\label{sec:rcu}

The three cortices above execute, but none of them learns: per the formulation of \Cref{sec:theory}, everything learned in the field is encoded into the fast-learning memory they read from, never overwriting what the base already holds (\Cref{fig:motivation}).
That memory's single shared unit is the \textbf{Competence Capsule (CC)} (\Cref{fig:cc_schema}). One capsule stores one local competence element of a skill---a \emph{perception} correction (a shift $\Delta\vz$ to the Sensor embedding) and an \emph{action} correction (a residual $\resid$ on the executed skill), bound to the situation $\vh$ that triggers them---and the system generalizes from these one- or few-shot writes. The Sensor and Action sides read their respective slices, the Reasoning module reads capsule descriptors to compose plans (it never writes), and a single one-shot write path ($\CFUR$) is the sole writer. All capsules live in one place, the capsule field $\rcufield$, held on edge hardware within a fixed memory budget.

\paragraph{Where capsules come from: building the library.}
\emph{Before deployment}, the library is \emph{built} in the lab: an operator provides a few demonstrations of each target skill (teleoperation, kinesthetic guidance, or a handful of successful rollouts), and each demonstration is encoded by the one-shot write, with no gradient descent over the frozen model. A demonstration covers a whole task; it is segmented at its phase boundaries (\Cref{fig:mission_chain}) into the task's two to four skills, and each segment becomes one capsule, so ``one demonstration per task'' yields one Build capsule per skill phase of that task. A skill is not limited to one capsule: as its envelope grows, further capsules of the same skill are added at its edge (\Cref{fig:growth_envelope}, $\rcu_i, \rcu_i', \rcu_i''$), so ``one capsule'' means one local competence element, and a skill is in general represented by several. The stored geometry then lets a single demonstration generalize across object poses and scenes, warped by GRT (\Cref{sec:grt}).
\emph{After deployment}, the same write extends the library on-device whenever new information is identified at the edges of stored competence (\Cref{sec:growth_exec}).

Every capsule carries eight fields (seven numeric and one regime tag), plus a descriptor tuple written once at creation:
\begin{itemize}[nosep]
 \item \textbf{Situation key} ($\vh_i$): a binary code that determines when this capsule activates. Similar scenes produce nearby keys, enabling local generalization.
 \item \textbf{Perception correction} ($\Delta\vz_i$): shifts the Sensor module's embedding toward the correct region.
 \item \textbf{Perception anchor} ($\vz^*_i$): the encoder's embedding captured during the supervised correction event, a fallback for when the encoder is completely unreliable.
 \item \textbf{Action correction} ($\resid_i$): the residual added to the frozen model's output.
 \item \textbf{Action anchor} ($\astaraction_i$): the action executed during the capture event, stored as the seed trajectory $\tau_i$ together with the 3D anchors $\mathcal{A}_i$ it was recorded against; this is the pair GRT reads in \Cref{eq:grt_transform}, and it is the fallback for when the base policy's output is unreliable.
 \item \textbf{Persistence} ($\sal_i$): how long the capsule survives without reactivation.
 \item \textbf{Confidence} ($\trust_i$): how strongly the correction is applied. Confidence grows with successful reuse.
 \item \textbf{Adaptation regime} ($\phase_i = (\phase^{\text{percep}}_i, \phase^{\text{action}}_i)$): a pair of per-side tags, each in $\{\text{passive},\,\textsc{corr},\,\textsc{ext}\}$ (a novelty tag is reserved for future use). Each component governs its side's support radius, generalization bandwidth, and lifecycle, with the design prior $\srad_{\textsc{corr}} < \srad_{\textsc{ext}}$: a corrective update stays tightly bounded around the situation that evidenced it, while an extension update is granted wider generalization bandwidth (the radius--capacity tradeoff behind this prior is analyzed in the supplementary material). The two sides can differ (e.g., perception in \textsc{ext}, action in \textsc{corr}).
\end{itemize}

\noindent The \textbf{canonical state} of a Competence Capsule is:
\begin{equation}
\label{eq:rcu_full}
\begin{aligned}
 \rcu_i &= \bigl(\vh_i,\; \Delta\vz_i,\; \vz^*_i,\; \resid_i,\; \astaraction_i \equiv (\tau_i,\,\mathcal{A}_i),\; \sal_i,\; \trust_i,\; \phase_i\bigr),\\
 d_i &= (\mathrm{tag}_i,\, \mathrm{pre}_i,\, \mathrm{post}_i;\; \mathcal{T}_i,\, \Phi_i),
\end{aligned}
\end{equation}
where the action anchor $\astaraction_i$ carries the seed trajectory $\tau_i$ and its anchors $\mathcal{A}_i$ (so \Cref{eq:grt_transform} and \Cref{fig:grt_mechanism} read only capsule fields), and $d_i$ is the descriptor tuple: the skill tag and pre-/post-conditions read by the Reasoning cortex in \Cref{eq:skill_plan}, the admissible transform family $\mathcal{T}_i$, and the phase objective $\Phi_i$ used by GRT. Descriptors are metadata fixed at creation, outside the eight fields that the blend, decay, and consolidation rules act on.

\Cref{sec:execution} covers what runs alongside execution: test-time growth and compression of this library.

\section{Test-Time Growth: Learning from the Extremes}
\label{sec:execution}

With the library built in the lab and the execution machinery of \Cref{sec:skill_composition} in place, this section covers what changes as the system runs: odd, near-edge cases are \emph{identified} from signals the system already computes, \emph{added} to the capsule field as new capsules, \emph{generalized} from, as new points from which GRT can warp, and \emph{compressed} so the library stays within its fixed memory budget.
Acquiring a genuinely new skill when none in the library applies (open-world novelty) is outside this paper's scope.

\begin{figure}[b]
\centering
\setlength{\tabcolsep}{2pt}%
\begin{tabular}{@{}cc@{}}
\includegraphics[width=0.487\columnwidth]{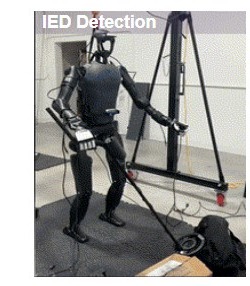} &
\includegraphics[width=0.487\columnwidth]{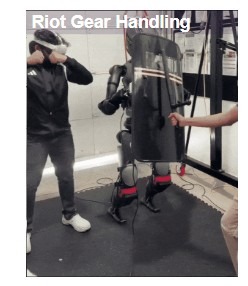} \\
\end{tabular}
\caption{\textbf{Test-time execution in the field.} Representative executions on the humanoid platform: IED recognition (the REC skill exercised here on the humanoid; the metal-detector IED-sweep mission of \Cref{tab:mission_set} runs on the quadruped) and protective-gear handling. Slightly out-of-distribution variants remain within the same skill family but change the physical arrangement---for example, a new combination of bag and box locations that requires a different maneuver to reach a shield, or a heavier shield that changes the verified grasp--carry execution.}
\label{fig:field_test}
\end{figure}

\subsection{Growth: Adding Skills at the Extremes}
\label{sec:growth_exec}
Growth captures \emph{extremes}: situations at the edge of a stored skill's range, where the base policy's generalization is starting to fall off (\Cref{fig:growth_envelope}). This is not a failure to be detected: the base is still operating, so the trigger is not an oracle but two signals the system already computes, a drop in the base policy's \emph{confidence} (higher output entropy, a smaller logit margin, the same signal that gates blending, \Cref{sec:inference}) and a rise in the \emph{retrieval distance} (the situation key lies in the outer margin of the nearest covering capsule's support radius); together they mark the edge of the envelope. When a warp lands at such an edge and still \emph{succeeds}, its outcome predicate $\Phi^{\mathrm{succ}}$ verified, it is written back on-device as a new capsule by the one-shot write path ($\CFUR$), so the library grows to cover that extreme and the same situation is met directly next time rather than re-warped from afar: an \textsc{extension} in the sense of \Cref{sec:adaptation_regimes}.
The autonomy rests on an error the system computes for itself. The base's own uncertainty and the memory's distance-to-support together say that this sample sits at the edge of what is stored; $\Phi^{\mathrm{succ}}$ says that the warped execution nevertheless succeeded within a known family, which is what makes the sample relevant rather than noise. What the write stores is not a new action (the old capsule already reached it) but a persistent new local support point: the next warp starts from this point rather than from the original demonstration, so the envelope the skill can reach moves outward with each capture (\Cref{fig:growth_envelope}). A sample that recurs inside an existing capsule's support does not create a new capsule: the trigger does not fire, and the verified reuse replenishes that capsule's confidence and persistence instead, so repeated experience strengthens the same capsule; only when captures accumulate densely around one key does consolidation merge them (\Cref{sec:compression_exec}).
Stated as the update rule, $\CFUR$ on this path is: \emph{trigger} when the base confidence falls below a threshold and the Hamming distance of the situation key from its nearest covering capsule lies within a fixed margin below that capsule's support radius (the sample is at the edge but still covered, so a warp from that capsule exists); \emph{verify} the executed emission with $\Phi^{\mathrm{succ}}$ on every phase (\Cref{eq:grt_transform}), with $J_t$ playing no part; \emph{write} a new capsule with key $\binarize(\vz_t)$, the executed warped trajectory and current anchors as $\astaraction$, the residual $\resid$ as the difference between the executed action and the base output, the perception fields carried over from the capsule that was warped, regime $\phase=(\text{passive},\textsc{ext})$, and initial confidence and persistence values; on every later verified reuse, \emph{replenish} $\trust_i \leftarrow \min(1, \trust_i + \eta_{\trust})$ and $\sal_i \leftarrow \min(1, \sal_i + \eta_{\sal})$; \emph{decay} persistence every cycle by \Cref{eq:decay_rule} and remove capsules below $\sal_{\min}$; and \emph{consolidate} when local density exceeds its threshold (\Cref{sec:compression_exec}). A second, rarer write trigger is the reasoning-mediated edit of \Cref{sec:grt}, written back after two verified successes on related scenes through the same write step. No existing capsule and no base weight is modified by a write. Genuine novelty (no stored skill of the right kind, where confidence and retrieval distance alone cannot separate a capturable extreme from a real breakdown) requires failure detection and is outside this paper's scope.
The library therefore compounds as it runs: every capsule added widens the region the Action module can warp into, so novel situations are increasingly met by warping a nearby skill in few, often zero, new shots; this mechanism predicts the rising post-deployment trajectories of \Cref{fig:build_ttgen_overview}. Each warp is one forward pass, and retrieval runs over a fixed memory budget, so the per-emission cost is bounded by the configured budget rather than growing without limit as the library fills.

\paragraph{Growth only: no training, and a coverage-routed decision.}
Neither network is trained: the slow-learning base stays frozen, and the fast-learning memory changes only by gaining new capsules, so \emph{growth is the only learning happening here}---continual learning in CFAM \emph{is} the accumulation of the fast memory. Decision-making is likewise not a joint optimization over the two networks but a routing by coverage, in the manner of complementary learning systems: when the current situation falls inside the support radius of a stored capsule, the fast path drives (the capsule supplies the skill and its corrections, confidence-weighted and clamped, \Cref{sec:inference}, with the base contributing only the reference output); when no capsule covers the situation, the slow path acts alone.

\subsection{Compression: Pruning and Consolidating}
\label{sec:compression_exec}
The library is held within a fixed memory budget, so growth is balanced by compression. When many capsules cluster around a single core skill, say a dozen near-identical \emph{pick} variations, they are consolidated into one representative, while capsules that go unused over time decay in salience and are removed. Coverage therefore rises without the stored count growing unbounded, and the most useful skills stay sharp.
\label{sec:consolidation}%
This keeps ongoing capsule creation, in the lab and in the field, inside a fixed budget (R3), running on-device on the same $\CFUR$ schedule.

\paragraph{Decay.}
Persistence decays geometrically:
\begin{equation}
\label{eq:decay_rule}
 \sal_i \leftarrow \gamma_{\phase_i}\, \sal_i,
\end{equation}
where $\gamma_{\phase_i} = \max(\gamma_{\phase^{\text{percep}}_i},\, \gamma_{\phase^{\text{action}}_i})$ takes the slower of the two side-specific rates. Within each side, $\gamma_{\textsc{corr}} < \gamma_{\textsc{ext}}$: corrective fixes decay fastest because they address immediate errors, whereas extension capsules persist longer because a newly covered family member needs time to be refined (a still-slower rate is reserved for the novelty regime). Taking the max-conservative side ensures that a capsule new on either pathway is retained until its rarer side is refined.
Capsules with $\sal_i < \sal_{\min}$ are removed. Frequently used capsules persist indefinitely because successful reuse replenishes their salience faster than decay reduces it (shown in the supplementary material); under $\CFUR$ the fixed budget prioritizes retention by reuse frequency, recency, and coverage value, following frequency- and recency-sensitive memory allocation under limited capacity~\cite{singh2020humanlike}.

\paragraph{Compression.}
When local density exceeds a threshold, similar capsules are merged into a single representative or a small core set of representatives. Full details, together with the memory substrate's implementation properties, are in the supplementary material.

\providecommand{\PH}[1]{\textcolor{red}{#1}}

\section{Experiments}
\label{sec:experiments}
\label{sec:analysis}

This section evaluates CFAM in four ordered stages on Skylark's in-house multi-embodiment dataset: \textbf{(A)} prior training on $D_\mathrm{train}$ (learning-curve efficiency at fixed data fractions, with $\pi_0$/CogACT/SpatialVLA trained on the same dataset as matched-data baselines), \textbf{(B)} held-out test on $D_\mathrm{test}$, \textbf{(C)} autonomous test-time growth on the variation stream $D_\mathrm{var}$, and \textbf{(D)} retention of earlier competence. This is the \emph{Build $\to$ Grow $\to$ Retain} cycle tagged by split.

\begin{table*}[t]
\centering
\caption{\textbf{Training datasets and evaluation protocols across synthetic and robotic settings.} $D_\mathrm{train}$ is the trajectory count used to train the frozen prior (a public backbone in simulation, the in-house multimodal prior on the five physical platforms); it is not per-task CFAM supervision. $D_\mathrm{test}$ is the held-out Stage-B test protocol, given as scored trials \emph{per seed} and the number of seeds (the column sums to $1{,}784$ scored trials per seed; rates are macro-averaged over task--condition cells and then over seeds, \Cref{sec:exp:setup}); the Stage-A learning curve is scored on a separate fixed held-out set $D_\mathrm{curve}$, disjoint from $D_\mathrm{train}$ and from $D_\mathrm{test}$ (\Cref{sec:experiments}). $D_\mathrm{var}$ is the variation stream on which autonomous test-time growth is measured (\Cref{fig:build_ttgen_overview}); each row lists the number of controlled axes (viewpoint, object position, orientation, lighting, occlusion) and the number of held-out variation conditions (a base task under one variation setting, not a trial count); $|D_\mathrm{var}|$ is $544$ scored trials per seed across the eight rows, about $68$ per row on average, which is $23.4\%$ of the $2{,}328$ scored trials per seed of $D_\mathrm{test}$ and $D_\mathrm{var}$ together. Each task contains two to four skills; the missions of \Cref{tab:mission_set} are longer chains (three to six skills) built from the same skill library.}
\label{tab:datasets}
\footnotesize
\setlength{\tabcolsep}{3pt}
\renewcommand{\arraystretch}{1.05}
\rowcolors{3}{white}{tabzebra}
\begin{tabularx}{\textwidth}{@{}P{1.1cm}P{2.15cm}C{2.1cm}C{1.05cm}C{1.55cm}C{3.0cm}>{\centering\arraybackslash}X@{}}
\toprule
\rowcolor{tabhead}
\textbf{Type} & \textbf{Dataset / platform} & \textbf{Embodiment} &
\textbf{Tasks} & \textbf{$D_\mathrm{train}$} & \textbf{$D_\mathrm{test}$} &
\textbf{$D_\mathrm{var}$ (axes / conditions)} \\
\midrule
\textbf{Synthetic} & \textbf{Bridge} & WidowX &
4 & 60,096 & 96 trials, 3 seeds & 4 / 13 conditions \\
& \textbf{Fractal} & Google Robot &
5 & $\sim$130K & 540--1{,}500 trials, 3 seeds & 5 / 17 conditions \\
& \textbf{LIBERO} & Franka Panda &
40 & 2{,}000 & 800 trials, 3 seeds & 4 / 11 conditions \\
\midrule
\textbf{Robotic} & \textbf{Arm} & Franka Panda &
10 & 1.1M$+$ & 200 trials, 3 seeds & 5 / 19 conditions \\
& \textbf{Dog} & Unitree Go2 &
4 & 87K$+$ & 48 trials, 3 seeds & 4 / 14 conditions \\
& \textbf{Humanoid} & Unitree H1 &
2 & 1.2M$+$ & 40 trials, 3 seeds & 5 / 16 conditions \\
& \textbf{Drone} & Quadrotor &
\mbox{3 sites} & 294K$+$ & 30 trials, 3 seeds & 4 / 18 conditions \\
& \textbf{Ground-vehicle} & Off-road vehicle &
3 & 65K$+$ & 30 trials, 3 seeds & 5 / 12 conditions \\
\bottomrule
\end{tabularx}
\end{table*}

\begin{itemize}
    \setlength{\itemsep}{0.25em}
    \setlength{\topsep}{0.35em}
    \setlength{\parsep}{0pt}
    \setlength{\parskip}{0pt}
    \item \textbf{Evaluation domains.} Controlled synthetic benchmarks (SimplerEnv, LIBERO) support repeatable analysis, while robotic experiments provide physical validation across five assets: a Franka Panda manipulator, Unitree Go2 quadruped, Unitree H1 humanoid, quadrotor, and off-road vehicle.
    \item \textbf{Baseline comparisons.} On the physical platforms, the standard-policy baselines ($\pi_0$, CogACT, SpatialVLA) are trained on the same in-house $D_\mathrm{train}$ (matched-data); on SimplerEnv and LIBERO they are the public policies; adaptation baselines (LoRA, MemoryVLA, CronusVLA) update the same standard policies after Build.
    \item \textbf{Four-stage protocol (A/B/C/D).} Every quantitative result below sits inside one of four stages, and every CFAM number in a caption or sentence carries the tag (split, stage, prior/backbone).
    \begin{itemize}[label={--},leftmargin=1.1em,nosep,topsep=1pt]
        \item \textbf{Stage~A --- Prior training on $D_\mathrm{train}$:} CFAM's slow-learning cortices are trained on the Skylark in-house multi-embodiment dataset (2.6 million$+$ trajectories across the five robotic platforms); on the physical platforms the standard-policy baselines are trained on the same dataset (matched-data), while in simulation CFAM's capsule field sits over a public backbone ($\pi_0$ for SimplerEnv, SpatialVLA for LIBERO). \emph{Learning-curve evaluation:} for each training fraction $f \in \{20, 40, 60, 80, 100\}\%$, CFAM and the matched-data standard policy are trained using only $f\,D_\mathrm{train}$. Each resulting checkpoint is evaluated on the same fixed held-out set $D_\mathrm{curve}$, which is disjoint from $D_\mathrm{train}$; no $D_\mathrm{curve}$ example is used for training or capsule construction, and the same tasks, trial counts, seeds, and success metric are used at every fraction. $D_\mathrm{test}$ is a separate held-out evaluation used only for Stage~B. The CFAM--standard-policy gap on $D_\mathrm{curve}$ at each fraction defines $\mathrm{Gain}_\mathrm{pre}$.
        \item \textbf{Stage~B --- Held-out test $D_\mathrm{test}$:} held-out task executions on the same platforms, disjoint from $D_\mathrm{train}$ and scored under the trials $\times$ seeds protocol of \Cref{tab:datasets} (a fixed trial protocol per platform, not a fraction of $D_\mathrm{train}$), evaluated after Build and before autonomous growth (\Cref{tab:action_pathway}); $\mathrm{Gain}_\mathrm{test}$ is the CFAM--standard-policy gap. $\CFUR$ test-time writes are enabled only in Stage~C and in the sequential suite of \Cref{sec:exp:continual}; in Stage~A, Stage~B, and the LIBERO ablations they are disabled, so every capsule present there was written at Build.
        \item \textbf{Stage~C --- Variation stream $D_\mathrm{var}$:} a further split of $544$ scored trials per seed, $23.4\%$ of the $2{,}328$ scored trials per seed of $D_\mathrm{test}$ and $D_\mathrm{var}$ together (\Cref{tab:datasets}), produced by controlled variations of the originals (viewpoint, object position, orientation, lighting, occlusion). CFAM does a few-shot Build, then captures verified near-edge cases (\Cref{tab:vla_adaptation,tab:real_growth_family,tab:mission_set}); the static standard policies fall along the stream (\Cref{fig:build_ttgen_overview}); adaptation baselines receive the same $D_\mathrm{var}$ exposure under their own update mechanism. $\mathrm{Gain}_\mathrm{post}$ is the CFAM--baseline gap.
        \item \textbf{Stage~D --- Retention:} in the sequential simulation suite, after each new environment, earlier environments are re-evaluated (\Cref{tab:continual}, \Cref{fig:retention}, \S\ref{sec:exp:continual}); the physical platforms are not re-evaluated on $D_\mathrm{test}$ after Stage~C.
    \end{itemize}
    Per-platform $D_\mathrm{train}$ and $D_\mathrm{test}$ sizes appear in \Cref{tab:datasets}.
    \item \textbf{Catastrophic forgetting.} Sequential experiments test whether CFAM retains earlier competence as new environments and skills are introduced.
\end{itemize}

\smallskip
\noindent\textbf{Results organization.}
Section~7.1 defines the baselines, metrics, and evidence boundary. Sections~7.2--7.5 then present Stage-A/B (Build and $D_\mathrm{test}$), Stage-C (post-deployment growth on $D_\mathrm{var}$), Stage-D (retention), and ablations, respectively.

\subsection{Evaluations and Baselines}
\label{sec:exp:setup}

This subsection establishes the common experimental frame used throughout the results. It introduces the evaluation domains, explains the two comparison classes, and defines the criteria and evidence boundary used to interpret later results. \Cref{tab:datasets} organizes the synthetic and robotic protocols by embodiment, build input, test scale, and task properties.
\label{sec:exp:missions}

\paragraph{Synthetic evaluations.}
Controlled benchmarks isolate four questions: initial manipulation competence at Build, continued competence acquisition through Grow across a sequential environment stream, retention of earlier competence as new environments are introduced, and component-level behavior under ablation. SimplerEnv provides the controlled, sequential, and retention settings, while LIBERO provides the ablation and fine-tuning-complementarity setting. \Cref{tab:datasets} distinguishes the Build input from the evaluated tasks and their controlled variations for each synthetic suite.

\paragraph{Real-world evaluations.}
Physical experiments test the same learning claims at increasing levels of difficulty: initial Build competence on single-phase Franka tasks (Real General, ten tabletop tasks) and continued Grow through mission chains across five physical assets. These experiments establish whether the measured gains persist outside simulation and across embodiments; retention is measured in the sequential simulation suite only (\Cref{sec:exp:continual}). \Cref{tab:datasets} identifies the embodiment, demonstration budget, test scale, and task properties for each physical platform.

\paragraph{Prior and baseline configurations.}
CFAM's fast-learning capsule field sits over a frozen slow-learning prior. On the five physical platforms of \Cref{tab:action_pathway,tab:vla_adaptation,tab:real_growth_family,tab:mission_set} that prior is our in-house multimodal model, trained on the in-house $D_\mathrm{train}$; on the simulation benchmarks it is a public backbone ($\pi_0$ for SimplerEnv, SpatialVLA for LIBERO). Two Franka testbeds are exceptions and use the public $\pi_0$ backbone: the ten-task Real General suite of \Cref{sec:exp:franka} and the sequential testbed of \Cref{subsec:complementarity}; numbers from them are always tagged as such. In every physical-platform table, $\pi_0$, CogACT, and SpatialVLA rows are those policies trained on the same in-house $D_\mathrm{train}$ (matched-data baselines), not the released public policies. The Reasoning cortex $\mathcal{R}$ is the off-the-shelf frozen Qwen2.5-VL-7B~\cite{bai2025qwen25vl} in every experiment (\Cref{sec:reasoning_module}). Adaptation-based (LoRA) and memory-based (MemoryVLA, CronusVLA) methods provide the post-deployment comparisons of \Cref{tab:vla_adaptation}; retrieval-prompting alternatives (RAG, VLM failure-prompt) are discussed qualitatively in \Cref{sec:exp:adaptation_methods}. \emph{Aggregation:} every reported rate is macro-averaged across task--condition cells and then averaged across the three seeds; consequently the percentages are not constrained to integer-success increments of the aggregate rollout count, and the per-seed scored-trial counts of \Cref{tab:datasets} are the denominators of the underlying cells, not of the reported rate.

\paragraph{Evaluation criteria.}
The analysis combines task success or accuracy and one-shot/few-shot efficiency with forward and backward transfer, forgetting, component ablations, and the source of observed gains. These measurements respectively test initial competence, continued learning without catastrophic forgetting, and the contribution of the component pathways.

\paragraph{Evidence boundary.}
Autonomous near-edge growth here means operator-free capture of a physically verified, successful within-family extreme after the supervised few-shot Build. A failed execution with no usable prior skill still requires an available corrective or successful trajectory. All reported growth remains within known skill families; open-world novelty injection and later field-learning results are outside this paper.


\subsection{Build: Initial Competence}
\label{subsec:one_shot}
\label{sec:exp:franka}

\noindent\textbf{Evaluation question.} What competence does one demonstration buy at deployment, and how does that operating point compare with standard build baselines?\par\smallskip

This subsection measures the competence available at the end of the supervised Build stage, before automatic field growth begins. Synthetic results establish the controlled manipulation operating point on SimplerEnv ($\pi_0$ backbone) and LIBERO (SpatialVLA backbone), while robotic results test one-shot acquisition across task families and the five physical embodiments (in-house prior).

\begin{table*}[t]
\centering
\caption{Comparison of VLA models with CFAM on the held-out test split $D_\mathrm{test}$; CFAM after few-shot Build, before autonomous test-time growth. The CFAM row uses the public $\pi_0$ backbone for the SE-Bridge and SE-Fractal columns and the in-house multimodal prior for the five robotic columns. In the five robotic columns the $\pi_0$, CogACT, and SpatialVLA rows are those policies trained on the same in-house dataset as matched-data baselines (not the released public policies); in the SE-Bridge and SE-Fractal columns they are the released public policies (trained on their own public data), and the $\pi_0$ row there is the same frozen backbone the CFAM row sits on.}
\label{tab:action_pathway}
\footnotesize
\renewcommand{\arraystretch}{1.12}
\setlength{\tabcolsep}{2.5pt}
\rowcolors{2}{white}{tabzebra}
\begin{tabularx}{\textwidth}{@{}>{\raggedright\arraybackslash}Xccccccc@{}}
\toprule
\rowcolor{tabhead}
\textbf{Method} & \textbf{SE-Bridge} & \textbf{SE-Fractal} & \textbf{Real Arm} & \textbf{Real Dog} & \textbf{Real Humanoid} & \textbf{Real Drone} & \textbf{Real Vehicle} \\
\midrule
CogACT & 51.3 & 68.1 & 61.5 & 54.5 & 52.5 & 55.2 & 56.3 \\
SpatialVLA & 42.7 & 75.1 & 60.4 & 50.5 & 49.6 & 48.9 & 45.6 \\
$\pi_0$ & 68.4 & 71.4 & 59.2 & 52.3 & 55.0 & 58.5 & 55.6 \\
\textbf{CFAM (Ours): $\pi_0$ backbone (sim) / in-house prior (real)} & \textbf{76.4} & \textbf{78.9} & \textbf{68.6} & \textbf{59.7} & \textbf{60.8} & \textbf{71.5} & \textbf{68.5} \\
\bottomrule
\end{tabularx}
\end{table*}

\paragraph{Rate of learning.}
We measure prior-training data efficiency on the fixed Stage-A learning-curve split $D_\mathrm{curve}$, not on $D_\mathrm{test}$. At every training fraction, both CFAM and the matched-data standard policy (the in-house-trained $\pi_0$) are evaluated on the same $D_\mathrm{curve}$ protocol. At 100\% of $D_\mathrm{train}$, the standard policy reaches $52.2\%$ on $D_\mathrm{curve}$; CFAM reaches the same operating point using 40\% of $D_\mathrm{train}$ ($52.3\%$), corresponding to $2.5\times$ fewer prior-training trajectories. At 100\%, CFAM reaches $74.6\%$, a $+22.4$\,pp advantage on this Stage-A evaluation, and the gap widens with data rather than saturating (from $+11.3$\,pp at 20\%).
These values should not be compared numerically with the Stage-B $D_\mathrm{test}$ results: $D_\mathrm{curve}$ and $D_\mathrm{test}$ are independent evaluation sets with different task and condition composition. On $D_\mathrm{test}$, the corresponding five-platform means are $56.1\%$ for $\pi_0$ and $65.8\%$ for CFAM, a $+9.7$\,pp gap (\Cref{tab:action_pathway}). Data efficiency compounds with the Stage-B and Stage-C gains reported below.

\subsubsection{Synthetic Build Results}

Across the two benchmark-level summaries, CFAM reaches $77.7\%$ on SimplerEnv ($\pi_0$ backbone) and $81.8\%$ on LIBERO (SpatialVLA backbone; this is the Full CFAM configuration of \Cref{tab:ablation}, with \textsc{corr} and \textsc{ext} capsules both written at Build and test-time writes disabled). Within SimplerEnv, performance is $76.4\%$ on Bridge and $78.9\%$ on Fractal. Relative to the frozen $\pi_0$ backbone, these are gains of $8.0$ and $7.5$ percentage points. The modification is the same in every row: one demonstration per task, segmented into one capsule per skill phase (\Cref{sec:capsule}), while the frozen backbone remains untouched.

\subsubsection{Real-World Build Results}

\paragraph{Per-task one-shot creation (Franka, Real General suite, public $\pi_0$ backbone).}
The one-shot write is measured on physical hardware on the Franka ten-task Real General suite with the public $\pi_0$ backbone: one task demonstration is segmented into its skill phases and encoded as one capsule per skill, with no gradient step.
Across the ten tasks, per-task gains range from $+10$\,pp (push, $72 \to 82$) to $+22$\,pp (insert, $36 \to 58$), a $+16.4$\,pp suite average ($53.8 \to 70.2$), with large improvements on the contact-rich stack and pour tasks ($+20$\,pp each) where the base model is systematically miscalibrated. The full per-task table is in the supplementary material.
Three Franka base/CFAM pairs appear in this paper and are three different measurements: this suite (public $\pi_0$, $53.8 \to 70.2$), the Real Arm column of \Cref{tab:action_pathway} (the in-house prior on $D_\mathrm{test}$, $68.6$, against matched-data baselines at $59.2$--$61.5$), and the Franka sequential testbed of \Cref{subsec:complementarity} (public $\pi_0$, $55.8 \to 71.6$). The in-house prior is a separate model from the public $\pi_0$ (\Cref{sec:exp:setup}); the two are never mixed within a table.
The gains come from both slices of the capsule: the perception-pathway studies in the supplementary material isolate the perception correction's independent contribution ($-6.1$\,pp when $\Delta\vz$ is zeroed on LIBERO) and show the two slices address genuinely distinct failure types.

\paragraph{Across the real-world datasets.}
\Cref{tab:datasets} widens the lens from the Franka suite to five platform-specific datasets. Each uses the same one-demonstration CFAM Build operation, but the embodiment, task family, action space, and test protocol differ.
\paragraph{In-house prior scale.} The in-house multimodal prior was trained on more than $2.6$ million trajectories across the five physical embodiments, $22$ task and site settings, and $79$ enumerated variations spanning manipulation, legged locomotion, aerial inspection, and off-road navigation (\Cref{tab:datasets}); these are prior-training trajectories, not per-task CFAM supervision---each robotic Build still uses one demonstration per evaluated task.
The resulting CFAM Build success on $D_\mathrm{test}$ ranges from $59.7\%$ on the quadruped to $71.5\%$ on the quadrotor inspection setting, with the contact-rich humanoid setting at $60.8\%$ and a $65.8\%$ cross-platform mean (\Cref{tab:action_pathway}); the post-deployment growth from this Build anchor is evaluated in \Cref{sec:exp:growth}.

\paragraph{Across five physical platforms.}
\label{sec:exp:cross_platform}
The same one-shot write generalizes across the five physical platforms (\Cref{tab:datasets}): CFAM leads the matched-data $\pi_0$ baseline on every one (+$5.8$ to +$13.0$\,pp) and the strongest matched-data baseline on every one (+$5.2$ to +$13.0$\,pp; per-platform values in \Cref{tab:action_pathway}).
A single demonstration written as a capsule, with the deployed weights untouched (preserving the non-interference guarantee of \Cref{sec:theory}), outperforms three standard policies trained on the full $D_\mathrm{train}$.
Qualitatively, the humanoid's gains concentrate on contact-rich manipulation where the in-house prior is most miscalibrated, and the quadruped's on unstable terrain (gravel, slip recovery) where a single demonstration of the recovery gait warps across the terrain family.
The load-bearing claim is that the \emph{capsule machinery is form-factor-agnostic}, not that skills are shared across form factors: the same GRT and the same write--grow--consolidate dynamics govern every library; only the embedding pipeline and action dimensionality change.

\paragraph{Component-level pathway comparisons.}
The Build evaluation decomposes CFAM into reasoning, perception, and the downstream action pathway. Detailed reasoning and perception pathway comparisons at Build are provided in \Cref{app:field_imagery} (\Cref{tab:reasoning_pathway,tab:perception_pathway}); the action pathway comparison is reported here as the downstream execution outcome.

\subsubsection{Build-Method Comparison}
\label{sec:exp:baselines}
\label{sec:exp:action_results}
\label{sec:exp_main_results}

The controlled build comparison is summarized by \Cref{tab:action_pathway}. On SimplerEnv, one-shot CFAM exceeds every public static policy on both splits with no per-task GPU adaptation (LoRA is compared on $D_\mathrm{var}$ in \Cref{tab:vla_adaptation}, not at Build). On the physical platforms, one-shot CFAM leads the strongest matched-data baseline by +$5.2$ to +$13.0$\,pp; the capsule representation converts one demonstration into a reusable, geometrically grounded unit of competence without touching the deployed weights. This subsection covers base competence building with standard supervised methods only; comparisons with post-deployment adaptation and memory methods are part of the field-learning analysis of \Cref{sec:exp:growth}.

\subsection{Autonomous Near-Edge Test-Time Growth}
\label{sec:exp:growth}

\noindent\textit{Evaluation question: after a few-shot supervised Build, can the deployed system autonomously expand competence from verified near-edge experience?}\par\smallskip

This subsection evaluates post-deployment growth separately in simulation and on physical systems. The synthetic stream measures whether forward transfer remains positive as five new environments arrive in sequence; the real-world stream measures the increase from the deployment-day Build anchor to the verified pre-novelty Grown endpoint across skill families, assets, and mission chains. Retention of earlier competence is analyzed separately in \Cref{sec:exp:continual}.

\begin{figure*}[t]
\centering
\captionof{table}{Master mission table: ten mission-critical scenarios (two per asset), each an ordered skill chain over $\rcufield$. \textbf{Build} = the supervised one-shot library at deployment; \textbf{Grown} = the pre-novelty endpoint after automatic capture of verified within-family extremes, i.e.\ after the whole Stage-C stream (Standard $=$ the static standard policy under the same mission protocol; $\dagger$ marks the mission plotted in \Cref{fig:build_ttgen_overview}, whose Build, $x{=}100$, and Grown, $x{=}160$, values are its anchors). Prefixes: L- $=$ LOCO-, M- $=$ MANIP-, REC $=$ recognize.}
\label{tab:mission_set}
\scriptsize
\setlength{\tabcolsep}{2.5pt}
\renewcommand{\arraystretch}{1.0}
\begin{tabularx}{\textwidth}{@{}lL{2.9cm}Xcccc@{}}
\toprule
\rowcolor{tabhead}
& & & \multicolumn{4}{c}{\textbf{Mission accuracy (\%)}} \\
\cmidrule(l){4-7}
\rowcolor{tabhead}
\textbf{Asset} & \textbf{Mission} & \textbf{Ordered skill chain} & \textbf{Standard Build} & \textbf{CFAM Build} & \textbf{Standard Grown} & \textbf{CFAM Grown} \\
\midrule
\multirow{2}{*}{Humanoid} & Hazardous-object recovery (firearm) & REC $\to$ L-legged-flat $\to$ M-reach $\to$ M-align $\to$ M-grasp $\to$ M-carry & 52.2 & 66.3 & 13.4 & 79.2 \\
 & Protective-gear donning (shield)$^\dagger$ & REC-gear $\to$ M-reach $\to$ M-grasp $\to$ M-don/wear $\to$ PA-posture-stance & 48.0 & 62.1 & 10.0 & 75.8 \\
\midrule
\rowcolor{tabzebra}
 & Metal-detector IED sweep & L-legged-rough $\to$ M-carry (payload) $\to$ REC-anomaly $\to$ PA-mark-target $\to$ L-slip-recovery & 47.8 & 68.9 & 6.1 & 77.5 \\
\rowcolor{tabzebra}
\multirow{-2}{*}{Robot Dog} & Confined-space structural recon$^\dagger$ & L-legged-rough $\to$ L-gap-cross $\to$ REC-hazard $\to$ PA-mark-target & 50.0 & 71.1 & 10.0 & 81.4 \\
\midrule
\multirow{2}{*}{Drone} & Perimeter recon \& track & L-aerial-hover $\to$ L-aerial-waypoint $\to$ REC-target $\to$ PA-loiter-track & 56.5 & 75.0 & 21.1 & 83.7 \\
 & Post-blast aerial damage survey$^\dagger$ & L-aerial-hover $\to$ L-aerial-waypoint $\to$ REC-damage $\to$ PA-orbit-inspect & 53.0 & 71.5 & 18.0 & 80.6 \\
\midrule
\rowcolor{tabzebra}
 & Precision part insertion (ordnance) & REC-part $\to$ M-reach $\to$ M-align $\to$ M-grasp $\to$ M-place $\to$ M-insert & 54.9 & 68.7 & 17.9 & 79.9 \\
\rowcolor{tabzebra}
\multirow{-2}{*}{Arm} & Render-safe / wire-cut (EOD)$^\dagger$ & REC-assembly $\to$ M-reach $\to$ M-align $\to$ M-grasp (fine) $\to$ M-cut & 53.0 & 66.8 & 15.0 & 77.0 \\
\midrule
\multirow{2}{*}{\shortstack[l]{Ground\\Vehicle}} & Off-road approach to objective & L-wheeled-terrain $\to$ REC-obstacle $\to$ route/avoid & 57.4 & 77.6 & 26.9 & 85.4 \\
 & Convoy follow \& checkpoint halt$^\dagger$ & L-wheeled-terrain $\to$ L-wheeled-follow $\to$ REC-checkpoint $\to$ M-precise-stop & 55.0 & 75.2 & 26.0 & 84.5 \\
\midrule
\rowcolor{tabhead}
\multicolumn{3}{@{}l}{\textbf{Overall}} & 52.8 & 70.3 & 16.4 & 80.5 \\
\bottomrule
\end{tabularx}
\end{figure*}

\begin{figure*}[t]
\centering
\includegraphics[width=\textwidth]{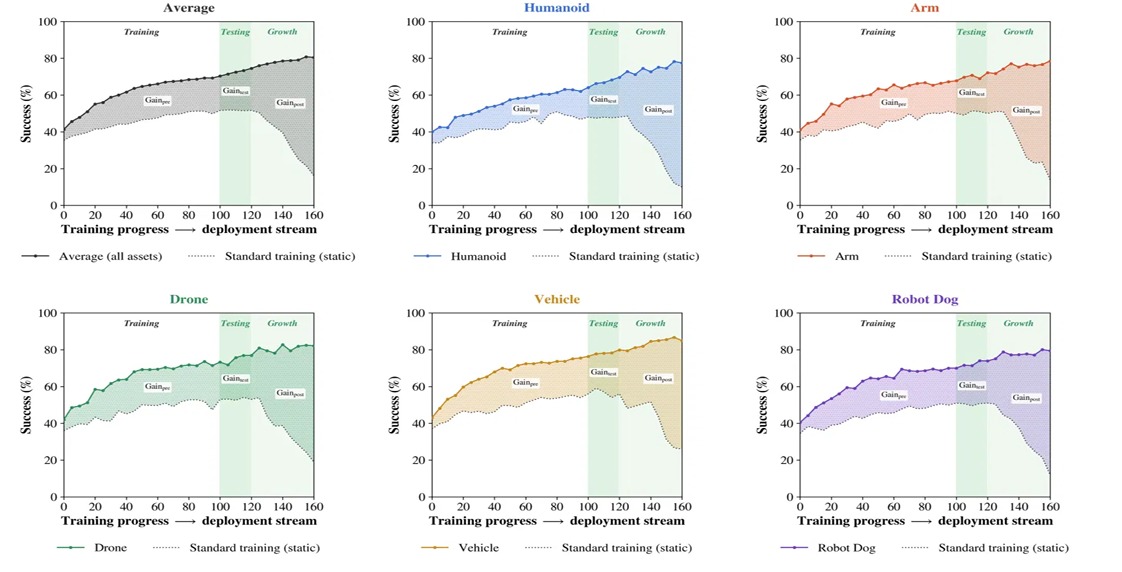}
\caption{\textbf{CFAM versus the standard policy across the deployment stream, per real-world embodiment and averaged.} Solid = CFAM; dotted = the standard (static) policy. In the \emph{Training} region ($0\text{--}100$, Stage~A) CFAM learns faster, rising above the standard policy at every fraction of $D_\mathrm{train}$. In the \emph{Testing} region ($100\text{--}120$, Stage~B) CFAM sustains that advantage on the held-out test. In the \emph{Growth} region ($120\text{--}160$, Stage~C) CFAM keeps growing as verified near-edge cases are captured, while the standard policy falls. The pattern holds on every embodiment and on the average; each panel shows one mission, marked $\dagger$ in \Cref{tab:mission_set}; its Build ($x{=}100$) and Grown ($x{=}160$) values are the anchors, and the Training region is the same mission scored at training checkpoints (Average: $70.3 \to 80.5$).}
\label{fig:build_ttgen_overview}
\end{figure*}

\subsubsection{Synthetic Growth Results}

CFAM maintains positive forward transfer in every synthetic environment ($+11.5$ to $+14.2$\,pp; $+12.8$\,pp on average; per-environment values in the FT columns of \Cref{tab:continual}), even as the stream introduces new objects, layouts, and task types. This is the synthetic growth result: the deployed memory adds useful competence at each stage without a gradient update to the frozen $\pi_0$ prior. \Cref{sec:exp:continual} uses the same stream to test whether those additions disturb earlier environments.

\subsubsection{Real-World Growth Results}

Growth and generalization are not only mechanisms of \Cref{sec:capsule,sec:execution}; on the near-edge deployment stream they are measurable results. At deployment the system warps stored skills onto the current scene. When confidence and retrieval distance mark a successful execution at the edge of a stored skill's support, the verified execution is written back as an extension capsule. This capture is autonomous because no operator selects, labels, or demonstrates the event and no gradient step changes the frozen prior; it is bounded because only verified near-edge extensions of known skill families are included in this paper.

Across the stream, the action side's per-skill success rises $74.0\to87.9$\% (\Cref{tab:real_growth_family}): execution rises as the warpable envelope expands. \Cref{tab:reasoning_pathway,tab:perception_pathway} report the reasoning and perception pathways at Build. The Build column of \Cref{tab:real_growth_family} is the one-shot library scored on $D_\mathrm{test}$ before growth, and the Grown column is the same library scored on $D_\mathrm{test}$ after the $D_\mathrm{var}$ stream has been traversed with writes enabled; at the skill-family level the pair therefore separates GRT generalization alone from GRT plus accumulated captures, on conditions the system never captured from.

\begin{table}[b]
\centering
\caption{Robotic growth by skill family on $D_\mathrm{var}$. Build = the one-shot library scored on $D_\mathrm{test}$ after Build, before growth; Grown = the same library scored on $D_\mathrm{test}$ after the $D_\mathrm{var}$ stream has been traversed with writes enabled (the verified pre-novelty endpoint). $\Delta$ is Grown minus Build success rate. Overall row is the unweighted mean across the five family rows.}
\label{tab:real_growth_family}
\small
\setlength{\tabcolsep}{5pt}
\renewcommand{\arraystretch}{1.1}
\rowcolors{2}{white}{tabzebra}
\begin{tabularx}{\columnwidth}{@{}>{\raggedright\arraybackslash}Xccc@{}}
\toprule
\rowcolor{tabhead}
\textbf{Skill family} & \textbf{Build (\%)} & \textbf{Grown (\%)} & \textbf{$\Delta$ (pp)} \\
\midrule
Legged locomotion & 74.4 & 86.8 & +12.4 \\
Aerial & 77.6 & 88.8 & +11.2 \\
Wheeled & 73.5 & 89.0 & +15.5 \\
Manipulation & 72.2 & 88.3 & +16.1 \\
Other perception--action & 72.5 & 86.6 & +14.1 \\
\midrule
\rowcolor{tabhead}
\textbf{Overall (unweighted mean)} & \textbf{74.0} & \textbf{87.9} & \textbf{+13.9} \\
\bottomrule
\end{tabularx}
\end{table}

\begin{table*}[b]
\centering
\caption{Comparison of adapted policies with CFAM after adaptation on the variation stream ($D_\mathrm{var}$); the metric is per-platform task success (\%) on $D_\mathrm{test}$, the same metric and split as \Cref{tab:action_pathway}. Each adaptation method receives the same $D_\mathrm{var}$ exposure under its own update mechanism (LoRA fine-tune, memory writes), and the comparison is head-to-head on the same split: CFAM leads the strongest adaptation baseline in every column, by $11.9$ to $20.8$\,pp. The CFAM row is above its own Build values in \Cref{tab:action_pathway} in every column, by $1.8$--$2.4$\,pp on the five real platforms and by $0.8$--$1.5$\,pp on the SimplerEnv splits, so the two tables are a before/after pair.}
\label{tab:vla_adaptation}
\footnotesize
\setlength{\tabcolsep}{4pt}
\renewcommand{\arraystretch}{1.1}
\rowcolors{2}{white}{tabzebra}
\begin{tabularx}{\textwidth}{@{}>{\raggedright\arraybackslash}Xccccccc@{}}
\toprule
\rowcolor{tabhead}
\textbf{Method} & \textbf{SE-Bridge} & \textbf{SE-Fractal} &
\textbf{Real Arm} & \textbf{Real Dog} &
\textbf{Real Humanoid} & \textbf{Real Drone} &
\textbf{Real Vehicle} \\
\midrule
\multicolumn{8}{@{}l}{\emph{Adaptation and memory baselines on $D_\mathrm{var}$}} \\
\textbf{CronusVLA} & 51.4 & 67.8 & 54.5 & 39.8 & 42.9 & 48.7 & 41.8 \\
\textbf{MemoryVLA} & 64.9 & 67.7 & 57.3 & 48.4 & 45.6 & 51.8 & 48.4 \\
\textbf{$\pi_0$ + LoRA} & 65.3 & 63.6 & 52.6 & 46.7 & 49.2 & 53.1 & 50.8 \\
\textbf{CogACT + LoRA} & 61.2 & 64.4 & 53.3 & 49.3 & 49.9 & 50.1 & 50.4 \\
\midrule
\textbf{CFAM (ours): 1-shot Build + test-time growth} & \textbf{77.9} & \textbf{79.7} & \textbf{70.8} & \textbf{61.9} & \textbf{63.2} & \textbf{73.9} & \textbf{70.3} \\
\bottomrule
\end{tabularx}
\end{table*}

\paragraph{Compression during growth.}
Consolidation carries performance weight as well as footprint: removing it costs $2.3$--$3.0$\,pp (\Cref{subsec:ablations}).

\paragraph{Mission-level reading.}
\Cref{tab:mission_set} reports the same arc at mission level (plotted per asset in \Cref{fig:build_ttgen_overview}): mean accuracy rises from $70.3$\% at Build to $80.5$\% at the pre-novelty Grown endpoint. The three Grown readings of this section are three different units and are reported separately rather than as nested readings of one measurement: per-skill success on $D_\mathrm{var}$ (\Cref{tab:real_growth_family}, $87.9$), per-platform task success on $D_\mathrm{var}$ after growth (\Cref{tab:vla_adaptation}, $68.0$ mean over the five real platforms), and per-mission accuracy (\Cref{tab:mission_set}, $80.5$); the three readings are therefore not nested and cannot be compared as a product of per-skill rates. The largest gains occur in the contact-rich arm and humanoid chains, matching manipulation's largest family-level gain in \Cref{tab:real_growth_family}; those families leave the greatest gap after one-shot build. Qualitatively, the field cases of \Cref{fig:field_deployment} illustrate the kind of near-OOD condition the stream contains (familiar objects in a new spatial arrangement, a heavier carried load, dense vegetation and a narrow passage on a grass-trained route, an unseen payload disturbance on a moving quadruped). They are slightly out-of-distribution changes inside an existing mission family, not unrelated tasks.

\subsubsection{Comparison with Post-Deployment Adaptation and Memory Methods}
\label{sec:exp:adaptation_methods}

The following comparison places CFAM's growth against the methods that also use information after deployment: gradient/adapter, memory-buffer, and retrieval-prompting approaches. Quoted literature results are not treated as matched controlled comparisons.

For a deployed policy that must improve after it ships, there are two natural architecture alternatives: the \emph{gradient/adapter route} (write the new competence into the weights: LoRA, full fine-tuning) and the \emph{memory-buffer route} (attach an episodic store the policy reads at inference: MemoryVLA, CronusVLA).
\Cref{tab:vla_adaptation} compares CFAM with VLA augmentation and adaptation approaches on the variation stream $D_\mathrm{var}$.
Against the memory-augmented systems on $D_\mathrm{var}$, CFAM leads on both SimplerEnv splits ($+12.0$ to $+13.0$\,pp over MemoryVLA, $+11.9$ to $+26.5$\,pp over CronusVLA) and exceeds MemoryVLA on the pooled robotic suite by $+13.5$ to $+22.1$\,pp per platform (\Cref{tab:vla_adaptation}). We attribute this performance to the regime-governed capsule structure: the two-regime memory organization enables qualitatively different storage for corrections and extensions.
 Against gradient-based adaptation, CFAM exceeds LoRA on both SimplerEnv splits and every robotic platform while requiring \emph{no retraining of the frozen model}; the demonstration budgets differ only at Build (one demonstration per task for CFAM against ten per task for the LoRA fine-tune), after which every method receives the same $D_\mathrm{var}$ exposure.
Neither alternative keeps learning: the gradient route stops when the fine-tuning budget is spent (and forgets when it resumes, \Cref{sec:exp:continual}), and the buffer route accumulates entries without consolidating them, whereas the capsule library keeps growing at test time.

\paragraph{Retrieval-prompting alternatives: RAG and failure-state prompting.}
A third alternative uses no new machinery at all: keep the frozen policy and its Reasoning VLM, and put the adaptation burden on the prompt. These alternatives are discussed qualitatively and are not scored in \Cref{tab:vla_adaptation}. \emph{VLM failure-prompt} feeds the failure state back to the Reasoning VLM at the next attempt with no persistent store; \emph{RAG} persists every failed-then-corrected episode as a retrievable document and prepends the top-$k$ matches at inference.
The failure modes are informative.
First, corrections must round-trip through language: continuous geometric detail (a grasp offset of a few centimeters, a wrist angle of a few degrees) is lost in verbalization, which is precisely the content a capsule stores as a numeric residual and GRT re-applies geometrically.
Second, retrieval-prompting requires a VLM decode on \emph{every adaptation step}, whereas CFAM invokes the Reasoning model once per skill phase and not at all per correction. This makes the capsule path less dependent on repeated language-model inference during adaptation.
In capsule terms (\Cref{sec:capsule}), RAG is a CC with its perception and action slices collapsed to text: the situation key survives as embedding similarity, but the numeric correction payload, bounded authority, and consolidation are lost.

\paragraph{Absolute performance context.}
CFAM's SimplerEnv-Fractal result is competitive with recent gradient-retrained policies on the same suite despite performing no retraining: representative reported values are $60.5\%$ for Dream-VLA~\cite{dreamvla2025} and $63.0\%$ for OpenVLA-OFT~\cite{openvlaoft2025} under visual matching.
We do not claim a like-for-like win (protocols and fine-tuning data differ across these reports, and our figures come from the configuration of \Cref{sec:exp:setup}), but the gradient-free operating point is not paid for in absolute success on this suite.
The architectural cost of refusing to update the base is instead visible in the ceiling of \Cref{tab:real_growth_family}: one-shot Build reaches roughly $84.2\%$ of the Grown endpoint ($74.0/87.9$), and test-time growth closes the remainder.

\newcommand{\CapTabContinual}{\textbf{Continual adaptation across five sequential simulated environments.} SimplerEnv task families, $\pi_0$ backbone; mean over $\ConfigValue{eval.seeds-sim}$ seeds. FT = forward transfer; BT = backward transfer; Forget = prior tasks with $>5\%$ success drop.}
\newcommand{\CapFigRetention}{\textbf{Retention of previously learned tasks across the sequential suite.} The curve shows competence retained after each successive environment (dashed: perfect retention). CFAM remains within $0.6$\,pp of perfect retention, while MemVLA drifts to $94.3\%$ and LoRA falls to $81.8\%$.}
\newcommand{\TabContinual}{%
\begin{tabular}{@{}lcccccc@{}}
\toprule
\rowcolor{tabhead}
& \multicolumn{2}{c}{\textbf{CFAM}} & \multicolumn{2}{c}{\textbf{LoRA}} & \multicolumn{2}{c}{\textbf{MemVLA}} \\
\cmidrule(lr){2-3}\cmidrule(lr){4-5}\cmidrule(lr){6-7}
\rowcolor{tabhead}
\textbf{Stage} & \textbf{FT} & \textbf{BT} & \textbf{FT} & \textbf{BT} & \textbf{FT} & \textbf{BT} \\
\midrule
Env 1 & +14.2 & --- & +16.8 & --- & +8.3 & --- \\
Env 2 & +12.8 & $-$0.3 & +14.1 & $-$4.7 & +7.1 & $-$1.2 \\
Env 3 & +11.5 & $-$0.5 & +11.3 & $-$8.9 & +6.4 & $-$2.8 \\
Env 4 & +13.1 & $-$0.4 & +9.2 & $-$13.6 & +5.8 & $-$4.1 \\
Env 5 & +12.4 & $-$0.6 & +7.8 & $-$18.2 & +5.1 & $-$5.7 \\
\midrule
\rowcolor{tabhead}
\textbf{Avg} & \textbf{+12.8} & \textbf{$-$0.5} & +11.8 & $-$11.4 & +6.5 & $-$3.5 \\
\rowcolor{tabhead}
\textbf{Forget} & \multicolumn{2}{c}{\textbf{2.1\%}} & \multicolumn{2}{c}{34.7\%} & \multicolumn{2}{c}{10.8\%} \\
\bottomrule
\end{tabular}}
\iflatexml
\begin{table}[t]
\caption{\CapTabContinual}
\label{tab:continual}
\centering
\footnotesize
\setlength{\tabcolsep}{4.2pt}
\renewcommand{\arraystretch}{1.02}
\rowcolors{3}{white}{tabzebra}
\TabContinual
\end{table}
\begin{figure}[t]
\centering
\includegraphics[width=0.5\linewidth]{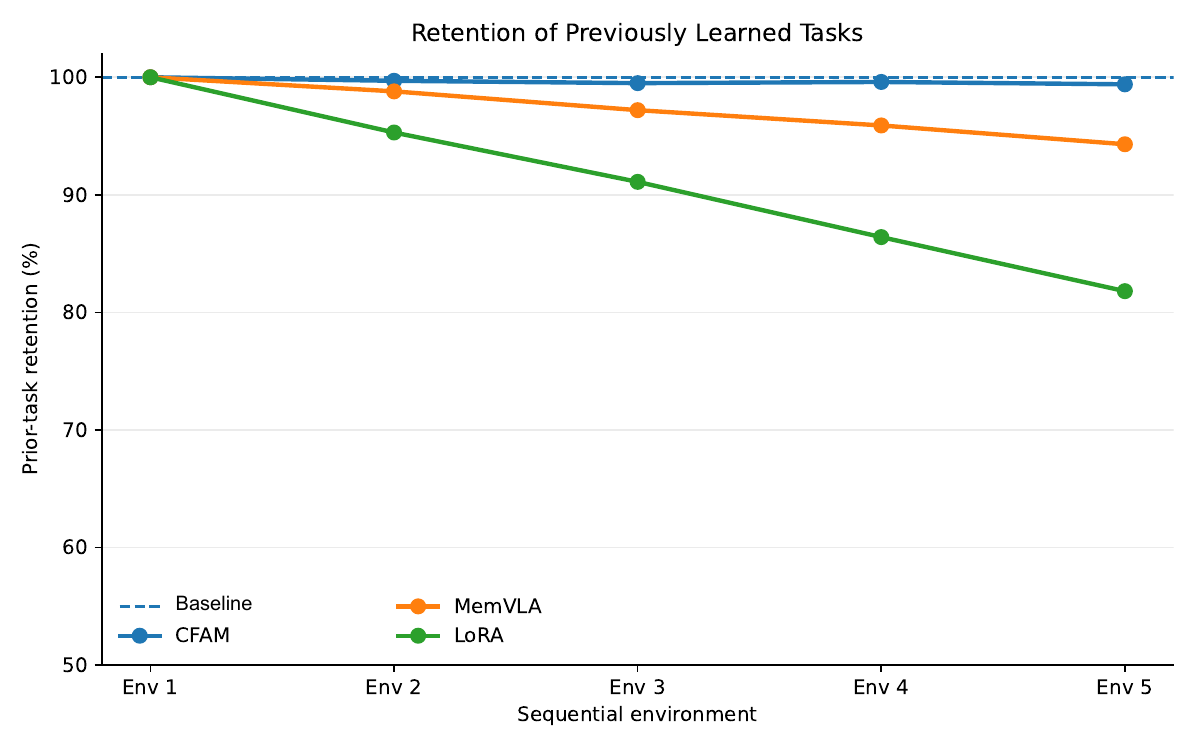}
\caption{\CapFigRetention}
\label{fig:retention}
\end{figure}
\else
\begin{figure*}[t]
\begin{minipage}[t]{0.49\textwidth}
\captionsetup{type=table}
\caption{\CapTabContinual}
\label{tab:continual}
\centering
\footnotesize
\setlength{\tabcolsep}{4.2pt}
\renewcommand{\arraystretch}{1.02}
\rowcolors{3}{white}{tabzebra}
\TabContinual
\end{minipage}\hfill
\begin{minipage}[t]{0.49\textwidth}
\captionsetup{type=figure}
\centering
\includegraphics[width=0.96\linewidth]{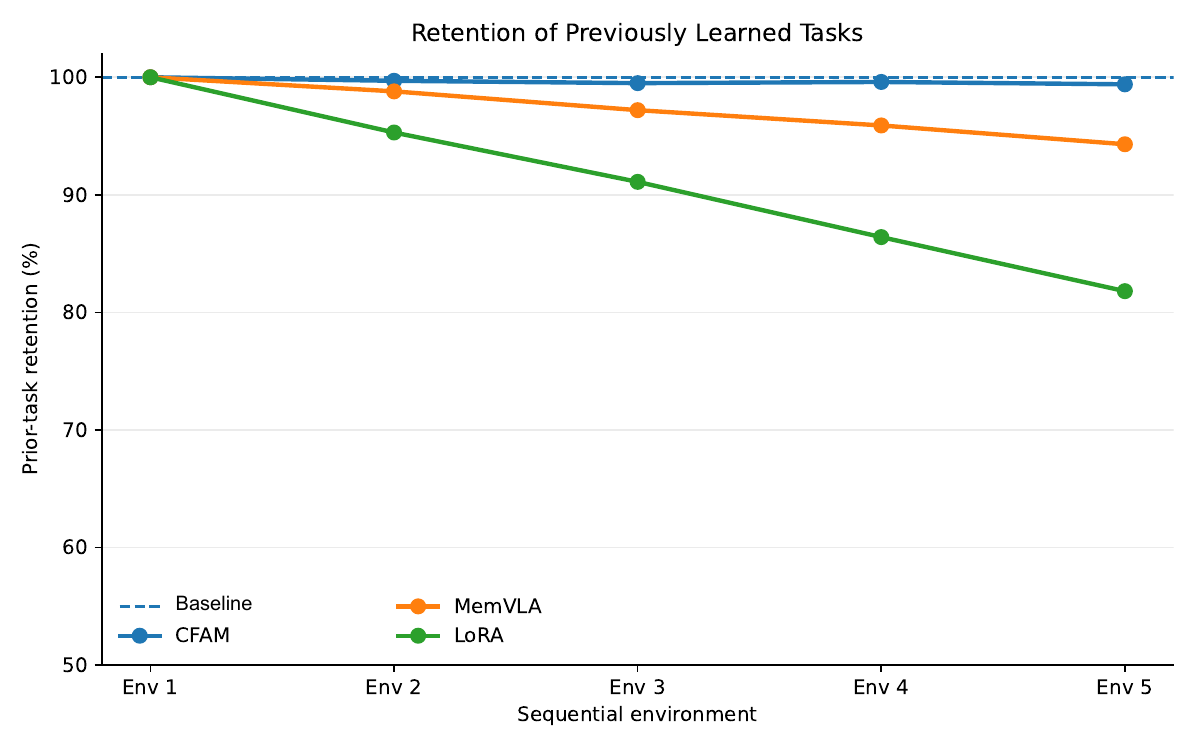}
\caption{\CapFigRetention}
\label{fig:retention}
\end{minipage}
\end{figure*}
\fi

\subsection{Growth Without Forgetting}
\label{sec:exp:continual}

\noindent\textit{Evaluation question: does adding new competence preserve performance on earlier environments?}\par\smallskip

The central architectural claim, that the memory grows without disturbing what the base or the earlier memory already holds, is tested head-on by sequential-environment adaptation: five simulated environments constructed from the SimplerEnv task families~\cite{simplerenv2024} ($\pi_0$ backbone) are encountered in succession, each introducing new objects and layouts within the known task families relative to its predecessors; after adapting in each, the agent is re-evaluated on all earlier ones.
\Cref{tab:continual} shows the result (\Cref{fig:retention} plots the retained fraction after each stage): near-zero backward transfer ($-0.5$\,pp) and minimal forgetting (only $2.1\%$ of prior tasks lose more than $5$\,pp) across five sequential environments, consistent with the locality guarantee and the no-forgetting invariant (both stated and proved in the supplementary material). LoRA suffers catastrophic forgetting ($34.7\%$ of prior tasks lose more than $5$\,pp; BT $-18.2$\,pp by Env 5) because it modifies shared parameters. CFAM's forward transfer remains stable ($+11.5$ to $+14.2$\,pp), with no measurable degradation attributable to saturation of the fixed-budget capsule field $\rcufield$ over this five-environment horizon. Retention is measured in the sequential simulation suite.

\paragraph{The no-interference principle, in its simplest form.}
The near-zero forgetting above is not peculiar to CFAM's machinery; it follows from a structural property CFAM shares with the simplest frozen-backbone classifier, nearest-class-mean (NCM)~\cite{mensink2013distance,rebuffi2017icarl,janson2022simple}: because each stored unit occupies a separate slot and enrolling a new one changes no existing parameters, prior knowledge is preserved by construction rather than by a regularizer.
CFAM generalizes this primitive along the axes a deployed policy needs---a capsule stores a per-situation action and perception correction rather than a per-class mean, is addressed over a support radius, weighted by confidence, typed by regime, warped by GRT, and consolidated under a fixed budget---while NCM is the degenerate case (one passive, radius-zero capsule per class, no warping, no growth). The $-0.5$\,pp backward transfer of \Cref{tab:continual} is this principle measured on a sequential suite.

\subsection{Ablations}
\label{subsec:ablations}
\begin{table*}[t]
\centering
\caption{Component ablations on LIBERO (SpatialVLA backbone, static-set evaluation). Mean and SD are over three seeds, per suite as Spatial/Object/Goal/Long success rate (\%). $\Delta$ is the change in the four-suite average against Full CFAM, the reference row; negative is worse. The blocks isolate the pre-novelty regimes (\textsc{corr}, \textsc{ext}), swap one architectural component at a time (storage, field controller, retrieval, GRT, capsule dynamics), and remove consolidation.}
\label{tab:ablation}
\footnotesize
\setlength{\tabcolsep}{6pt}
\renewcommand{\arraystretch}{1.05}
\rowcolors{2}{white}{tabzebra}
\begin{tabularx}{\textwidth}{@{}>{\raggedright\arraybackslash}XC{4.0cm}C{3.7cm}r@{}}
\toprule
\rowcolor{tabhead}
\textbf{Configuration} & \textbf{Mean} & \textbf{SD} & \textbf{$\Delta$} \\
\midrule
Base only & 74.1/69.8/67.5/51.2 & 1.4/1.6/1.7/2.3 & $-16.1$ \\
\textsc{corr} only & 81.7/76.5/74.2/57.8 & 1.2/1.4/1.5/1.9 & $-9.2$ \\
\textsc{ext} only & 79.4/75.1/72.8/56.4 & 1.3/1.5/1.6/2.0 & $-10.9$ \\
Full CFAM & \textbf{87.9/85.2/83.1/70.9} & 1.0/1.1/1.3/1.5 & --- \\
\midrule
Dense memory & 83.1/80.4/78.3/65.2 & 1.2/1.3/1.5/1.7 & $-5.0$ \\
No field controller & 80.2/77.1/74.8/61.8 & 1.4/1.5/1.7/1.9 & $-8.3$ \\
Nearest-neighbor action & 81.6/79.0/76.5/63.7 & 1.3/1.4/1.6/1.8 & $-6.6$ \\
GRT only & 78.4/74.5/72.1/56.8 & 1.3/1.5/1.6/2.0 & $-11.3$ \\
CC dynamics only & 82.7/79.8/77.2/62.4 & 1.2/1.3/1.5/1.8 & $-6.3$ \\
No intra-regime consolidation & 85.2/82.4/80.1/67.3 & 1.1/1.2/1.4/1.6 & $-3.0$ \\
No cross-regime consolidation & 85.8/83.1/80.9/68.0 & 1.1/1.2/1.4/1.6 & $-2.3$ \\
\bottomrule
\end{tabularx}
\end{table*}

\noindent\textit{Evaluation question: which memory, control, geometric, and consolidation components account for the observed gains?}\par\smallskip

Ablations use the SpatialVLA backbone and report average success rate across the four LIBERO suites under conventional static-set evaluation.

\paragraph{Per-regime and architectural ablation.}
\Cref{tab:ablation} isolates each component and reports per-suite numbers.
Each pre-novelty regime contributes independently: corrective $+6.9$\,pp, extension $+5.3$\,pp (Full CFAM $=$ \textsc{corr} $+$ \textsc{ext}; the \textsc{corr} capsules are written at Build from the supplied demonstrations, as in every experiment here, and no novelty-regime capsules are used anywhere in this paper). The full combination ($+16.1$\,pp over base) exceeds the sum of the two isolated contributions ($6.9 + 5.3 = 12.2$\,pp), indicating an interaction between the regimes: a correction capsule and an extension capsule active on the same query recover cases that neither recovers alone.
Architecturally, the capsule-field controller is the most load-bearing of the component swaps (removing it costs $8.3$\,pp). Dense storage costs $5.0$\,pp. GRT and capsule dynamics are complementary ($+4.8$\,pp and $+9.9$\,pp alone, $+16.1$\,pp together).

\paragraph{Consolidation ablation.}
\label{subsec:consolidation_analysis}
Removing intra-regime consolidation, merging of redundant capsules \emph{within} a regime, costs $3.0$\,pp, and removing cross-regime consolidation costs $2.3$\,pp (\Cref{tab:ablation}). Without consolidation, capsules accumulate redundancy and retrieval signal-to-noise degrades: compression is not only a footprint mechanism but a performance one.

\paragraph{Complementarity with fine-tuning.}
\label{subsec:finetuned}
\label{subsec:complementarity}
\label{sec:positioning}
A natural concern is that CFAM's gains vanish as the base model improves.
We test this by layering CFAM atop a \emph{fine-tuned} base (LoRA, rank 32, $\alpha = 32$) on two testbeds: SpatialVLA on LIBERO-Long, and the Franka sequential tabletop testbed with the public $\pi_0$ backbone (a different suite and backbone from the in-house-prior Real Arm column of \Cref{tab:action_pathway} and from the Real General suite of \Cref{sec:exp:franka}).
LoRA + CFAM reaches $74.3 \pm 1.3\%$ on LIBERO-Long (SpatialVLA) and $75.1 \pm 1.5\%$ on the real-robot testbed ($\pi_0$, Franka), outperforming both LoRA alone ($64.7$/$65.2\%$, i.e.\ $+9.6$/$+9.9$\,pp) and frozen + CFAM ($70.9$/$71.6\%$, i.e.\ $+3.4$/$+3.5$\,pp), from frozen bases of $51.2$/$55.8\%$ ($\ConfigValue{eval.seeds-sim}$ seeds).
The gain from adding CFAM on top of LoRA is smaller than on the frozen model. This is expected: as the base model improves, fewer errors remain for the memory to correct; the remaining gains come from extension-regime capsules that address gaps LoRA cannot fill.
Crucially, the LoRA + CFAM configuration requires no additional GPU compute beyond LoRA's initial fine-tuning: capsule dynamics are gradient-free and run entirely at inference time.
Practitioners need not choose between fine-tuning and CFAM: a robot can be fine-tuned once in the lab and then equipped with the memory for growth.

\section{Discussion}
\label{sec:discussion}

The central result is a separation between the competence installed when a physical AI system is built and the competence it can acquire after deployment: matched-data efficiency and leadership at Build, autonomous growth on the variation stream while adaptation baselines fall behind, and near-perfect retention in the sequential simulation suite (\Cref{sec:experiments}).

\subsection{What Test-Time Growth Means for Physical AI}

CFAM does not keep training the frozen foundation model after deployment. It expands a separate capsule field whose entries are local, inspectable perception--action corrections. In the measured pre-novelty streams, confidence and retrieval distance identify a successful execution at the edge of a stored skill's support, and the physical phase objective verifies that success before the execution is written back. The result is a larger reusable envelope without a gradient step on the deployed base. The rise from the Build to Grown anchors therefore measures added post-deployment competence, not merely repeated inference from a fixed library.

This is \emph{autonomous learning from near-OOD cases at test time}. The Build library is few-shot and supervised: each installed task begins with a supplied demonstration. After that initialization, however, no operator chooses or labels the successful cases that are captured; the deployed system identifies, verifies, and stores them from its own execution signals on-device.

\paragraph{What a learned near-OOD case looks like.}
The difference is small enough to preserve the task's structure but large enough to sit outside the Build examples: a new bag--box arrangement forcing a different maneuver to reach the shield, a heavier shield in the same reach--grasp--carry family, a grass-trained route meeting dense weeds and a narrow opening, a moving gait meeting an unseen impulsive payload disturbance (\Cref{fig:field_deployment}). CFAM learns these slightly out-of-distribution combinations by extending an applicable prior skill; they do not imply autonomous weapon selection or engagement.

The build comparisons and the growth results answer different questions: what one demonstration buys at deployment relative to the matched-data and adaptation alternatives, and whether that one-shot operating point remains fixed. Growth is bounded by the skill families the library already holds.

\subsection{Limitations and Scope}
\label{subsec:limitations}

\paragraph{Evidence boundary and duration.}
The evidence ends at the defined near-edge boundary: known skill families remain relevant throughout. Autonomous capture has been demonstrated only on the evaluated hardening streams and within the measured horizon; longer-duration and open-world deployment remain untested. Open-world novelty injection, failure detection, correction from detected field failures, field-time perception writes, and language-to-trajectory synthesis are outside this paper's scope and are not evaluated here. Several quantities that bear on the claims are not measured here either. \emph{Data and controls:} the in-house prior with an empty capsule field, the same prior fine-tuned on the single Build demonstration, and the same demonstration exploited by trajectory replay or nearest-neighbour retrieval without capsules, the controls that would isolate the capsule representation's share of the Build gain; a per-platform Stage-C run with test-time writes disabled (the skill-family Build column of \Cref{tab:real_growth_family} is the only such control) and a held-out post-growth split of unseen conditions; the reachable envelope as a function of the number of captures; growth of the perception and reasoning pathways along the stream; the number of unique skills per platform; the size and composition of $D_\mathrm{curve}$; the per-platform $D_\mathrm{var}$ denominators and condition allocation; and the simulated (Isaac Sim) share of $D_\mathrm{train}$ per platform. \emph{Statistics:} per-seed dispersion, confidence intervals, and paired tests on the reported rates; and the number of distinct $\Phi$ predicate instances and their false-write and false-abstention rates. \emph{Deployment footprint:} the footprint behind R3 and R4 (edge device per platform, whether the Reasoning cortex ran on the asset or off-board during the quadrotor and quadruped trials, GRT, retrieval, and write latencies against the control period, capsule size and counts before and after $D_\mathrm{var}$, memory budget and occupancy, and consolidation merges and evictions during Stage~C). \emph{Protocol:} the mission trial counts, retry rules, and scoring of \Cref{tab:mission_set}; whether the SpatialVLA base of \Cref{tab:ablation} was LIBERO-fine-tuned; and the starting checkpoints, parameter counts, optimization budgets, and tuning used to train $\pi_0$, CogACT, and SpatialVLA on the in-house dataset and to adapt MemoryVLA and CronusVLA to the non-manipulation embodiments. The architectural argument for R4 is that a write is one forward pass and one memory insertion, gradient-free; measurements of this write path in on-field deployments are future work, so the deployment claim here is stated as a property of the design.

\paragraph{Supervision and write quality.}
Autonomous near-edge extension is not open-world discovery. Build requires supplied demonstrations, and failed executions with no usable prior skill require an available corrective or successful trajectory. Moreover, the confidence and retrieval signals used to route an event are not correctness guarantees. A confidently wrong base can suppress the fallback, false-write and false-abstention rates are not yet calibrated, and a persistent wrong capsule can be reused. The independent phase objective is the principal backstop, but it verifies an execution after emission rather than guaranteeing it beforehand.

\paragraph{Frozen-base and representation limits.}
CFAM adapts around the frozen policy; it cannot repair every deficiency inside that policy. If the encoder maps physically different states to indistinguishable addresses, or if an action lies outside the base system's sensing, actuation, and executable skill family, local residuals cannot create the missing capability. The formal properties bound authority, locality, and convergence under stated conditions; they do not guarantee task success.

\paragraph{Geometry and morphology.}
GRT depends on reliable correspondences. Occluded or mismatched anchors degrade the candidate transform, while rigid and affine warps are poorly suited to strongly deformable objects. Cross-morphology transfer requires a compatible action representation and is not established here.

\paragraph{Retention and capacity.}
The retention evidence is simulation-only (\Cref{tab:continual}); the physical platforms have not been re-evaluated on $D_\mathrm{test}$ after Stage-C growth, so real-platform retention is untested. The no-interference result applies to disjoint supports left untouched. Consolidation can merge nearby capsules and therefore trades redundancy against fidelity. The memory budget is finite, and a sufficiently long or diverse deployment will eventually require eviction or a larger store; which knowledge should be retained under saturation remains an open systems and governance question.

\paragraph{Evaluation breadth.}
The present experiments cover tabletop manipulation, sequential simulated environments, ten mission chains, and five physical platforms, but they do not establish generality across coordinated two-handed manipulation (the humanoid tasks of \Cref{fig:lab_training,fig:mission_chain} use one hand at a time), deformable-object handling, or multi-room navigation. These results should be read as evidence for the CFAM growth primitive and its Build $\to$ Grow $\to$ Retain sequence, not as a complete demonstration of autonomous lifelong learning.
\section{Conclusion}
\label{sec:conclusion}

CFAM demonstrates that the competence of a physical AI system need not be fixed at deployment. Across four ordered stages on Skylark's in-house multi-embodiment dataset with matched-data baselines, CFAM reaches the standard policy's operating point with $2.5\times$ less prior-training data, leads every matched-data baseline on the held-out split, autonomously captures verified slightly out-of-distribution cases on-device, raising action success by $13.9$ percentage points while adaptation baselines fall behind, and, in the sequential simulation suite, retains earlier competence, with backward transfer of $-0.5$ percentage points versus $-11.4$ percentage points for LoRA.

The result does not depend on continual gradient retraining of the frozen base. New competence is written into bounded, inspectable perception--action capsules and reused through geometric warping. This gives post-deployment learning a localized form: the system can add coverage while preserving the shared substrate and the provenance of each addition.

The present evidence establishes autonomous near-edge extension, not open-world discovery; open-ended novelty, longer-duration field learning, and cross-morphology generalization remain future work (\Cref{subsec:limitations}).

Today's physical AI systems largely stop learning when training ends. CFAM provides evidence that they can instead continue acquiring bounded competence post-deployment.

\bibliographystyle{ieeetr}
{\fontsize{6.0}{6.45}\selectfont%
\makeatletter
\renewcommand{\@openbib@code}{\setlength{\itemsep}{0pt}\setlength{\parsep}{0pt}}
\makeatother
\bibliography{references}}

\clearpage
\onecolumn
\appendix
\section{Field Deployment Imagery and Pathway Comparisons}
\label{app:field_imagery}
The two ground platforms of the pre-novelty mission set (\Cref{tab:mission_set}) are shown during field trials, providing the embodiment context for the slightly out-of-distribution cases discussed in \Cref{sec:discussion}. The reasoning- and perception-pathway comparisons referenced from \Cref{subsec:one_shot} follow (\Cref{tab:reasoning_pathway,tab:perception_pathway}).
\iflatexml
\begin{figure}[t]
\centering
\begin{tabular}{@{}p{0.48\textwidth}@{\hspace{0.03\textwidth}}p{0.48\textwidth}@{}}
\includegraphics[width=0.48\linewidth]{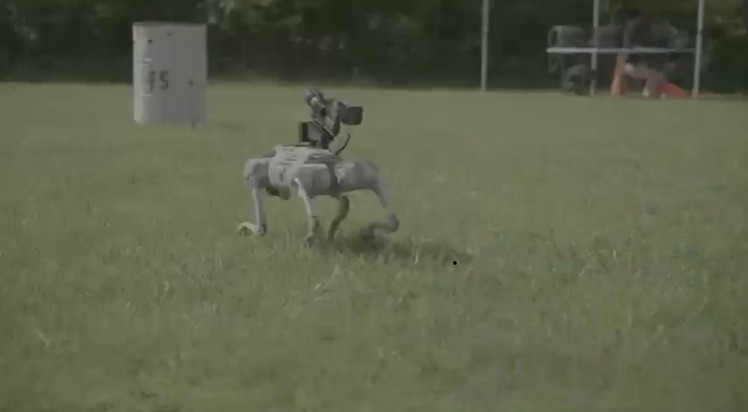}\hfill
\includegraphics[width=0.48\linewidth]{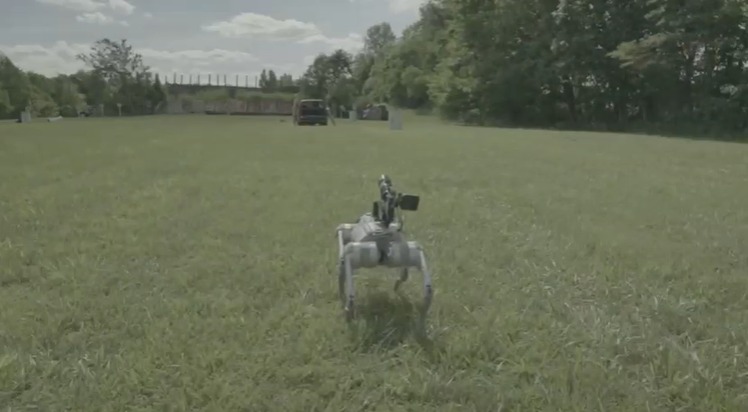}\par\smallskip
\small \textbf{(a) Robot Dog mission.} Uneven-terrain traversal with a moving payload. &
\includegraphics[width=0.48\linewidth]{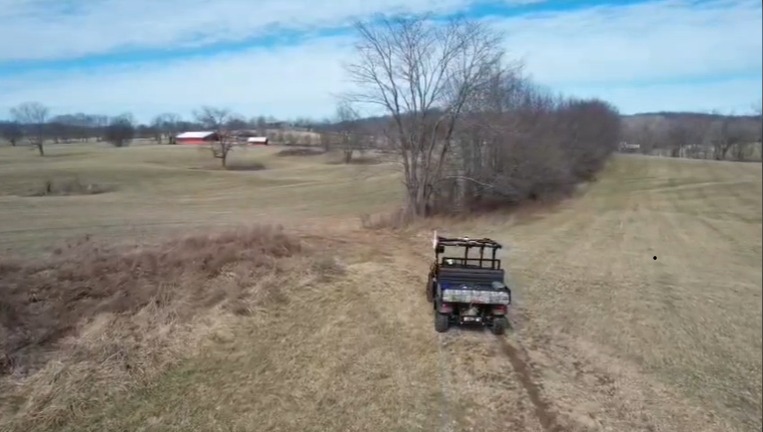}\hfill
\includegraphics[width=0.48\linewidth]{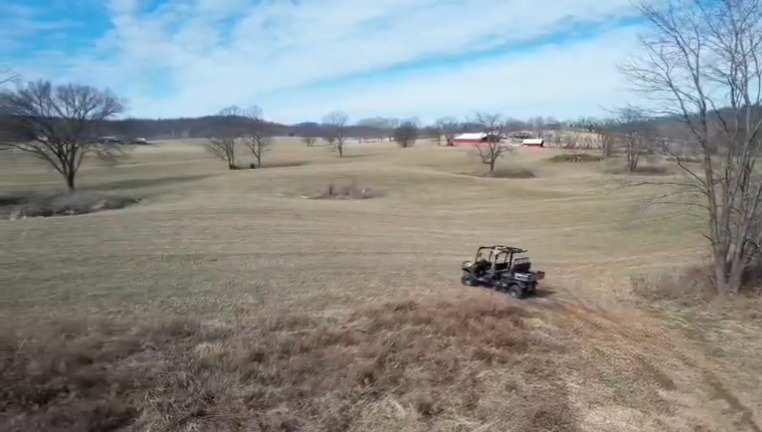}\par\smallskip
\small \textbf{(b) Ground Vehicle mission.} First and final views of the off-road passage.
\end{tabular}
\caption{\textbf{Field deployment imagery and concrete near-OOD cases.} \textbf{(a) Robot Dog:} the legged platform provides the embodiment for extending a moving gait to an unseen impulsive or recoil-like payload disturbance while preserving stability. \textbf{(b) Ground Vehicle:} a route learned on open grass is extended when dense weeds leave only a narrow opening. These are new, slightly out-of-distribution conditions inside known skill families; the system identifies, verifies, and stores the successful extension autonomously on-device.}\label{fig:field_deployment}
\end{figure}
\else
\begin{center}
\centering
\begin{minipage}[t]{0.492\textwidth}
\centering
\includegraphics[width=0.492\linewidth,height=2.6cm,keepaspectratio]{image_assets/field_dog_2.jpg}\hfill
\includegraphics[width=0.492\linewidth,height=2.6cm,keepaspectratio]{image_assets/field_dog_1.jpg}
\smallskip\par\small\textbf{(a) Robot Dog mission.} Uneven-terrain traversal with a moving payload.
\end{minipage}\hfill
\begin{minipage}[t]{0.492\textwidth}
\centering
\includegraphics[width=0.492\linewidth,height=2.6cm,keepaspectratio]{image_assets/field_vehicle_1.jpg}\hfill
\includegraphics[width=0.492\linewidth,height=2.6cm,keepaspectratio]{image_assets/field_vehicle_3.jpg}
\smallskip\par\small\textbf{(b) Ground Vehicle mission.} First and final views of the off-road passage.
\end{minipage}
\captionof{figure}{\textbf{Field deployment imagery and concrete near-OOD cases.} \textbf{(a) Robot Dog:} the legged platform provides the embodiment for extending a moving gait to an unseen impulsive or recoil-like payload disturbance while preserving stability. \textbf{(b) Ground Vehicle:} a route learned on open grass is extended when dense weeds leave only a narrow opening. These are new, slightly out-of-distribution conditions inside known skill families; the system identifies, verifies, and stores the successful extension autonomously on-device.}\label{fig:field_deployment}
\end{center}
\fi

\vspace{2pt}
\centering
\captionof{table}{Reasoning pathway comparison across simulated and real-robot suites. Success rate (\%) is reported for the reasoning pathway. The CFAM row is the same frozen Qwen2.5-VL-7B as the second row, operating inside the CFAM stack and prompted with the capsule descriptors of \Cref{eq:rcu_full} (\Cref{sec:reasoning_module}); the difference between the two rows is what the descriptors add, not a different model. Baselines: GPT-4o~\cite{openai2024gpt4o}, Qwen2.5-VL-7B~\cite{bai2025qwen25vl}, Gemini Robotics-ER~\cite{geminirobotics2025}.}
\label{tab:reasoning_pathway}
\footnotesize
\renewcommand{\arraystretch}{1.08}
\setlength{\tabcolsep}{3pt}
\rowcolors{2}{white}{tabzebra}
\begin{tabular}{@{}lccccccc@{}}
\toprule
\rowcolor{tabhead}
\textbf{Method} & \textbf{SE-Bridge} & \textbf{SE-Fractal} & \textbf{Real Arm} & \textbf{Real Dog} & \textbf{Real Humanoid} & \textbf{Real Drone} & \textbf{Real Vehicle} \\
\midrule
GPT-4o & 38.8 & 35.4 & 35.2 & 36.5 & 39.6 & 25.6 & 31.3 \\
Qwen2.5-VL-7B & 45.3 & 42.8 & 38.5 & 39.2 & 32.6 & 29.5 & 34.5 \\
Gemini Robotics-ER & 39.3 & 36.5 & 31.5 & 33.6 & 35.6 & 27.1 & 32.5 \\
\textbf{CFAM Reasoning (Ours; Qwen2.5-VL-7B in stack)} & \textbf{54.3} & \textbf{58.8} & \textbf{45.6} & \textbf{48.7} & \textbf{43.7} & \textbf{46.9} & \textbf{43.6} \\
\bottomrule
\end{tabular}

\vspace{2pt}
\centering
\captionof{table}{Perception pathway comparison across simulated and real-robot suites. Success rate (\%) is reported for the visual representation used by the downstream action policy. The CFAM row is the in-domain-trained SHDL Sensor cortex, whereas the baselines are frozen generic encoders. Baselines: CLIP~\cite{radford2021clip}, DINOv2~\cite{oquab2023dinov2}, R3M~\cite{nair2022r3m}, MoCo-v3~\cite{chen2021mocov3}, ViT-IN (ImageNet-supervised ViT~\cite{dosovitskiy2020vit}).}
\label{tab:perception_pathway}
\footnotesize
\renewcommand{\arraystretch}{1.05}
\setlength{\tabcolsep}{3.5pt}
\rowcolors{2}{white}{tabzebra}
\begin{tabular}{@{}lccccccc@{}}
\toprule
\rowcolor{tabhead}
\textbf{Perception Method} & \textbf{SE-Bridge} & \textbf{SE-Fractal} & \textbf{Real Arm} & \textbf{Real Dog} & \textbf{Real Humanoid} & \textbf{Real Drone} & \textbf{Real Vehicle} \\
\midrule
CLIP & 61.8 & 68.4 & 61.1 & 43.1 & 46.5 & 47.4 & 42.2 \\
DINOv2 & 64.7 & 71.2 & 64.3 & 44.8 & 42.3 & 48.7 & 44.3 \\
R3M & 66.9 & 73.5 & 66.7 & 45.1 & 48.2 & 50.8 & 45.6 \\
MoCo-v3 & 59.8 & 67.1 & 58.3 & 42.3 & 41.1 & 46.3 & 40.2 \\
ViT-IN & 58.6 & 65.9 & 65.4 & 43.7 & 45.9 & 46.8 & 42.3 \\
\textbf{CFAM Perception (Ours)} & \textbf{73.2} & \textbf{78.6} & \textbf{76.8} & \textbf{52.1} & \textbf{63.4} & \textbf{65.5} & \textbf{63.8} \\
\bottomrule
\end{tabular}

\end{document}